\documentclass[letterpaper, 10 pt, conference]{ieeeconf}
\IEEEoverridecommandlockouts          % This command is only needed if
\usepackage{graphicx}
\usepackage{xcolor}
\usepackage{epstopdf}
\usepackage{url}

\usepackage{amsmath}
\usepackage{amssymb}

\usepackage{algorithm, algpseudocode}  %algorithmic

\usepackage{cite}
\newcommand{\real}{\mbox{$I \!\! R$}}
\newcommand{\norm}[1]{\mbox{$\| #1 \|$}}
\def \eex{ \hfill\mbox{\large $\circ$} }
\def \epf{ \hfill $\square$ }

\newcommand{\Ag}{\mbox{$\alpha$}}

\newcommand{\As}{\mathcal{A}}

\newcommand{\Fs}{\mathcal{F}}
\newcommand{\Is}{\mathcal{I}}
\newcommand{\Ls}{\mathcal{L}}
\newcommand{\Ns}{\mathcal{N}}
\newcommand{\Os}{\mathcal{O}}
\newcommand{\Ps}{\mathcal{P}}

\newcommand{\Ss}{\mathcal{S}}
\newcommand{\Ts}{\mathcal{T}}
\newcommand{\Us}{\mathcal{U}}
\newcommand{\Vs}{\mathcal{V}}
\newcommand{\Ws}{\mathcal{W}}

\DeclareMathOperator*{\argmin}{Arg\,Min}

\newtheorem{theorem}{Theorem}

\newtheorem{lemma}[theorem]{\bf Lemma}
\newtheorem{prop}[theorem]{\bf Proposition}

\newcommand{\rtxt}[1]{ \hspace{1em}\mbox{#1} }

\newcommand{\barr}{ \begin{array}{cl} }
\newcommand{\earr}{ \end{array} }

\title{\LARGE \bf  Mobile Robot Sensory Coverage in 2-D Environments:\\
  An Optimization Approach with Efficiency Bounds}

\author{\bf \large E. Fourney and J. W. Burdick and E. D. Rimon
\vspace{-1.0in}
}
\begin{document}

\maketitle
\thispagestyle{empty}
\pagestyle{empty}

%%% ---------------------------------------------------%%
\vspace{-.2in}

\begin{abstract}
This paper considers three related mobile robot multi-target sensory coverage and inspection planning problems in {\small 2-D} environments.  In the first problem, a~mobile robot must find the shortest path to observe multiple targets with a~limited range sensor in an~obstacle free \mbox{environment.} In the second problem, the mobile robot must efficiently observe multiple~targets~while taking advantage of multi-target views in an~obstacle free environment.  The third problem considers multi-target sensory coverage in the presence of obstacles that obstruct sensor views of the targets.  We show how all three problems can be formulated in a~{\small MINLP} optimization framework.  Because exact solutions to these problems are {\small NP-hard}, we introduce polynomial time approximation algorithms for each problem. These algorithms combine polynomial-time methods to approximate the optimal target sensing order, combined with efficient convex optimization methods that incorporate the constraints posed by the robot sensor footprint and obstacles in the environment. Importantly, we develop bounds that limit the gap between the exact and approximate solutions.  Algorithms for all problems are fully implemented and illustrated with examples. Beyond the utility of our algorithms, the bounds derived in the paper contribute to the theory of optimal coverage planning algorithms. 
\vspace{-.02in}
\end{abstract}

%Using a Big-M formulation, that this problem can be rigorously formulated as a MINLP.
%The paper presents a  whose path is proven to have a  10/3 approximation ratio with respect to the optimal mobile robot path.  Extension of the first %approximation method provides a polynomial-time algorithm that simplifies the coverage path
%with a bounded ratio of number of the sensing locations relative to the optimal number.  Finally, we show how the complete sensory coverage problem of a %bounded planar environments populated by obstacles can is formulated  as an MINLP over a decomposition of the space into convex cells.  Extension of the %multi-target algorithm provides a polynomial time solution with a bounded approximation ratio.  Examples illustrate the methods.

%%%-----------------------------------------------------%%
\section{Introduction} \label{sec:intro}
\vspace{-.02in}

%\noindent This paper is motivated by {\small DARPA}'s Subterranean Challenge \cite{AutoSpot},
% where robot teams rapidly~navigate~and~map~complex
% underground environments with the goal of locating objects of 
% interest.

\noindent In the robot sensory coverage problems~considered~in~this~paper, a~mobile robot must choose its motions~so~that~its sensors can inspect multiple targets in a~bounded environment. Robot sensory coverage applications include mine sweeping~\cite{demining_choset}, crop sampling~\cite{tokekar_sensor_2016}, intruder detection~\cite{danner_randomized_2000} and sewage system inspection~\cite{loizou_cdc16}. Other applications include
aerial inspection of industrial installations~\cite{inspection_survey,bircher_3D}, aerial search~\cite{lauterbach_eins3d_2019}, seafloor mapping~\cite{Galceran_underwater} and ship hull inspection~\cite{englot_three-dimensional_2013}. See surveys by
Acar and Choset~\cite{acar&choset_06} and Galceran~\cite{galceran_survey}.

This paper studies three related mobile robot sensory coverage planning problems in \mbox{\small 2-D} environments. In the first two problems, a~mobile robot must efficiently observe multiple targets with a~limited range sensor in an~obstacle free environment. First observing one target at a~time, then exploiting
%The sensor observes one target at a~time.
%Sensing is performed at discrete nodes along the robot path.
%The sensor footprint need not be circular. It can be any convex shape and in particular a~sector of finite detection range.
%The robot coverage path must specify the sensor heading at discrete observation nodes along the robot path.
%In the second problem, a~mobile robot must efficiently
%observe with its limited range sensor multiple targets while
%then reduce the number of sensing positions by taking advantage of 
possible multi-target views along the robot path.
In the third problem, a mobile robot must efficiently observe with a~limited range sensor multiple targets among stationary obstacles that
%In this problem,
% the environment populated by obstacles.
obstruct sensor views of the targets.
% then minimize the number of sensing positions using multi-target viewsOn

Why do we study the multi-target sensory coverage planning problem, and why do we emphasize an optimal shortest path framework?  On every day until the year 2030, the world will add $\sim$1 gigawatt of solar power generation capacity.  Some solar installations will be as large as 5 miles on a side.  These facilities must be inspected on a regular basis, presumably by robots.  Simple boustrophedon paths will suffice to inspect the solar panel surfaces by aerial drones.  But the electrical inverter modules and electrical junctions boxes, whose failure can be detected by thermal cameras, are irregularly sited inspection targets.  The vast size of these inspection problems requires a shortest path strategy.

One of the novelties of this paper is to formulate these problems in a rigorous optimization framework. This framework brings the computational difficulty of coverage to the forefront.  We show that all problems can be posed as mixed integer nonlinear optimization problems ({\small MINLP}).  
%The integer variables represent the order of target observations while the continuous variables represent the robot sensing positions in the environment. 
Because these {\small MINLP}s are NP-hard, we introduce polynomial time algorithms to approximately solve these problems.  In addition to new algorithms, we provide rigorous efficiency bounds on the mobile robot path length that set new upper bounds on the length of a robot path to solve an inspection problem. These are the paper's two main conceptual contributions.

The {\small MINLP} approach and our approximate solution methods provide a~flexible framework for coverage planning that can be extended to target specific observation requirements, different sensor scanning patterns, terrain dependent travel cost and control input limits.  Prior works have often made restricted choices of these parameters
%The use of {small MINLP} in robotics is next described.

%In the fourth problem, the mobile robot has no apriori knowledge
% on the target locations in an~environment populated by obstacles, and it must efficiently search with its
%limited range sensor the entire environment in order to observe all targets.

{\bf Related work:}
We are not the first to recognize that sensory coverage planning (which is NP-hard) can be solved under {\small MINLP} formulation.  Gentilini~\cite{gentilini_travelling_2013}
adapted the {\small MINLP} formulations of Current ~\cite{current_covering_1989-1} and Golden~\cite{golden_generalized_2012} to plan the motion of a robot arm to inspect {\small 3-D} objects with a distal camera.
%a~body of complex geometry.
%The robot arm needs only position the camera at a~single point in each neighborhood in order to fully inspect the object.
%Gentilini relies on earlier works by for the formulation of the optimization constraints.
  %also 2020 paper \cite{carrabs_adaptive_2020}.
The camera visits viewing neighborhoods specified as convex sets. A direct optimization of the viewing path was proposed.
%the integer variable part of this {\small MINLP} is {\small NP}-hard. Hence only
%for a~small number of viewing neighborhoods.
%This  paper solves the integer and continuous parts of the {\small MINLP} formulation as separate sub-problems.
%Gentilini was able to use
However, direct {\small MINLP} solvers can only find globally optimal solutions in reasonable computation times for small numbers of regions~\cite{belotti_minlp,BONMIN,minlp_solvers}. Large number of targets can be solved using branch and bound methods, with much longer computation time~\cite{coutinho_branch-and-bound_2016}.

%a~significantly larger number~of
%whose number equals the number of the {\small MINLP} integer plus continuous variables.  This possibility is discussed in the conclusion.

%Wang~\cite{wang_icra07}
%uses pure integer optimization to
%plans the motion of a~camera mounted on a~mobile manipulator to inspect all facets of a~{\small 3-D} structure.
%The camera's motions are limited to a~fixed graph that surrounds the structure. The graph allows the use of pure integers as optimization variables but the %authors do not discuss how to select the graph in a~manner that ensures efficient inspection paths.

This paper solves off-line~\mbox{sensory}~\mbox{coverage} planning ~\mbox{problems} using a~two-\-stage approach
% according to their {\small MINLP}  formulation.
%integer and continuous variable parts.
that is related to a long tradition in robotics. Danner~\cite{danner_randomized_2000} uses a~two-stage scheme to plan mobile robot intruder detection routes in {\small 2-D} environments.
% populated by obstacles.
The first stage samples candidate robot sensing locations. The second stage computes in polynomial time
%in $O(k^2 \log k)$ time
a~tour that visits the  selected sensing locations.
%, with path length bound $l(path) \!\leq\! 2 \cdot l_{opt}$, where $l_{opt}$ is the length of the shortest tour through the sampled sensing locations.
% (a {\small $TSP$} tour).

Englot~\cite{christensen_planning_2017} used a two-stage scheme for underwater inspection of ship hulls. A~mesh representation of the hull defines a collection of targets to observe. The first stage uses random sampling to construct a graph that connects sensing nodes with multiple target views. The second stage ~\mbox{computes} a~tour through the sensing nodes.
%with path length bound~\mbox{$l(path) \!\leq\! \tfrac{3}{2} \cdot l_{opt}$,} where $l_{opt}$ is the length of the shortest tour through the sampled sensing nodes. 
%This stage is followed by pruning redundant viewpoints, 
%along the robot path, 
%an approach that is also used in our paper.
%Johnson's set cover algorithm~\cite{johnson_setcover},
Bircher~\cite{bircher_3D} also used a two-stage scheme to plan aerial inspection of industrial sites. Mesh-based facets of the environment form the collection of targets to observe. Initial sensor node locations are chosen according to the sensor viewing pattern.  The first stage computes a~tour that visits all sensing nodes.  The second stage adjusts the sensing positions to reduce the robot path length. This process repeats several times, but  efficiency bounds on the coverage path length are not discussed.
%The two-step approach taken in this paper emphasizes the {\small MINLP} formulation of the problem with explicit bounds on the coverage path length and %number of sensing nodes.

%\textcolor{blue}{
%As our paper uses {\small MINLP}, % to solve sensory coverage problems,
The {\small MINLP} formulation has been~applied~to~a~different problem of computing collision free paths for robot arms~\cite{tobia_obstacles}.
%Rather than divide the {\small MINLP} into integer and continuous sub-problems,
The robot arm joint space is partitioned into collision free convex cells whose visitation order is represented by the {\small MINLP} integer variables. The integer variables are  then {\em relaxed}
into continuous variables and the {\small MINLP} is solved as a~convex optimization problem having only continuous variables. However, the solution needs rounding back into integer variables in a~manner that does not guarantee efficiency bounds on the solution path.
%the total number of continuous variables can become large
%under the relaxation approach, and the resulting path

The mobile robot sensory coverage problem studied in this paper is related to the {\em Traveling Salesperson Problem ({\small TSP}) with  neighborhoods}%~\cite{arkin_approximation_1994}%
~\cite{mitchell-2017} in computational geometry. 
This {\small NP}-hard problem is approximately solved in polynomial time by finding a~tour that visits each neighborhood, and then bounding the length of this tour in terms of the optimal tour length through the specified neighborhoods.

In this paper, the neighborhoods represent {\em sensing regions} modeled as circles centered at the targets.\footnote{Our framework is readily extended to sensing patterns that can be modeled by a convex polygon.} When the circles have equal size and do not overlap, Dumitrescu~\cite{dumitrescu_approximation_2003} computes in polynomial time a~tour that visits each circle with path length
%that satisfies the upper bound
$l(path) \!\leq\!   (\mbox{\small $1$} \!+\! \tfrac{8}{\pi})  (\mbox{\small $1$}\!+\! \epsilon) \!\cdot\! l_{opt}$, where $l_{opt}$ is the optimal tour length. The $\mbox{\small $1$} \!+\! \epsilon$ coefficient represents an~approximation of an~initial {\small TSP} tour that passes through the circle centers. For instance,  $\mbox{\small $1$} \!+\! \epsilon \!=\! \tfrac{3}{2}$ when Christofides~\cite{christofides_n_worst-case_1976} is used to~app\-roximate the {\small TSP} tour. When the sensing circles possibly overlap, Dumitrescu~\cite{dumitrescu_approximation_2003} computes  a~tour that visits each circle with path length bound $l(path) \!\leq\!   \pi (\mbox{\small $1$} \!+\! \tfrac{8}{\pi}) (\mbox{\small $1$} \!+\! \epsilon) \!\cdot\! l_{opt}$, where $l_{opt}$ is the optimal tour length.
%in polynomial time

Isler adapted these techniques to sample crop data from circular areas (with path length efficiency bound~\cite{tokekar_sensor_2016}), and then to aerial robot inspection of ground targets visible through vertical cones of variable heights with path length efficiency bound~\cite{plonski&isler}.

\begin{figure}
%\vspace{-.05in}
\centering \includegraphics[width=.5\textwidth]{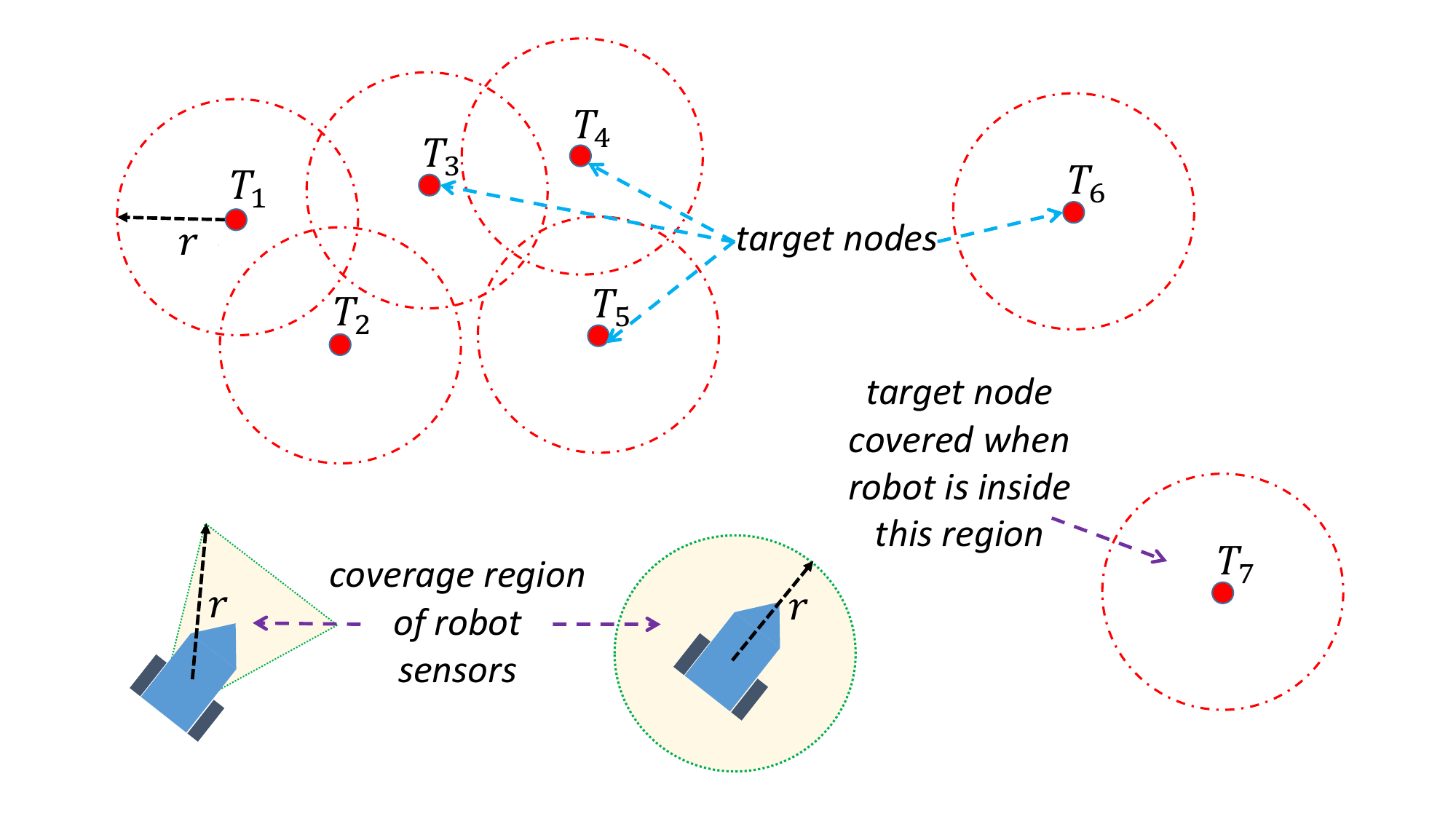}
% motivation_picture1.pdf
\vspace{-.375in}
	\caption{The problem geometry showing circular sensing regions centered at the targets and typical
robot sensor footprints with~detection~range~$r$.}
\label{fig:sensing_region}
\vspace{-.2in}
	\end{figure}

{\bf Paper contributions and format:} In the sensory coverage problems considered in this paper, the mobile robot starts at the first target location $T_1$, moves to observe intermediate targets $T_2,\ldots,T_{n-1}$ in any order, then stops within the detection range of a~final target $T_n$.
%Note that a~closed loop tour is a~special case of this $S$-to-$T$ sensory coverage problem.
This problem is termed {\em $S$-to-$T$ sensory coverage,} where $S \!=\! T_1$ and $T \!=\! T_n$. The mobile robot observes the targets at discrete sensing locations along its path while ensuring that the targets lie within the detection range~$r$ of its on-board coverage sensors.  The robot sensing locations thus~lie~in~{\em circular sensing regions} of radius~$r$ centered at the targets (Fig.~\ref{fig:sensing_region}).
%whose %equal
%size is determined by the sensor detection range 

The paper first describes the {\small MINLP} formulation of multi-target coverage in obstacle free environments with non-overlapping sensing regions. An approximate shortest sensory coverage path is computed in two stages. First, a polynomial time algorithm finds a~bounded approximation to the optimal $S$-to-$T$ sensory coverage path, which establishes the sensing node visitation order.
% the integer variables in the {\small MINLP} problem. 
% using Hoogeven's algorithm~\cite{hoogeveen_analysis_1991} which is fully explained.
%Hoogeven's path satisfies
%the bound $l(path) \!\leq\! \tfrac{5}{3} \!\cdot\! l^*$, where $l^*$ is the length of the shortest $S$-to-$T$  path that visits all targets.\footnote{
%oogeven's algorithm~\cite{hoogeveen_analysis_1991}  is based on Christofides' algorithm~\cite{christofides_n_worst-case_1976}, and the  $5/3$ path length %bound is shown to be tight.}
The sensing nodes positions are then optimized for the shortest path using efficient convex optimization. The robot's path length bound is shown to approach
%This part includes analysis of the robot path length expressed in terms of the true optimal path length. The path length
$l(path) \!\leq\!   \tfrac{10}{3} \!\cdot\! l_{opt}$, where $l_{opt}$ is the optimal $S$-to-$T$
sensory coverage path length. % of the targets.
The approximation coefficient improves on \cite{dumitrescu_approximation_2003} and is shown to be tight.
%Dumitrescu's  path length bound for coverage tours of disjoint equal size discs is {\em higher} than $\tfrac{10}{3}\!\cdot\! l_{opt}$.
%coefficient $ 1 \!+\! \tfrac{8}{\pi} \!=\! 3.55$

The paper next develops a novel {\small MINLP} formulation for multi-target sensory coverage when the sensing regions may overlap
in an~obstacle free environment.
% thus allowing the robot to take advantage of multiple targets observations.
% Multi-target views form disjunctive constraints that are incorporated into the {\small MINLP} formulation using integer viewing variables. %\cite{vecchietti_modeling_2003}.
%The two-stage approach solves the {\small MINLP} problem
%as follows.
%First a~{\em base set} of %non-overlapping
%sensing circles that overlap all other sensing circles  is computed.
%Sensing nodes are %initially located selected within
%at the center of the segments where
A modified two-stage approximation algorithm first places sensing nodes on the perimeter 
%and centers 
%in polynomial time
of {\em base set} sensing circles that overlap all other sensing circles. 
%The next two stages resemble the solution of the previous problem. 
Application of the algorithm described above (approximate path construction, followed by sensing location optimization) is followed by a process of  
%In the second stage, 
%The sensing node locations are then optimized for the shortest path using efficient convex optimization, followed by
sensing node reduction. The constructed path satisfies~the~bound  $l(path) \!\leq\! \frac{5}{3} \cdot 9 \cdot l_{opt}$ (with some constants neglected), where $l_{opt}$ is the length of the shortest path that observes all targets while taking advantage of multi-target views. The approximation bound improves on~\cite{dumitrescu_approximation_2003}
%which is roughly $\tfrac{11}{6} (9 \cdot l_{opt})$ when~\mbox{$1 \!+\! \epsilon \!=\! \tfrac{3}{2}$}~is~used,
but~is~{\em not}~a~tight~bound.

%%-----------------------------------------------------------------------------------------------%%

%% ------------------------------ %%
%\begin{figure}
%  \includegraphics[height=1.8 true in]{updatedfigure2.pdf}
%  %ProblemGeometry.jpg
%  \vspace{-.1in}
%  \caption{The problem geometry showing circular sensing regions centered at the targets and typical
%robot sensor footprints with~detection~range~$r$.}
%  \label{fig:sensing_region}
%    \vskip -0.2 true in
%\end{figure}
%% ------------------------------ %%

The third sensory coverage problem \mbox{includes} polygonal obstacles in the environment. The \mbox{targets} can only be viewed from obstacle bounded {\em viewing regions} of maximal detection range~$r$. The sensory coverage path among obstacles can still be computed by a polynomial time two-stage approach. Path length optimization occurs within the obstacle free space, and the number of sensing is reduced, when possible.
%of the {\small MINLP} optimization problem. 
% Initial sensing node locations are selected on the perimeter of base set viewing regions, that overlap all other sensing regions in the environment.  A~bounded approximation to the optimal $S$-to-$T$ sensory coverage path is computed using shortest paths that avoid obstacle collisions. This path fixes the sensing nodes visitation order. The sensing node locations are then optimized within obstacle free corridors using efficient convex optimization, followed by sensing node reduction along the optimized path. \\
The path length approximation bound for this problem uses two parameters. The number of obstacle vertices within detection range of the targets, $m$, and the circular radius representing the average area of the viewing regions, $\bar{\rho} \!\leq\! r$. 
The sensory coverage path satisfies the bound  $l(path)  \!\leq\!
\frac{5}{3} ( \mbox{\small $1$} \!+\! \mbox{\small $8$} \frac{r^2}{ \bar{\rho}^2}) \cdot (l_{opt} + 2r \!\cdot\! m)$ (with some constants neglected), where $l_{opt}$ is the length of the optimal $S$-to-$T$ sensory coverage path in the  environment. All algorithms are fully implemented and illustrated with execution examples.
%with software available as part of the paper. 

%The free environment
%that contains the obstacle free portion of each sensing circle
%is partitioned into convex cells. \textcolor{green}{The intersection of these cells with each sensing circle  forms {\em viewing cells} that surround  each target.}
%However, in this problem the robot {\em physical size} partitions the viewing cells into two types.
%Accessible cells that can be visited %and fully observed
%by the mobile robot and inaccessible cells that correspond to %that cannot be visited by the robot
%disconnected components of the mobile robot free c-space.
%be partially or fully observed only from adjacent accessible cells.
%This paper solves the sensory coverage problem for targets located in accessible cells.
%The extension to arbitrarily placed targets using {\small MINLP} is detailed but not implemented.

%The last sensory coverage problem assumes that the target positions are {\em not} apriori known to the robot.
%The mobile robot must efficiently observe {\em full cells} of the environment in order to observe all targets with its limited range sensor.
% while taking advantage of multi-target sensing locations.

A very preliminary version of this work appeared in \cite{9561213}.  The present paper sharpens Proposition \ref{prop:bound1} and provides the missing proof of this proposition.  The bulk of Section \ref{sec:fewer} on sensor node reduction is entirely new.  Moreover, all of Section \ref{sec:obstacles} and its novel bounds are new material.  

The paper is structured as follows. Section~\ref{sec:covering_path}
formulates the first multi-target sensory coverage problem as a~{\small MINLP} problem. Section~\ref{sec:polynomial} describes a~two-step approximation algorithm for this problem and develops a path length approximation bound. Section~\ref{sec:fewer} adds multi-target views to the  {\small MINLP} formulation, then describes an analogous approximation algorithm, a sensing node reduction algorithm, and a path length approximation bound.  Section~\ref{sec:obstacles} generalizes the {\small MINLP} formulation to multi-target sensory coverage in the presence of obstacles.  The section describes an approximation algorithm, followed by analysis of the algorithm's path length approximation bound. The conclusion discusses an extension of the {\small MINLP} approach to partially known environments and {\small 3-D} sensory coverage planning problems.

%The majority e.g. %\cite{carrabs_adaptive_2020,coutinho_branch-and-bound_2016,smith_glns_2017,au_improved_2017,behdani_integer-programming-based_2014,wang_steiner_2019,mennell_heuristics_2009})
%do not provide any bounds on the solution quality, or even on the algorithm's computational complexity.

%%------------------------------------------------%%
\section{ {\small MINLP} Formulation of the Multi-Target Sensory Coverage Problem }\label{sec:covering_path}
\vspace{-.02in}

\noindent This section describes the {\small MINLP} formulation of the basic multi-target sensory coverage problem.  The mobile~robot moves in a~{\small 2-D} environment free of obstacles using
an ideal position sensor and a~finite detection range coverage sensor, whose
%(Fig.~\ref{fig:sensing_region}).
%The sensor is mounted on the mobile robot platform  and
%The robot sensor position in a~fixed world frame is denoted $(x,y)$.
footprint forms a~circular or convex sector with detection range $r$ (Fig.~\ref{fig:sensing_region}).
%The sensor footprint can also be circular or  a~convex polygon of maximal detection range $r$.
The robot must~observe~targets $T_1,\ldots,T_n$ from
discrete {\em sensing nodes} $S_1,\ldots,S_{n_s}$ ($n_s \!\leq\! n$)~along~its path. The robot starts at target $T_1$ (which is also labeled as sensing node $S_1$) and ends within detection range of $T_n$, but otherwise observes the intermediate targets $T_2,\ldots,T_{n-1}$ in any order.  The {\em sensing region} that surrounds each target,  $R(T_i)$, is the circle of radius $r$ centered at $T_i$ (Fig.~\ref{fig:sensing_region}). This is a specific type of {\em viewing region}, $V(T_i)$, which describes the area from which target $T_i$ can be observed.  When the robot sensor is located in $R(T_i)$, there exist sensor viewing directions that can observe target~$T_i$.

{\bf The multi-target sensory coverage problem:}  The simplest problem assumes one sensing node per target, $n_s \!=\! n$, in an~obstacle free environment.  The goal is to find the {\em shortest} path, $\mathcal{P}$, that allows the robot to observe all targets $T_1,\ldots,T_n$ from sensing nodes $S_1,\ldots,S_n$,  such that $S_1 \!=\! T_1$~and~\mbox{$S_n \! \in \! R(T_n)$.}

%A corner point along the shortest path may be the single point at which the robot visits a~particular sensing region. In this case the corner
%point must form a sensing node.
%corner points must be used as sensing nodes, as the robot. All other sensing nodes
%vary along segments where the shortest path passes through individual sensing regions.

%The paper objective  is to compute  a~reasonably short sensory coverage path, $\mathcal{P}$,
%with path length bound expressed in terms of the shortest sensory coverage path length.
%can be computed using efficient polynomial time techniques and have a~

The path $\mathcal{P}$ is chosen to
minimize the cost
  \begin{equation}\label{eq:min_cost}
     Cost(\{ S_1,\ldots,S_n \}, \{ \xi_{ij} \} ) =  \ \sum_{i=1}^n \sum_{j=1,j\neq i}^n \! \xi_{ij} \!\cdot\! d(S_i,S_j)
  \end{equation}
over the sensing node locations $S_1,\ldots,S_n \in \real^2$ ({\small $S_1 \!=\! T_1$})
and {\em incidence variables,} $\xi_{ij}$ for $i,j\in \{ 1,\ldots,n \}$, such that
\[
  \xi_{ij} = \begin{cases} 1 &
                  \mbox{\rm $T_j$ is observed immediately after $T_i$ is observed} \\
                              0 & {\rm otherwise} \end{cases}
\]
with $d(S_i,S_j) \!\geq\! 0$ denoting the distance between sensor nodes $S_i$ and $S_j$.
Terrain dependent  travel costs can be handled by non-symmetric travel costs.
This paper assumes symmetric travel costs, $d(S_i,S_j) \!=\!  d(S_i,S_j)$.
% with incidence variables
%\mbox{\small $\norm{S_i \!-\! S_j}$}$. In the symmetric travel cost case
% \[
% \xi_{ij} \!=\! \xi_{ji} \!=\! \begin{cases} 1 \! &
%                \mbox{\rm $\!\! % T_j$~is~observed~immediately~after~or~before~$\! T_i$ } \\
%               0 \! & $\!\!$ {\rm otherwise} \end{cases}.
% \]
In the symmetric case it suffices to use the
reduced indices $j \!>\! i$ in Eq.~\eqref{eq:min_cost}.

\begin{figure}
\centerline{  \includegraphics[height=1.4 true in]{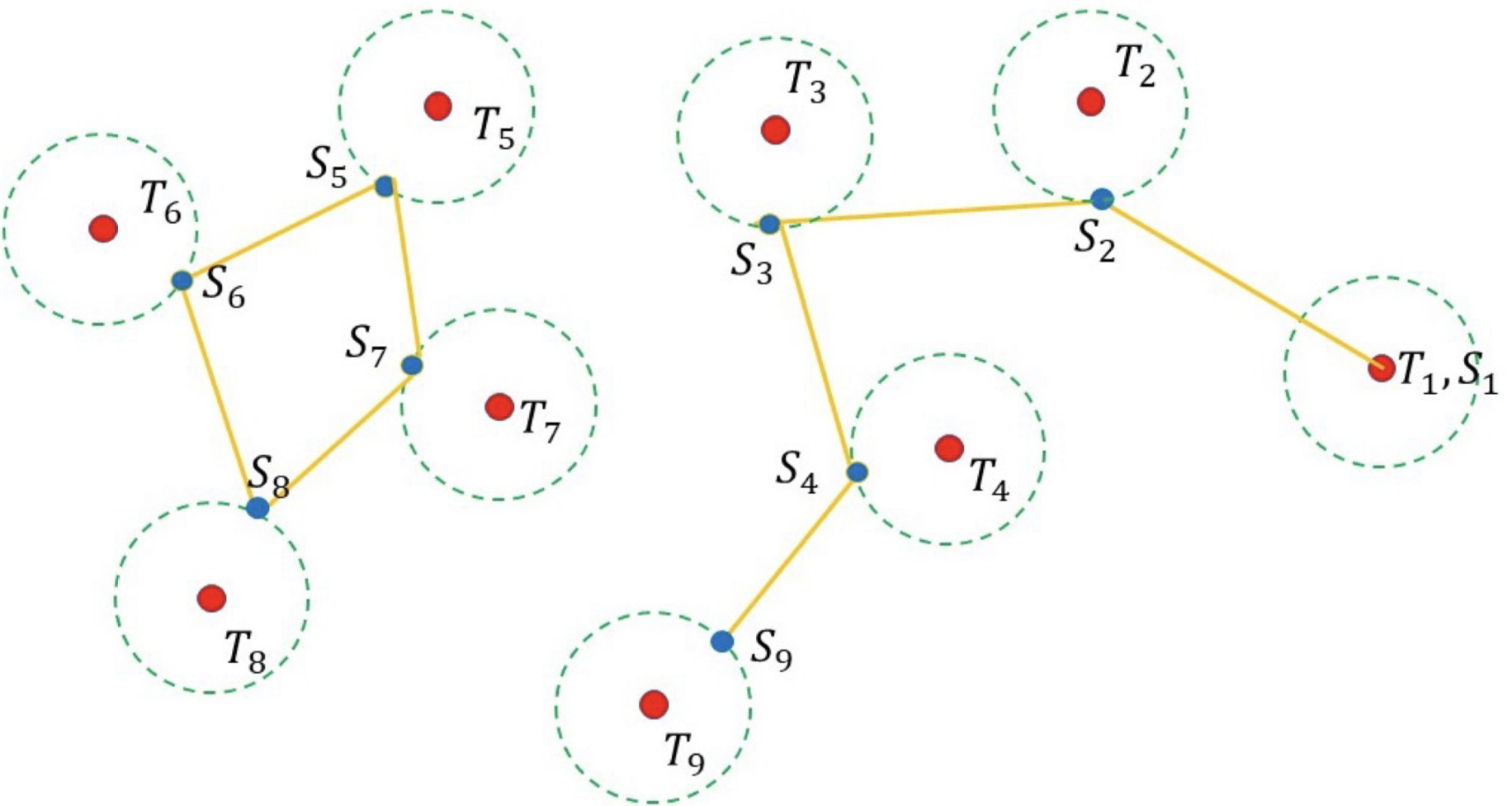}}
\vspace{-.1in}
\caption{A non-valid sensory coverage path~starts~at~\mbox{$S_1 \!=\! T_1$}~and~ends at~$S_9 \!\in\! R(T_9)$. The disjoint
loop through the sensing nodes \mbox{\small $W \!=\! \{ S_5,S_6,S_7,S_8 \}$} does {\em not}~satisfy~the~\mbox{constraint}~\mbox{\scriptsize  $\sum_{i\in W,j\in W}  \xi_{ij}$}$ \!\leq\! 3$.} %  where $|W| \!=\! 4$.}
    \label{fig:subtour}
  \vskip -0.2 true in
\end{figure}

For the sensory coverage path to be well defined, the cost minimization
%of the cost in Eq.~(\ref{eq:min_cost})
must satisfy the following constraints.
\begin{enumerate}
\item{} {\bf Full path constraints:}  The path $\mathcal{P}$ must pass through
  each sensing node
\vspace{-.08in}
\begin{equation} \label{eq:edge0}
\sum_{j=2}^n \xi_{1 j} = 1, \quad \quad \sum_{i=1}^{n-1}\ \xi_{i n} = 1,
\vspace{-.13in}
\end{equation}
\noindent and
\vspace{-.1in}
\begin{equation} \label{eq:node_constraint0}
\sum_{j=1}^{i-1} \xi_{ji} + \sum_{j=i+1}^n\ \xi_{ij} = 2 \rtxt{\hspace{1.5em} $i=2 \ldots n \!-\! 1$.}
\vspace{-.08in}
\end{equation}
\noindent Eq.~\eqref{eq:edge0} ensures that a~single edge of $\mathcal{P}$ exits $S_1$
and a~single edge enters $S_n$. Eq.~\eqref{eq:node_constraint0} ensures that a~single edge of $\mathcal{P}$
enters  (first summand) and a~single edge %of $\mathcal{P}$
exits (second summand)~each~\mbox{intermediate}~\mbox{sensing}~node.
\item{} {\bf Connected path constraints:} Additional constraints are needed to ensure that $\mathcal{P}$  forms a~connected path.
%  sub-paths, such as the one seen in Fig.~\ref{fig:subtour}.
Let $\Ss \!=\! \{ S_1,\ldots,S_n\}$ be the set of
  sensing nodes. Let $W$ be the index set of three
  or more distinct sensing nodes, and let $|W|$ denote the set size. The constraints  %excluding $S_1$ and $S_n$
  \[
  \sum_{i\in W,j\in W} \! \xi_{ij} \leq |W| \!-\! 1, \ \forall W \subseteq \mathcal{S} \!-\! \mbox{\small $\{S_1,S_n\}$}, |W|\geq 3
  \]
  prevent disconnected sub-paths by preventing sets of $|W| \!\geq\! 3$ sensing nodes from forming
  disjoint loops (Fig.~\ref{fig:subtour}).  The number of such constraints
  % $ 2^{n-1}-(n-1)(n-2)/2-1$.
  \[
         \sum_{k=3}^{n-1} {{n \!-\! 2}\choose k} = 2^{n-1} \!-\! \frac{1}{2} (n \!-\! 1)(n \!-\! 2) -1
\]
is  exponential in~$n$, which highlights the {\small NP}-hardness of the multi-target sensory coverage problem.
\item{} {\bf Sensing region constraints:}  Each sensing node $S_i$ must lie inside the circular sensing region
  $R(T_i)$, expressed by the quadratic constraint
  % of the sensing node $S_i$
  %$T_i\ \in\ R(S_i)$.  For disc sensory patterns, this constraint is
     \begin{equation}\label{eq:quadratic}
        ||S_i - T_i||^2\ \leq \ r^2 \rtxt{\hspace{1em} $S_i \in \real^2$, \ $i=1 \ldots n$.}
     \end{equation}
%A convex polygon sensing region bounded by $p$ edges is modeled by
%$p$ linear constraints at each $S_i$:
%  \begin{equation}\label{eq:polytope}
%    \vec{v}_{i,j}(q_i)\cdot(T_i-C_{i,j}(q_i))< 0\ \ \  {\rm for}\  j=1,\ldots, p
%  \end{equation}
%where $v_{ij}$ is normal to the $j^{th}$ line $H_j$ bounding the sensing polygon at $S_i$, and $C_{ij}$ is a point in $H_j$.
\end{enumerate}

%% ----------------------------------- %%
%% ----------------------------------- %%

\noindent {\bf Problem \boldmath{$\# 1$:}} Compute the minimum cost $S$-to-$T$ path $\Ps$ that visits variable sensing nodes $S_1,\ldots,S_n$ such that all targets $T_1,\ldots,T_n$ are observed at the sensing nodes.
%the respective sensing nodes.
This problem is formulated as the {\small MINLP} optimization problem
% solves the symmetric distance multi-target coverage problem\\
%  \begin{equation}\label{eq:OptProblem}
\[
    (\mathcal{S}^*,\{\xi^*_{ij} \}) =   \argmin_{\mathcal{S},\{\xi_{ij} \}}\bigg(
             \sum_{i=1}^n \sum_{j=i, j > i}^n \xi_{ij} \! \cdot \! d(S_i,S_j) \bigg)
\]
where $d(S_i,S_j) \!=\! \mbox{\small $\norm{S_i \!-\! S_j}$}$ is the Euclidean distance.
The sensing nodes $\mathcal{S} \!=\! \{S_1,\ldots,S_n\}$ and the incidence variables $\{ \xi_{ij} \}$ %for $i,j\in \{ 1,\ldots,n \}$
are subject to the constraints
\vspace{-.1in}
\begin{eqnarray}
  && \sum_{j=2}^n \xi_{1 j} = 1, \quad \quad \sum_{i=1}^{n-1}\ \xi_{i n} = 1  \label{eq:edge} \\
  && \sum_{j=1}^{i-1} \xi_{ji} + \sum_{j=i+1}^n\ \xi_{ij} = 2 \rtxt{\hspace{1.5em} $i=2 \ldots n  \!-\! 1$}             \label{eq:node_constraint}\\
  && \!\! \sum_{i\in W,j\in W} \!\!\! \xi_{ij}\! \leq \! |W| \!-\! 1, \ \forall W \subseteq \mathcal{S}  \!-\! \mbox{\small $ \{S_1,S_n\}$}, |W| \!\geq\! 3      \ \ \          \label{eq:subtour}\\
   &&  \xi_{ij} \in  \{0,1\} \hspace{1.5em} \rtxt{\hspace{1.5em} $1\leq i,j \leq n$}    \label{eq:binary}  \\
   && S_i\ \in R(T_i)  \rtxt{\hspace{2.75em} $S_i \in \real^2$, \ $i = 1\ldots n$}  \label{eq:inrange}
\end{eqnarray}
where the sensing range constraints take the quadratic form of Eq.~\eqref{eq:quadratic}. A convex polyhedral sensing pattern can alternatively included by a set of linear constraints. Generically, for any convex sensing footprint, these constraints are simply $S_i \in V(T_i)$.
%
%% ----------------------------------- %%
\begin{figure}[H]
%  \vskip -0.1 true in
\centerline{\includegraphics[height=2.6 true in]{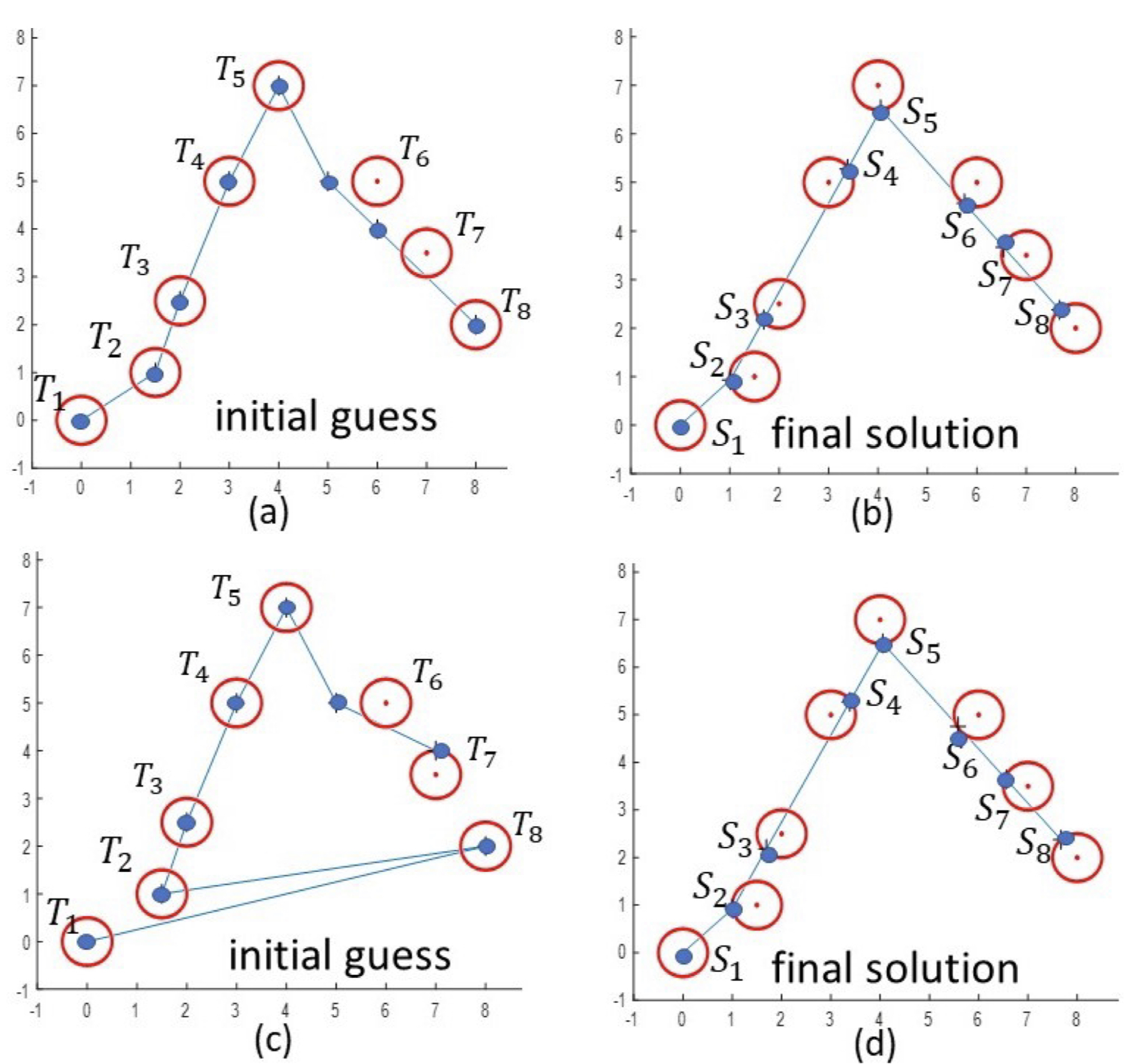} }
\vspace{-.16in}
  \caption{{\scriptsize MINLP} solution using {\scriptsize BONMIN} for eight targets that  lie at the centers of the red sensing circles.
  (a)  Initial path guess, and (b) the optimal sensory coverage path where each sensing node (blue dots) lies in a~target centered sensing circle.
  (c) Another initial path guess with different visitation order for $T_8$, and (d) the {\scriptsize MINLP} solution gives the same globally optimal path.}
  \label{fig:MINLP_example}
\end{figure}

\noindent {\bf Exact MINLP example:} Since Problem~$\# 1$ is \textsc{NP}-hard, direct \textsc{MINLP} solvers can only solve simple instances in practical computation time.  The example in Fig.~\ref{fig:MINLP_example} uses \textsc{BONMIN} ~\cite{BONMIN}, a branch-and-bound solver with nonlinear trust region that computes globally optimal solutions. \textsc{BONMIN} requires an initial, but not necessarily feasible, sensing nodes path guess (Fig.~\ref{fig:MINLP_example}(a)). The optimal solution in Fig.~\ref{fig:MINLP_example}(b) was computed in 5.8~seconds.\footnote{All evaluations using  \textsc{BONMIN} use \textsc{MATLAB} 19a, the \textsc{MATLAB} Opti ToolBox\cite{OptiToolbox} with the \textsc{BONMIN} solver, and an Intel i7-4770 CPU with 16 Gbyte memory. Evaluations of the proposed algorithm use MATLAB R2023b and the \textsc{CVX} solver \cite{cvx} run on an Intel core i9-13900H with 64 GB RAM. }
Another initial path with different visitation order  (Fig.~\ref{fig:MINLP_example}(c)) gives the same optimal solution (Fig.~\ref{fig:MINLP_example}(d)), in roughly the same computation time.~\eex

%%--------------------------------------------------------- %%
\section{Multi-Target Sensory Coverage with Path Length Bound}\label{sec:polynomial}
\vspace{-.02in}

\noindent This section describes a~two-stage \textit{polynomial time} algorithm that approximatly solves Problem~$\# 1$, then provides a bound on the worst case path length produced by the algorithm.
% with computational complexity that is {\em polynomial} in $n$, the number of targets.
%Section~\ref{sec:Hoogeveen} describes
{\em Hoogeven's algorithm} \cite{hoogeveen_analysis_1991} is first used to determine the sensing nodes
%are initially located at the targets and their
visitation order. The Hoogeveen path solution
%fixes the sensing nodes visitation order with
assigns fixed values to the incidence variables, thereby approximating a solution of the S-T TSP problem on the targets.  Then, initial sensing node locations are placed at the target locations.
%Section~\ref{sec:fixed_optimize}
Next, convex optimization finds the sensing node locations that yield the shortest sensory coverage path, given the visitation order. The section then analyzes this algorithm's path length approximation bound.

{\bf Hoogeveen  $S$-to-$T$ shortest path approximation:} Hoogeveen's algorithm \cite{hoogeveen_analysis_1991}, summarized in Algorithm ~\ref{alg:hoogeveen}, generates in polynomial time a~path  that visits all sensing nodes, {\em which are initialized to the target locations}, with a path length approximation bound that is described below. The algorithm first computes in line ~\ref{line:MaxCompleteGraph} a~complete graph, $G$, by joining
every pair of sensing nodes $S_i$ and $S_j$ with edge weight $d(S_i,S_j) \!=\! \norm{S_i \!-\! S_j}$. This computation takes $O(n^2)$ time where $n$ is the total number of sensing nodes (one per target). The shortest $S$-to-$T$ path that visits all sensing nodes lies in $G$, though it is {\small NP}-hard to find this path \cite{hoogeveen_analysis_1991}.
%%% -------------------------------------------------------------- %%%
\begin{algorithm}[htb]
\caption{Hoogeveen S-to-T Path Approximation}
\begin{small}
\hspace{.2em} {\bf Input:} Initial Sensing node locations $\Ss = \{S_1,\ldots,S_n\}$
\begin{algorithmic}[1]
\Procedure{Hoogeveen}{$\Ss$} %{$\mathcal{N}$ = target node set}
\State $G\ =\ {\rm CompleteWeightedGraph}(\mathcal{S})$ \label{line:MaxCompleteGraph}
\State $T_{min}\ =\ {\rm MinSpanningTree}(G)$ \label{line:MST}
\State $V_{join}\ =\ {\rm WrongNodeDegree}(T_{min})$ \label{line:Tjoin}
\State $M_{opt}\ =\ {\rm MinPerfectMatching}(V_{join})$ \label{line:MPM}
%\State $T_{union} = M_{opt} + T_{min}$    \label{line:union}
\State $\mathcal{P}_{E}\ =\ {\rm EulerianPath}(M_{opt} \!+\! T_{min})$ \label{line:EulerianTour}
\State $Return({\rm Shortcut}(\mathcal{P}_{E}))$ \label{line:Hamiltonian}
\EndProcedure
\end{algorithmic}
\end{small}
\label{alg:hoogeveen}
\end{algorithm}
%\vspace{-10pt}
%%% -------------------------------------------------------------- %%%
%\vskip -0.2 true in

Hoogeveen's algorithm computes in line~\ref{line:MST} a~minimum spanning tree, $T_{min}$, that spans all nodes of $G$ with minimum total edge cost. This stage is computed with Prim's algorithm in $O(n \log n)$ time. The  minimum spanning tree is augmented with
edges that ensure {\em odd degree} of nodes $S_1$ and $S_n$
and {\em even degree} of all other nodes. This stage is performed in
Line~\ref{line:Tjoin}, where nodes $S_1$ and $S_n$ have {\em wrong degree} when their node degree is even in $T_{min}$, all other nodes have {\em wrong degree} when their degree is odd in $T_{min}$. The set of wrong degree nodes, $V_{join}$, is computed in $O(n)$ time and always contains an even number of nodes. In line~\ref{line:MPM}, the edges to be added to $T_{min}$ are computed as a {\em minimum perfect matching} over the nodes of $V_{join}$,\footnote{In {\em minimum perfect matching} all nodes of $V_{join}$ are pairwise joined by edges whose  total cost over all possible pairings is minimal.}  using Blossom's algorithm~\cite{kolmogorov_blossom_2009} in $O(n^3)$ time. The pairwise matchings of the nodes in $V_{join}$ forms the set of edges, $M_{opt}$, that are added to $T_{min}$ in line~\ref{line:EulerianTour}.

The larger graph,  $T_{min} \!+\! M_{opt}$,  is next used to compute the desired path.  An $S$-to-$T$ path that traverses every edge of the latter graph exactly once (nodes may be revisited) forms an {\em Eulerian path.}  The graph $T_{min} \!+\! M_{opt}$ possesses an~Eulerian $S_1$-to-$S_n$ path iff $S_1$ and $S_n$ have odd degree while all other nodes have even degree. The latter graph is constructed in a~manner that ensures this property, and line~\ref{line:EulerianTour} computes in $O(n)$ time an~Eulerian $S_1$-to-$S_n$ path. The Eulerian path is finally shortened in line~\ref{line:Hamiltonian} into a~path from $S_1$ to $S_n$ that visits a~new sensing node in each step, by taking shortcuts along the graph $G$ in $O(n^2)$ time. Hoogeveen ~\cite{hoogeveen_analysis_1991} established that the resulting path
visits every sensing node, starting at $S_1$ and ending at $S_n$, with path length bound
%\vspace{-.1in}
   \[ l(path) \!\leq\!  l(T_{min}) \!+\!  l(M_{opt}) \leq l_{TSP} + \tfrac{2}{3} l_{TSP} = \tfrac{5}{3} l_{TSP}
    \]
\noindent  where $l_{TSP}$ is the length of the shortest path from $S_1$ to $S_n$ that visits all sensing nodes located at the targets.

{\bf Efficient optimization of sensing node locations:}~Algo\-rithm~\ref{alg:hoogeveen} computes a~sensory coverage path that~does~{\em not} optimize the sensing node locations.  When the targets are sparsely located with non-overlapping sensing circles (overlapping sensing circles are considered in Problem~\#2), the sensing nodes visitation order found by Algorithm~\ref{alg:hoogeveen} is a good initial path for the next step that optimizes the sensing locations. The path computed by Algorithm ~\ref{alg:hoogeveen} {\em fixes the incidence variables,} $\xi_{ij}$, thus reducing the {\small MINLP} optimization cost to a function only of the sensing node locations, $S_1,\ldots,S_n \!\in\! \real^2$:
\vspace{-.08in}
   \begin{equation}\label{eq:min_Scost}
       Cost(S_1,\ldots,S_n) \ =  \ \sum_{j=1}^{n-1}  \norm{S_{j+1} - S_j }  .
\vspace{-.08in}
\end{equation}
\noindent Each term \mbox{\small $||S_{j+1} \!-\! S_j||$} is a~convex function,~since~it~is~the composition of a~convex norm function with the linear function $S_{j+1} \!-\! S_j$.  The sum of convex functions is a~convex function, hence the cost (\ref{eq:min_Scost}) is a~convex function of $S_1, \ldots, S_n$.
The sensing circles form convex sets for the individual sensing node locations.  Their conjunction therefore forms a~convex set in the composite space $S_1,\ldots,S_n$.  Computation of the shortest sensory coverage path under fixed sensing node visitation order
thus forms  {\em a~convex optimization problem,} which is summarized as Algorithm~\ref{alg:s_t_heuristic} and computationally solved using \cite{gb08},\cite{cvx}.

\setlength{\textfloatsep}{5pt}
\begin{algorithm}[t]
\caption{Shortest Cover}
\begin{small}
\hspace{.2em} {\bf Input:} Targets $\mathcal{T} = \{T_1,\ldots,T_n\}$
\begin{algorithmic}[1]
\Procedure{ShortestCover}{$\mathcal{T}$} %{$\Ns$ = target node set}
\State Set \ $S_i = T_i$ \ for \ $i=1\ldots n$, \ $\mathcal{S} = \{S_1,\ldots,S_n\}$
\State $\mathcal{P}_{order}\ =\ {\rm Hoogeveen}(\mathcal{S})$  \label{line:init}
\State $\{\xi_{ij}\}\ =\ {\rm Incidence}(P_{order})$ \label{line:getincidence}
\State $\Ss_{opt} = \argmin {\rm Cost}(\{\xi_{ij}\},\mathcal{S},\mathcal{T})$ \label{line:argmin}
\State $\quad \quad {\rm subject \  to}  \ S_1 = T_1 \ {\rm and} \ S_i \!\in\! R(T_i) \ {\rm for} \ i=2 \ldots n$ \label{line:SenseConstraint}
\State $Return( {\rm path} \ \mathcal{P},  \  {\rm sensing \ nodes} \ \Ss_{opt} )$ \label{line:path}
\EndProcedure
\end{algorithmic}
\end{small}
\label{alg:s_t_heuristic}
\end{algorithm}

\begin{figure}[h]
\vskip -0.1 true in
    \centerline{
    \includegraphics[height=2.2 true in]{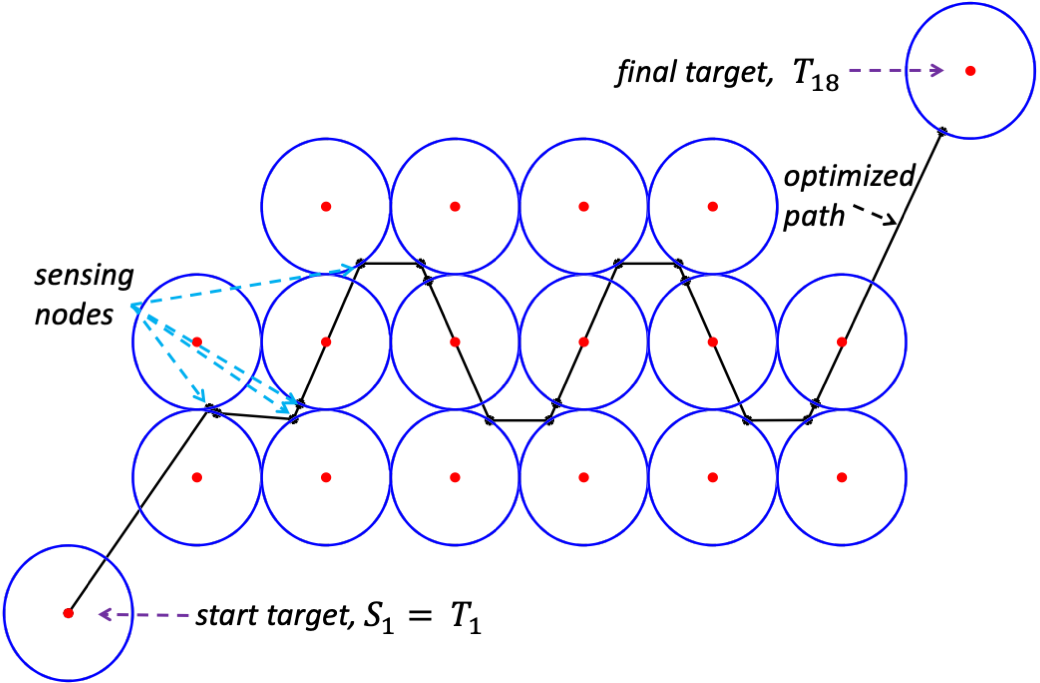}
    }
\vspace{-.125in}
    \caption{Optimized sensory coverage path computed by Algorithm~\ref{alg:s_t_heuristic} for
      $18$ targets surrounded by sensing circles whose radii equal the robot
      detection range, $r=0.49$.  The sensory coverage path is the black curve while the $18$ blue dots indicate the
      optimized sensing node locations.}
    \label{fig:ManyDisjointNodes}
\vspace{-.1in}
\end{figure}

%%% ---------------------------------------------------- %%
{\bf Examples:} Algorithm~\ref{alg:s_t_heuristic} requires an~initial coverage path, provided by the Hoogeveen path computed in line~\ref{line:init}, to perform convex optimization in line~\ref{line:argmin}.  When Algorithm~\ref{alg:s_t_heuristic} is applied to the problem of Fig.~\ref{fig:MINLP_example}, it yields the same optimal path as {\small BONMIN} in $0.1$~seconds (versus 5.8~seconds required by {\small BONMIN}). Algorithm~\ref{alg:s_t_heuristic} is next applied to the $18$ targets placed at integer locations shown in Fig.~\ref{fig:ManyDisjointNodes}, with sensing circles of detection range $r \!=\! 0.49$ units (the {\small BONMIN} solver was {\em not} able to solve this problem in reasonable time). Algorithm~\ref{alg:s_t_heuristic} computed a~sensory coverage path for these targets in $0.575$~seconds, with the optimized sensing nodes (blue dots) and  the coverage path (black curve)~shown~in~Fig.~\ref{fig:ManyDisjointNodes}.~\eex
% The BONMIN MINLP solver could not solve this problem.
%
%% ------------------------------ %%
%% ------------------------------ %%

{\bf  Path length approximation bound:} The {\em approximation bound} of Algorithm~\ref{alg:s_t_heuristic} is defined
as the largest ratio of the algorithm's path length measured relative to the optimal path length, taken over all problem instances. Our approximation bound is based on the following lemma.

\begin{lemma}[\cite{hoogeveen_analysis_1991}] \label{lemma:hoogeveen}
Algorithm~\ref{alg:hoogeveen} computes in $O(n^3)$ time a %~Hoogeveen
path that visits all initial sensing nodes. Its length satisfies $l(Alg.~\ref{alg:hoogeveen})  \ \leq \  \tfrac{5}{3} l_{TSP}$, where $l_{TSP}$ is the length of the shortest $S$-to-$T$ path that visits all sensing nodes.
% located at the targets.
\vspace{.05in}
\end{lemma}

\noindent The next proposition describes the path length approximation bound for Algorithm~\ref{alg:s_t_heuristic}.
\vspace{.01in}

\begin{prop} \label{prop:bound1}
%Consider $n>4$
%When the targets $T_1,\ldots,T_n$ are located
%at least $2r$
%twice the maximal sensing radius
%with non-overlapping sensing circles,
Algorithm~\ref{alg:s_t_heuristic} computes in $O(n^3)$ time
a~sensory coverage path whose length satisfies the approximation bound
\vspace{-.12in}
%of covering path, $|\mathcal{P}_h|$,  is bounded relative to the optimal
%covering path length $|\mathcal{P}^*|$ as
   \begin{equation}\label{eq:bound}
     l(Alg.~\ref{alg:s_t_heuristic})  \ \leq \ \tfrac{10}{3}\mbox{\small $(1 \!+\! \delta(n)$} ) \cdot l_{opt}  % |\mathcal{P}_h|
\vspace{-.08in}
   \end{equation}
\noindent where $\delta(n) \! \ll \! 1$ for large $n$, $\delta(n)\rightarrow 0$ as $n\rightarrow\infty$, and $l_{opt}$ is the optimal $S$-to-$T$ sensory coverage path
in the obstacle free environment.
%Moreover, $|\mathcal{P}_h|$ is computed in time proportional $\mathcal{O}(Bn^3)$, where $B$ is the
%desired number of bits of accuracy in the sensing node location.
\end{prop}

% \frac{10(n-1)}{3(n-5 + ((n-1)\bmod 2) + 2\sqrt{3})}|\mathcal{P}^*|\ .

%For large $n$, the approximation ratio denoted by ${\Rs_{app}}$, is bounded by $10/3$. When
%applied to our case of uniform sized sensing radii at each target, the approximation ratio developed by Tokekar et. all \cite{tokekar_sensor_2016} is %$(214.8+\epsilon)$, which is nearly 2 orders of magnitude larger than our bound.
%% which is better than the best ratios in \cite{de_berg_tsp_2005,Mitchell}.xo{sec:{sec:covering_path}covering_path}

\vskip 0.07 true in
\noindent {\bf Proof:}
%Correctness of Algorithm~\ref{alg:s_t_heuristic} follows by construction:
Line 3 of Algorithm~\ref{alg:s_t_heuristic} computes in $O(n^3)$ time an~initial sensory coverage path for the convex optimization procedure.
%No step in Algorithm~\ref{alg:hoogeveen} has complexity greater than $O(n^3)$.
The complexity of Algorithm~\ref{alg:s_t_heuristic} therefore depends upon the relative complexity of the convex optimization procedure.  When convex optimization is performed with interior point methods \cite{ellipsoid_survey}, %such as the ellipsoid algorithm
an optimal solution is computed within $\epsilon$ accuracy in $k n^2 \!\cdot\! \log(1/\epsilon)$ steps, where $n$ is the number of optimization
variables and $k$ is the number of steps required to evaluate the cost and constraints, which is $O(n)$ in the {\small MINLP} formulation of Section~\ref{sec:covering_path}. Algorithm ~\ref{alg:s_t_heuristic} thus takes $O(n^3 \!\cdot\! \log(1/\epsilon))$ time.\footnote{The {\em simplex method} solves second-order convex problems extremely well in practice but its computation time is non-polynomial.} %in~$n$.}

%\note{ Also note that in the non-arXiv version of the paper, the figure caption must have the  "(see appendix)" replaced by a references to the arXiV appendix}
%%----------------------------------------------------------------%%
\begin{figure}
%\vspace{-.05in}
\centering \includegraphics[width=.5\textwidth]{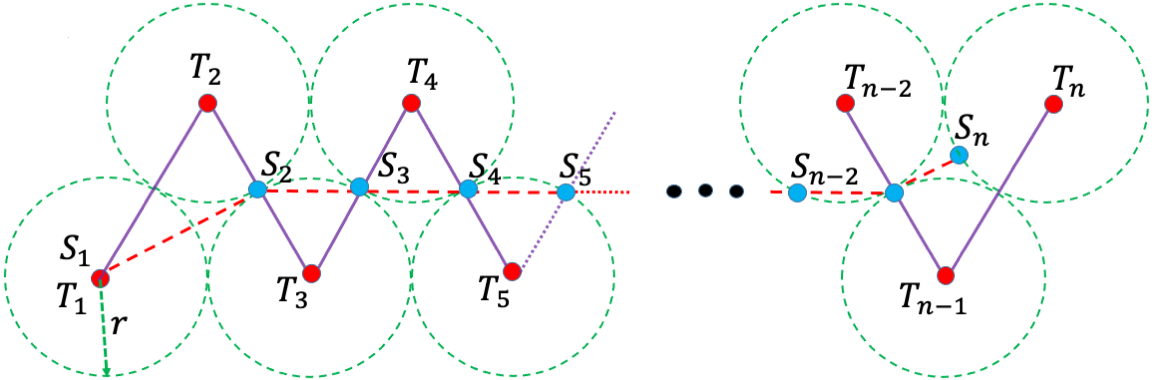}
% motivation_picture1.pdf
\vspace{-.15in}
\caption{Geometry of a worst case ratio of path length through the targets, $l_{TSP}$,  to the optimal sensory coverage path length $l_{opt}$ (see appendix).}
\label{fig:RatioDiagram}
%\vspace{-.2in}
\end{figure}
%%----------------------------------------------------------------%%

By Lemma~\ref{lemma:hoogeveen}, the initial sensory coverage path satisfies the bound $l(Alg.~\ref{alg:hoogeveen}) \!\leq\! \tfrac{5}{3} l_{TSP}$. Algorithm~\ref{alg:s_t_heuristic} starts with this path,  then computes a~{\em shorter path} that visits optimized sensing nodes locations. Hence
\vspace{-.1in}
\begin{equation} \label{eq.53}
 l(Alg.~\ref{alg:s_t_heuristic})   \leq \tfrac{5}{3} l_{TSP}
\vspace{-.09in}
\end{equation}
\noindent where $l_{TSP}$ is the length of the shortest $S_1$-to-$S_n$ path  through sensing nodes located at the targets. As shown in Fig.~\ref{fig:RatioDiagram}, the geometry that leads to the {\em worst case ratio} between $ l_{TSP}$ and $l_{opt}$ occurs when the sensing circle are packed in alternating linear pattern (see appendix). In this packing, the length of the shortest $S_1$-to-$S_n$ path through the targets is
     \[ l_{TSP} \!=\! 2(n - 1)r \]
while the length of the shortest sensory coverage path is
\[
    l_{opt} = (n - 4 + \kappa'_o) r \rtxt{\hspace{1em} $n > 4$.}
    %= (n-5 + \bmod(n-1,2) + 2\sqrt{3}) r
\]
where $\kappa'_o$ is from Eq. (27) and (25) in the appendix.
%where $\bmod(n-1,2)$, which is shorthand notation for $\big[(n-1)\ modulo\ 2\big]$, always takes the value of $0$ when $n$ is odd.
It follows that the ratio $l_{TSP} / l_{opt}$ is bounded by
\[
\frac{l_{TSP}}{l_{opt}} \leq \frac{2(n - 1)}{n - 4 + \kappa'_o } \rtxt{\hspace{1em} $n > 4$.}
\]
Substituting for $l_{TSP} / l_{opt}$ in Eq.~\eqref{eq.53} gives
\begin{equation} \label{eq:ratio}
l(Alg.~\ref{alg:s_t_heuristic}) \leq \frac{10}{3} \cdot \frac{n \!-\! 1}{n - 4 + \kappa'_o } \cdot l_{opt} \rtxt{\hspace{1em} $n \geq 4$.}
\end{equation}
The term that multiplies $10/3$ in Eq.~\eqref{eq:ratio} can be written as
\[
\frac{n \!-\! 1}{n \!-\! 4 \!+\! \kappa'_o } \!=\! 1 \!+\! \delta(n) \hspace{.7em} \mbox{where} \hspace{.7em}
\delta(n)  \!=\! \frac{ 4 \!-\! \kappa'_o} { n \!-\! 4 \!+\! \kappa'_o}.
\]
Note that as $n \to \infty $, $\delta(n) \to 0$.~\epf
\vspace{.01in}
%can never be longer than the length of the approximate TSP solution found in
%Algorithm \ref{alg:hoogeveen}.  Likewise, the length of the path found by Algorithm
%\ref{alg:hoogeveen} is bounded by the approximation ratio bound of Lemma \ref{lemma:hoogeveen}:
%$\bar{L}_h\le (5/3)L^{*}_{TSP}$.  Substituting these inequalities into Eq. (\ref{eq:comp1}) yields.
%  \begin{equation} \label{eq:comp12}
%      \Rs_{app} \ \le \  \bigg(\frac{5}{3}\bigg) \frac{L_{TSP}^{*}}{\underbar{\em L}^{*}_{CPP}}
%  \end{equation}
%

%%-------------------------------------------------- %%
%% Multi-Targrt Views %%

\section{Fewer Sensing Nodes than Targets} \label{sec:fewer}
\vspace{-.02in}

\noindent When the targets are located close together ~in ~an ~\mbox{obstacle}-free environment, the covering robot might be able to take advantage of opportunities to view multiple targets from a single sensing location. Multi-target view opportunities can  simplify and shorten the sensory coverage path. This section formulates the {\small MINLP} optimization problem that allows multi-target views, then describes a polynomial time approximation algorithm, followed by sensing node reduction along the coverage path. The section concludes with analysis of the algorithm's path length approximation bound.
% in terms of the optimal path length.
%Let us first illustrate the potential of sensing node reduction.

%% ------------------------------ %%
\begin{figure}
 \centerline{\includegraphics[width=0.3\textwidth]{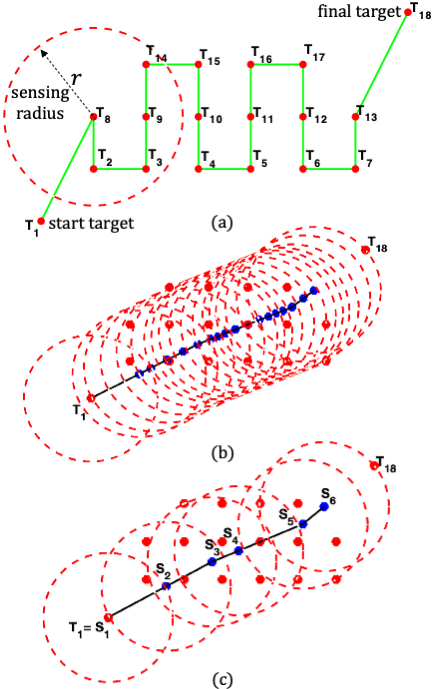}}
   \vspace{-.12in}
    \caption{(a) An $18$ target (red dots) coverage problem overlaid with a~dashed red circle showing the robot sensing range. (a)~The $S$-to-$T$  Hoogeveen  path
    %through the targets
    computed by Alg.~\ref{alg:hoogeveen}.  (b) The optimized sensing  node locations computed by  Alg.~\ref{alg:s_t_heuristic} (blue dots). (c) When redundant sensing nodes are removed, six sensing nodes can observe all targets.}
   \label{fig:overlap1}
%  \vskip -0.1 true in
\end{figure}
%% ------------------------------ %%

{\bf Motivational example:} Fig.~\ref{fig:overlap1}(a) shows the $18$ target example of Fig.~\ref{fig:ManyDisjointNodes}, now with
a~larger sensing radius of~\mbox{$r \!=\! 1.75$} units which allows individual sensing nodes to cover multiple targets.
The initial Hoogeven path computed by Alg.~\ref{alg:hoogeveen}~is shown in Fig.~\ref{fig:overlap1}(a), while
%Note that Alg. \ref{alg:hoogeveen} often produces {\em boustrophedan-like} \cite{choset_coverage_2000} paths
%when applied to problems with repeating geometries.
the optimized sensing node~loc\-ations computed by Alg. ~\ref{alg:s_t_heuristic} are shown in Fig.~\ref{fig:overlap1}(b).
The optimized sensing nodes are arranged along a nearly straight path, which is clearly the shortest path that observes all targets.
% starting with the Hoogeveen path of Fig.~\ref{fig:overlap1}(a).
When the optimized coverage path takes advantage of multi-target views, Fig.~\ref{fig:overlap1}(c) shows that six sensing nodes suffice to observe all targets. This example shows that path optimization can be performed under maximally flexible sensing node placement, followed by sensing node pruning, or reduction, along the optimized coverage path.~\eex

{\bf {\small MINLP} formulation:} Let $n_s \!\leq\! n$ denote the number of sensing nodes that now becomes an~additional
optimization variable.  To allow fewer sensing nodes than targets,
%the sensing constraints of Eq.~\eqref{eq:inrange} are modified as follows.  For
for each target $T_i$ %($i \!=\!1,\ldots,n$)
the sensing constraints of Eq.~\eqref{eq:quadratic}
%condition $||S_j \!-\! T_i|| \!\leq\! r^2$
must be satisfied by {\em at least one} sensing node $S_j$ ($j \!=\!1 \ldots n_s$).  This constraint can be expressed in disjunctive
form
\vspace{-.06in}
\[
\mbox{ $\bigvee$}_{j=1}^{n_s} \big( ||S_j \!-\! T_i||^2-r^2 \leq 0 \big) \rtxt{\hspace{.5em} $i=1\ldots n$.}
\vspace{-.06in}
\]
%Using a {\em Big-M} methodology \cite{vecchietti_modeling_2003},
\noindent The disjunctive constraints can be expressed as a set of mixed integer constraints using {\em viewing variables,} $\eta_{ij} \!\in\! \{0,1\}$, and {\em Big-M} constants $M_i $ for $i \!=\! 1\ldots n$~\cite{vecchietti_modeling_2003}. Each constant $M_i$ is
chosen sufficiently large according to the rule
  \begin{equation}
       M_i \geq \max_{x \in R(T_j), \, j = 1 \ldots n} \big\{ ||x - \mbox{\small $T$}_i||^2 \big\} \rtxt{\hspace{1em} $i \!=\! 1\ldots n$}
  \end{equation}
where $R(T_j)$ is the sensing circle of radius~$r$ centered at $T_j$.
%($j \!=\! 1 \ldots n$).
Using the Big-M constants,  each target $T_i$  for $i \!=\! 1\ldots n$ must be observed according to the {\em conjunctive} constraints
\begin{eqnarray}
    && ||S_j - T_i||^2 \, \leq \, \textcolor{black}{\eta_{ij}} \!\cdot\! r^2 + (1 \!-\! \eta_{ij}) \cdot M_i  \rtxt{$j=1 \ldots n_s$} \nonumber \\ %  \label{eq:bigM}\\
    && \sum_{j=1}^{n_s} \ \eta_{ij} \geq 1   \label{eq:atleastone} \\
    && \eta_{ij} \!\in\!  \{0,1\}  \rtxt{\hspace{1em} $ j=1 \ldots n_s$.} \nonumber %\label{eq:etadefn}
\end{eqnarray}
The viewing variables indicate sensing relationships: $\eta_{ij} \!=\! 1$ when $S_j  \!\in\! R(T_i)$ ~and ~$\eta_{ij} \!=\!0$ ~\mbox{otherwise.} Eq.~\eqref{eq:atleastone} ensures ~that at least one sensing node observes each target, with multi-views of the same target allowed.  To our knowledge, this formulation represents the first complete statement of the multi-view problem
with variable number of sensing nodes.

%The sensory coverage problem with multi-target views is as follows.
% (for each $i$) found in Eq.s
%(\ref{eq:bigM}), (\ref{eq:atleastone}), and (\ref{eq:etadefn}) below.  The constants $M_i$ must be
%chosen sufficiently large:
%  \begin{equation}
%       M_i \ge \max_j\{ ||S_j-T_i||^2\ |\ S_j^{lower} \le S_j \le S_j^{upper}\}
%  \end{equation}
%where $S_j^{lower}$ and $S_j^{upper}$ are lower and upper bounds on $S_j$.  The additional binary
%variables $\{\eta_{ij}\}$ indicate sensing relationships: $\eta_{ij}=1$ if $T_i\ \in \ R(S_j)$, else
%$\eta_{ij}=0$.

{\bf Problem \boldmath{$\# 2$}:}  Compute the shortest $S$-to-$T$ path $\Ps$ that visits $n_s \leq n$ variable position sensing nodes such that all targets $T_1,\ldots,T_n$ are observed at the sensing nodes. This problem is formulated as the {\small MINLP} optimization problem
\vspace{-.08in}
%\begin{equation}\label{eq:OptProblemFull}
\[
    (n^*_s,\mathcal{S}^*,\{\xi_{ij}^*\}, \{\eta_{ij}^*\})  \!=\!  \argmin_{n_s,\mathcal{S},\{\xi_{ij}\},\{\eta_{ij}\} }\
      \!\!   \bigg(\!    \sum_{j=i,j > i}^{n_s} \! \!\xi_{ij} \cdot d(S_i,S_j) \!\bigg)
\]
where $d(\mbox{\small $S$}_i,\mbox{\small $S$}_j) \!=\! \mbox{\small $\norm{S_i \!-\! S_j}$}$ is the Euclidean distance. The sensing nodes, $\mathcal{S} \!=\! \{S_1,\ldots,S_{n_s}\}$,  and incidence var\-iables, $\xi_{ij}$, are subject to the constraints of Eqs \eqref{eq:edge}-\eqref{eq:binary} while the sensing nodes and viewing variables, $\eta_{ij}$, must satisfy for each target $T_i$ ($i \!=\! 1\ldots n$) the multi-view constraints (\ref{eq:atleastone}).
%\begin{eqnarray}
%    && ||S_j - T_i||^2 \, \leq \, \textcolor{black}{\eta_{ij}} %\!\cdot\! r^2 + (1 \!-\! \eta_{ij}) \cdot M_i  \rtxt{$j=1 \ldots %n_s$} \nonumber \\
%    && \sum_{j=1}^{n_s} \ \eta_{ij} \geq 1,   \eta_{ij} \!\in\!  \%{0,1\}  \rtxt{\hspace{1em} $ j=1 \ldots n_s$.} \nonumber
%\end{eqnarray}

%------------------------------------------------------------------------------------%

%% ----------------------------------------------------------------------------------------------------------------------------- %%
\begin{figure}
%\mbox{ \hspace{-.5in} \centering{\includegraphics[width=.6\textwidth]{overlap18_hoog.png}}}
 \centering{
 \includegraphics[width=.39\textwidth]{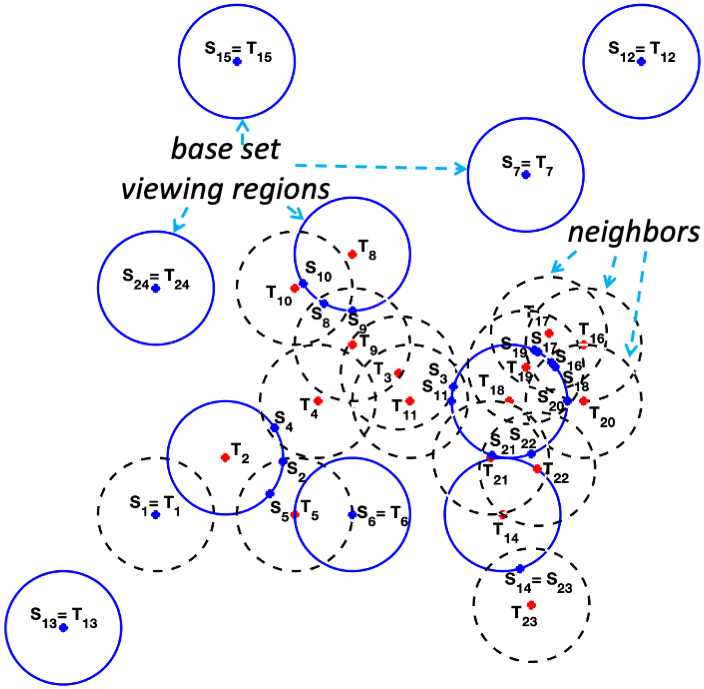}
 }
 \vspace{-.125in}
 \caption{A $24$ targets example with overlapping sensing circles.~The~base set circle
 indices are $\Is \!=\! \{ 2, 6, 7, 8, 12, 13, 14, 15, 18, 24 \}$ (blue circles). Initial sens\-ing nodes selected on the perimeter
 or center of the base set circles (blue dots).}
   \label{fig:base_set}
%  \vskip -0.1 true in
\end{figure}
%% ---------------------------------------------------------------------------------------------------------------------------------- %%

{\bf  Polynomial time 
%approximation 
algorithm for multi-target views:} The algorithm that approximately solves Problem~$\# 2$
first computes a~{\em base~set} of disjoint sensing circles, such that all targets can be observed from sensing nodes in the base set sensing circles. Consider the collection of 
%overlapping 
sensing circles centered at the targets, $\Ws \!=\! \{ R(T_i) : i=1\ldots n \}$,  each having radius $r$. % which is the robot detection range.
%\textcolor{blue}{The first base set circle is $R(T_1)$, centered at the robot start point. This circle and its overlapping circles are removed~from~$\Ws$.
All isolated sensing circles are first selected as base set circles and removed from~$\Ws$.
In each subsequent iteration, one randomly selects a sensing circle from $\Ws$,
%the circle  that overlaps the {\em highest} number of remaining circles in $\Ws$,
then removes this circle and its overlapping circles from $\Ws$. This iteration repeats until all sensing circles have been removed from $\Ws$.
The base set circles are computed in $O(n)$
%$O(n^2\log n)$
time and their indices are stored in the set $\Is$.\\
\indent {\bf Example:} Fig.~\ref{fig:base_set} illustrates the selection of  base set circles on a~$24$ target example.
% shows anwith overlapping sensing circles (dashed circles).
All isolated sensing circles
%at $T_7,T_{12},T_{13},T_{15},T_{24}$
are first selected as base set circles and removed from~$\Ws$.  Next $R(T_8)$ is randomly selected as a~base set circle and removed with its overlapping circles %at $T_9,T_{10}$
from~$\Ws$. This step repeats for $R(T_2)$, $R(T_{18})$ and $R(T_{14})$.  Finally $R(T_6)$  is left in~$\Ws$ and  selected as the last base set circle.~\eex

\indent {\bf Initial Sensing node location selection:} Each base set circle, $R(T_i)$ for $i \!\in\! \Is$,  is used to select initial sensing node locations as follows. If $R(T_i)$ has no neighbors, no other target's sensing circles overlap $R(T_i)$, the target $T_i$ at its center is selected as an initial ~sensing node location. These are sensing nodes $S_6, S_7, S_{12}, S_{13}, S_{15}, S_{24}$ in Fig.~\ref{fig:base_set}. Otherwise, $R(T_i)$ overlaps other sensing circles and sensing nodes are selected on $R(T_i)$'s perimeter at the midpoint of its overlap with each neighboring circle. These are $S_4,S_5$ on the perimeter of $R(T_2)$,
$S_9,S_{10}$ on the perimeter of $R(T_8)$, several sensing nodes on the perimeter of $R(T_{18})$ and $S_{23}$ on the perimeter of $R(T_{14})$ in Fig.~\ref{fig:base_set}.  The sensing node for each non-isolated base set circle is selected on its perimeter, at the center of its overlaps with the neighboring sensing circles. These are the~\mbox{perimeter}~of~$R(T_2),R(T_8),R(T_{14}),R(T_{18})$ in Fig.~\ref{fig:base_set}. The initial sensing nodes are computed in $O(n)$ time and $n_s \!=\! n$ at this stage.

%\indent {\bf Hoogeveen  \boldmath{$S$}-to-%\boldmath{$T$} %shortest 
{\bf Sensing nodes optimization:} Having identified ~\mbox{initial} ~sen\-sing node locations,  the algorithm that solves Problem~$\# 2$ proceeds much like the solution to Problem~$\# 1$. Algorithm~\ref{alg:hoogeveen} computes a~Hoogeveen path that starts at $S_1 \!=\! T_1$, visits the intermediate sensing nodes in any order and ends at a sensing node that observes the final target $T_n$. The Hoogeven path fixes the incidence variables, $\xi_{ij}$, and the sensing nodes are indexed according to the Hoogeven path visitation order as $S_1,\ldots,S_n$. The  sum-of-norms cost function
\vspace{-.08in}
   \begin{equation}\label{eq:min_Scost}
       Cost(S_1,\ldots,S_n) \ =  \ \sum_{i=1}^n  \norm{S_{i+1} - S_i }
\vspace{-.08in}
\end{equation}
\noindent forms a~convex function of the sensing node locations. Since each target has its own sensing node at this stage,
the viewing variables
%of each target $T_i$
are set as $\eta_{ij} \!=\! 1$ for $j \!=\! i$ and $\eta_{ij} \!=\! 0$ for $j \! \neq \! i$ ($i \!=\! 1\ldots n$).
Each sensing node $S_j$ except $S_1$ thus may vary in its sensing circle, $R(T_j)$. Computation of the shortest sensory coverage path
%for the reduced set of sensing nodes
under fixed visitation order and fixed viewing assignments forms a~convex optimization problem that can be efficiently solved using Algorithm~\ref{alg:s_t_heuristic} of the previous section.

%for the shortest sensory coverage path.
% under the fixed visitation order.

%% --------------------------------------------------%%
\begin{figure}
%\mbox{ \hspace{-.5in} \centering{\includegraphics[width=.6\textwidth]{overlap18_hoog.png}}}
 \centering{
 \includegraphics[width=.45\textwidth]{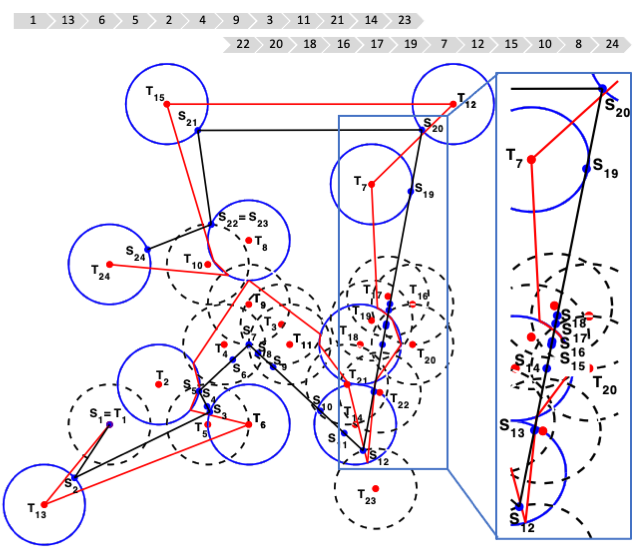}
 }
 \vspace{-.14in}
\caption{The $24$ targets example of Fig.~\ref{fig:base_set}.
 The Hoogeveen path through the initial sensing nodes shown as red path, ordered by gray incidence arrows above. The optimized sensory coverage path computed under the Hoogeveen visitation order shown as black path, with
 the optimized sensing nodes overlaid as blue dots along this path.}
    \label{fig:overlap_hoog}
%  \vskip -0.1 true in
\end{figure}
%% --------------------------------------------------%%

\indent {\bf Example:} Fig.~\ref{fig:overlap_hoog} illustrates the two-stage solution~of~Alg\-orithm~\ref{alg:s_t_heuristic} on the~$24$ targets example of Fig.~\ref{fig:base_set}. The Hoogeveen path passing through the initial sensing nodes was computed in $0.4$~seconds (red path). The sensing node locations are next optimized under the Hoogeveen visitation order as a~convex optimization problem, with each sensing node allowed to freely vary in its target-centered sensing circle. The  optimized sensory coverage path shown as black path in Fig.~\ref{fig:overlap_hoog} was computed in $1.11$~seconds. The $24$ optimized sensing nodes appear as blue dots along the optimized path.~\eex

%% --------------------------------------------------%%
\begin{figure}
%\mbox{ \hspace{-.5in} \centering{\includegraphics[width=.6\textwidth]{overlap18_hoog.png}}}
 \centering{
 \includegraphics[width=.5\textwidth]{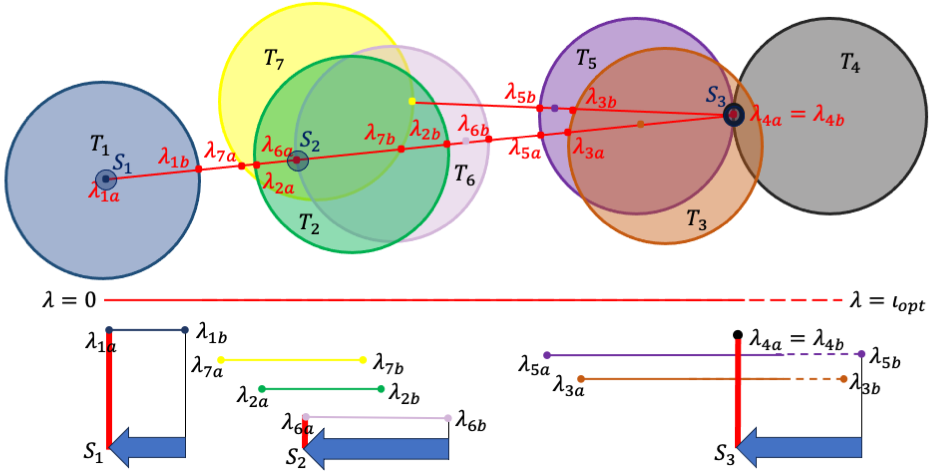}
 }
 \vspace{-.14in}
\caption{The sensing node reduction in Alg. \ref{alg:node_reduction} is illustrated, with intervals identified, sorted, reversed along to place sensing nodes.}
    \label{fig:sensredux}
%  \vskip -0.1 true in
\end{figure}
%% --------------------------------------------------%%

{\bf Sensing node reduction:} The final stage of the algorithm 
%for Problem~$\# 2$
%(summarized as Alg.~\ref{alg:overlap_reduce})
selects a~reduced set of sensing nodes along the optimized sensory coverage path.  General sensing node reduction in {\small 2-D} environments forms an~{\small $NP$}-hard {\em set cover} problem~\cite{approx_alg}.
However,  sensing node reduction along the optimized path can be done in polynomial time as next described.

The {\em overlap segments} on the optimized path, denoted as $I_i \!=\! [a_i,b_i]$ for $i \!=\! 1\ldots k$ ($k \!\geq\! n)$, are defined as the intervals along the path where each sensing circle is first entered, at $a_i$ and exited, at $b_i$.
%where $a_i$ and $b_i$ are values of the path length %parameter $s$.
%which ranges from $s \!=\! 0$ to 
%$l(\mbox{\rm Alg.~\ref{alg:s_t_heuristic}})$. 
%When the optimized path intersects 
%a~sensing circle multiple times, 
%as we require one $\lambda$-interval per sensing %circle. 
%Selecting the \textit{first} intersection along a path %would allow for the most information quickly, %desirable if there is live relay of target %information. 
%we select the first overlap segment.
%along the optimized path. 
The overlap segments are thus uniquely associated with the sensing circles along the optimized path. Sensing node reduction starts by sorting the overlap segments along the path into list $\Us$. Moving in reverse from the end of the optimized path until the first $a_i$ entry is found, a~sensing node is placed at $a_i$, allowing the robot to observe $T_i$ and possibly  many targets along the $[a_i,b_i]$ segment, since other overlap segments have only begun in the region of $a_i$, but none has ended. This is illustrated in Fig. \ref{fig:sensredux}, with $S_3$ as the first node. All targets viewable from $a_i$, $T_j$ s.t. $a_j \leq a_i \leq b_j$, are removed from the list of uncovered target intervals $\Us$.  This process repeats until all targets have been observed, shown in Fig. \ref{fig:sensredux} as $S_1, S_2,$ and $S_3$.

%of the  or randomly select overlaps, regardless, %\textit{the method will return the minimum number of %sensing nodes required to observe all targets along %the given path subject to that order}.

%segment and we assigned one that observes all %possible other targets along this segment,  %Thus, 
%the fewest sensing nodes were selected for this %segment and we know 
%the single 
%must exist along this segment, and 

The sensing node reduction process is summarized as Algorithm ~\ref{alg:node_reduction}. Lines~\mbox{3-6} compute the overlap segments, which are stored in $\Us$. Line~7 sorts these segments by their start point, $a_i$ for $i \!=\! 1 \ldots n$. Lines~8-19 select sensing nodes by reversing along $\Ps$ and removing redundant overlap segments from~$\Us$, while storing the selected sensing nodes in $S_{min}$. The path $\Ps$ consists of $n \!-\! 1$ line segments checked against $n$ sensing circles to identify  overlap segments. Computation and sorting of the overlap segments thus takes $O(n^2)$ time, then sensing node reduction takes additional $O(n)$ time.

%The overlap intervals and sensing node reduction can %thus be computed in $O(n^2)$ time.

%has been used, subject to the segment's identified for %use. For the sake of contradiction, if a set of %sensing nodes with fewer nodes fully observed all %targets, there must be overlaps in the key domains %used to place sensing nodes, because every domain used %required a sensing node somewhere from $a_i-b_i$. %However, every time a sensing node is placed, all %possible overlapping intervals are removed, and the %next key domain thus must be $\epsilon>0$ away from %the first. Thus, this sensing node set has the fewest %possible elements, given the conditions.

%The algorithm
%Algorithm~\ref{alg:node_reduction}
%computes in $O(n^2)$ time a~reduced set of sensing nodes that observe all targets.}
%along a~given optimized sensory coverage path.}

%%% ------------------------------------------ %%%
\begin{algorithm} %[H]  %[ht]
\caption{Sensing Node Reduction}
 \begin{small}
%\textcolor{blue}{
\hspace{.2em} {\bf Input:} $\!$ Targets \mbox{\small $\Ts \!=\! \{T_1\ldots T_n\}$}.~Optimized~path~$\mathcal{P}$.~Sensing \\
\mbox{\hspace{.29em}} nodes  \mbox{\small $\Ss \!=\! \{S_1\ldots S_n\}$} along $\mathcal{P}$.
%}
\begin{algorithmic}[1]
\Procedure{SensingNodeReduction}{$\Ts$,  $\mathcal{P}$, $\Ss$} \label{line:reduced}
%\textcolor{blue}{
\State {$\Us \gets \emptyset$, \ \ $\Ss_{min}\gets\emptyset$}
 \For{$i=1$ to $n$}     \label{line:findcover_start}
    \State {$I_i \gets OverlapInterval( R(T_i), \mathcal{P})$}
    \State {$\Us \gets \Us \cup \{I_i\}$}
 \EndFor
% \For{$i=1$ to $n$}
%    \If{$I_i = \{S_i\}$ }
%    \State {$\Ss_{min} \gets \Ss_{min} \cup \{S_i\}$}
%    \State {Remove $I_i$  from $\Us$}
%    \State {Remove all $I_j$  s.t. $Covers(S_i,T_j)$ from $\Us$}
%       \EndIf
% \EndFor                    \label{line:findcover_end}
%   \State {If $\Us \!\neq\! \emptyset$, re-index~overlap~intervals~in~$\Us$~as~$I_1,\ldots,I_k$}
%    \State {Identify disjoint path segments of $\cup_{i=1}^k I_i$ as $\Pi_1,\ldots,\Pi_l$}
%    \For{$i=1$ to $l$}
   \State {Set $\Us = SortOverlapSegments(\Us)$}
 \While{$\Us \!\neq\! \emptyset$} \label{line:while}
       \State{$\mathrm{S}_{new} = \Us(end)(a_i)$}
       \State{$\Ss_{min} = \Ss_{min} \cup \{\mathrm{S}_{new}\} $}
       \State{Set $j=1$}
        \While{$j\leq size(\Us)$}   
            \If{$\mathrm{S}_{new} \in I_j$}
            \State {Discard $I_j$ from $\Us$}
            \State{$j--$}
            \EndIf
            \State{$j++$}
        \EndWhile
 \EndWhile  \label{line:end_while}
%     \EndFor
%   \EndFor
%   }
   \State {Return \big(reduced sensing nodes  $\Ss_{min}$ \big)}
\EndProcedure
\end{algorithmic}
\end{small}
\label{alg:node_reduction}
 \end{algorithm}

\begin{figure}
\centering{
\includegraphics[width=.5\textwidth]{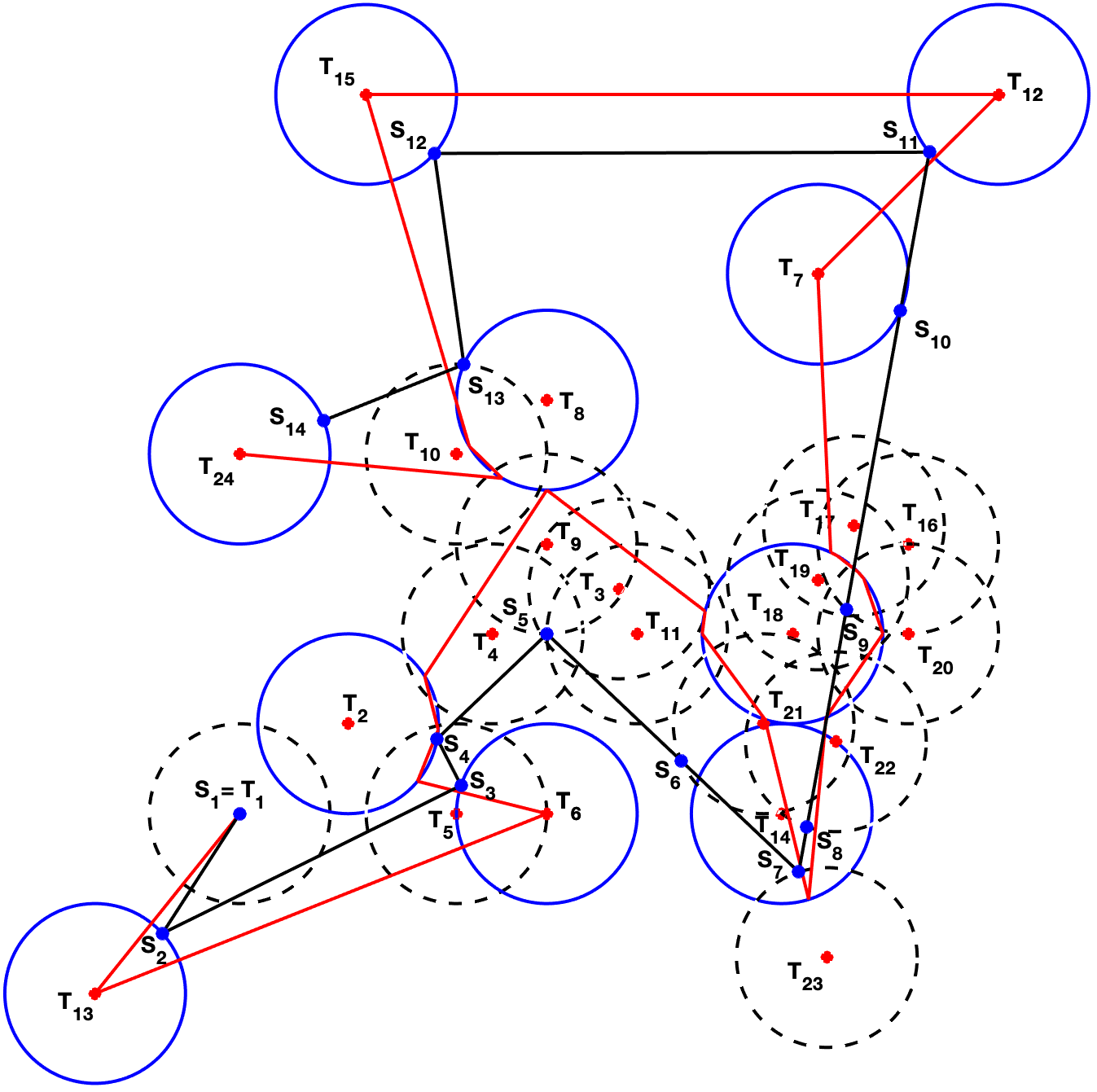}}
 \vspace{-.2in}
 \caption{The $24$ targets example of Fig.~\ref{fig:overlap_hoog} showing
 the optimized sensory coverage path (in black) with reduced sensing nodes  $S_1,\ldots,S_{14}$ that use  multi-target views to observe all  targets
along this path.}
   \label{fig:reduce_opt}
  \vskip -0.1 true in
\end{figure}

{\bf Example:} Algorithm~\ref{alg:node_reduction} was applied to the $24$ sensing nodes located on the optimized sensory coverage path of Fig.~\ref{fig:overlap_hoog}. The algorithm found the reduced sensing nodes $S_1,\ldots,S_{14}$ shown in Fig. ~\ref{fig:reduce_opt} in $2.0$~seconds. These sensing nodes exploit multi-target views to observe all $24$ targets.~\eex 
%along the optimized path.~\eex
%\vspace{.01in}

%%% -------------------------------------------------------------- %%%
%\vskip -0.15 true in
\begin{algorithm}[ht]
%\textcolor{blue}{
\caption{Shortest Cover Under Multi-Target Views}
\hspace{.2em} {\bf Input:} $\!$ Targets \mbox{\small $\Ts \!=\! \{T_1,\ldots ,T_n\}$}
%~\mbox{Sensing}~\mbox{nodes}~\mbox{\small $\Ss \!=\! \{S_1\ldots S_m\}$}
\begin{small}
\begin{algorithmic}[1]
\Procedure{ReducedSensingPath}{$\Ts$}
\State { $\Is = BaseSetCicrcles(\Ts)$ }
\State { $\{ S_1, \ldots, S_n \} = BaseSetSensingNodes(\Is, \Ts)$ }
%\State { Set $\Ss = \{ S_1, \ldots, S_n \}$ }
\State { $\mathcal{P}_{order}$\ =\ {\rm Hoogeveen}($S_1, \ldots, S_n$) }
\State { $\{\xi_{ij}\}\ =\ {\rm Incidence}(P_{order})$ }
\State { $\eta_{ij} \!=\! 1 \ {\rm for}  \ j \!=\! i, \eta_{ij} \!=\! 0  \ {\rm for} \ j \! \neq \! i$ }
\State { ($\Ps, \Ss_{opt} ) = \argmin {\rm Cost}\big\{ \{\xi_{ij}\},\{\eta_{ij}\},\Ss,\mathcal{T} \big\}$ } \label{line:convex}
\State { $\quad  {\rm subject \  to}  \ S_1 \!=\! T_1 \ {\rm and} \ S_i \!\in\! R(T_i)$  $i =2 \ldots n$ }
\State { $\Ss_{min}$ = SensingNodeReduction($\Ts$, $\Ps$, $\Ss_{opt}$) }
%\State{Set $n_s = |\Ss_{min}|$, \ $S_1 = T_1$}
\State { $Return\big( \mbox{\rm path} \ \Ps, \mbox{\rm reduced sensing nodes} \ \Ss_{min} \big)$ }
\EndProcedure
\end{algorithmic}
\end{small}
\label{alg:overlap_reduce}
%}
\end{algorithm}
\vspace{-0.05 true in}
%%% -------------------------------------------------------------- %%%

%% ----------------------------------------------------------------------------------------------------------------------------- %%
%% ---------------------------------------------------------------------------------------------------------------------------------- %%

%{\bf Remark:} Further sensing nodes reduction and shortening of the sensory coverage path can be achieved
%by iteratively running Algorithms~\ref{alg:node_reduction}~and~\ref{alg:overlap_reduce}.
%%The optimized sensing node positions, $\Ss_{opt}$,
%%have changed as a~result of the convex optimization performed in line~\ref{line:convex} of Algorithm~\ref{alg:overlap_reduce}.
%In each iteration one runs Algorithm~\ref{alg:node_reduction} on the optimized nodes of $\Ss_{opt}$ for sensing nodes reduction. If the number of sensing nodes
%has been reduced,  one runs Algorithm~\ref{alg:overlap_reduce} to optimize the positions of the smaller number of sensing nodes, until no more sensing node reduction is achieved.~\eex
%\vspace{.02in}

The approximation algorithm that solves Problem~$\# 2$ is summarized as Algorithm ~\ref{alg:overlap_reduce}. Its main steps are base set selection in $O(n)$ time, initial sensing nodes assignment in $O(n)$ time, Hoogeveen path computation in $O(n^3)$ time, convex optimization in $O(c \cdot n^2 \ln(1/\epsilon))$ time, and sensing node reduction in $O(n^2)$ time, $n$ is the number of targets.

%--------------------------------------------------%
{\bf  Path length bound:} Proposition \ref{prop:bound2} describes the path length approximation bound for Algorithm~\ref{alg:overlap_reduce}.  The proof follows Dumitrescu ~\cite{dumitrescu_approximation_2003} with a~slightly improved bound.
%\vspace{.01in}

\begin{prop} \label{prop:bound2}
%Consider $n>4$
When the locations of targets $T_1,\ldots,T_n$ leads to overlapping sensing circles of radius~$r$, Algorithm~\ref{alg:overlap_reduce} computes in $O(n^3)$ time a~sensory coverage path whose length satisfies the approximation bound
\vspace{-.08in}
\[
     l(\mbox{\rm Alg.~\ref{alg:overlap_reduce}})  \ \leq \ \textcolor{black}{\tfrac{5}{3}} 
     \big( 9 \cdot  l_{opt} \!+\! (1 \!+\! 8 \pi ) r \big)
 \vspace{-.06in}
 \]
\noindent where $l_{opt}$ is the length of the optimal $S$-to-$T$ sensory coverage path that takes advantage of multi-target views
in an~obstacle free environment.
%Moreover, $|\mathcal{P}_h|$ is computed in time proportional $\mathcal{O}(Bn^3)$, where $B$ is the
%desired number of bits of accuracy in the sensing node location.
\end{prop}

%\vskip 0.07 true in
\noindent {\bf Proof:}
Let $\Ag_{opt}$ be the shortest $S$-to-$T$ sensory~\mbox{coverage}~path that takes advantage of multi-target views, with $l_{opt}$ denoting its length. The optimal path must visit every~base~set circle, otherwise the targets at the centers~of~these~circles~can\-not be observed by the robot. Consider the area swept by a~disc of radius $2r$ while its center moves along the path $\Ag_{opt}$
\vspace{-.06in}
\begin{equation} \label{eq.area}
\As = 4r \cdot l_{opt} + 4 \pi r^2
\vspace{-.06in}
\end{equation}
\noindent where $4 \pi r^2$ is the area of two half-discs at the endpoints of~$\Ag_{opt}$. The area swept by the disc 
%of radius $2r$ 
covers all base set circles, each having radius~$r$. Since the base set circles are disjoint
\vspace{-.04in}
\[
\As \geq |\Is| \cdot (\pi r^2)
\vspace{-.04in}
\]
\noindent where $|\Is|$ is the number of base set circles. Substituting for $\As$ according to Eq.~\eqref{eq.area} gives
\vspace{-.06in}
\begin{equation} \label{eq.m}
 |\Is|  \leq 4 ( \frac{l_{opt}}{\pi r} + 1 ).
\vspace{-.02in}
\end{equation}
\noindent Next consider the 
%$S$-to-$T$ Hoogeveen 
path computed by Alg.~\ref{alg:overlap_reduce}. It starts at~$S_1$, visits  $S_2,\ldots,S_{n-1}$ in some order, then ends at $S_n$.
%the sensing node that observes the final target~$T_n$. 
Using Lemma~\ref{lemma:hoogeveen}, the length of this path, $l_H$, satisfies the bound $l_H \leq \tfrac{5}{3} l_{TSP}$ where $l_{TSP}$ is the length of the shortest $S$-to-$T$ path that starts at~$S_1$
%, visits $S_2,\ldots,S_{n-1}$ in any order 
and ends at~$S_n$.  The convex optimization stage of Alg.~\ref{alg:overlap_reduce} only {\em shortens} the Hoogeveen path. Hence
$l(\mbox{\rm Alg.~\ref{alg:overlap_reduce}}) \leq l_H \leq \frac{5}{3} l_{TSP}$.

An alternative path $\mathcal{P}$ that visits the sensing nodes takes advantage of the nodes' initial locations on the base set circle perimeters and centers.  By construction, each base set circle has sensing nodes on its perimeter or a~single sensing node at its center.  Path $\mathcal{P}$ starts at $S_1$ and follows $\Ag_{opt}$. Each time $\Ag_{opt}$ reaches a base set circle, imagine that the robot's path executes full circumnavigation of the circle when it contains perimeter sensing nodes, and straight in-and-out motion to its center when the circle contains a~sensing node at its center. The path then continues along $\Ag_{opt}$ to the next base set circle. The length of this path, $l_b$, satisfies the upper bound
\vspace{-.04in}
\[
l_b \leq l_{opt} + 2\pi r \cdot |\Is| + r.
\vspace{-.04in}
\]
\noindent The path that circumnavigates the base-set-circles is {\em longer} than the shortest path that visits the sensing nodes. Hence $l_{TSP} \leq l_b$ and consequently
\[
\begin{array}{l}
l(\mbox{\rm Alg.~\ref{alg:overlap_reduce}})
\! \leq \! \tfrac{5}{3}  l_{TSP} \! \leq \! \tfrac{5}{3} l_b \\[2pt]
%\hspace{-4.5em} 
\! \leq \! \tfrac{5}{3} \big( l_{opt} \!+\!  8 \pi r  \cdot  ( \frac{l_{opt}}{\pi r} \!+\! 1 ) + r \big)  \!=\!   \tfrac{5}{3} ( 9 \cdot  l_{opt} \!+\! (1 \!+\! 8 \pi ) r )
\end{array}
\]
where we substituted the upper bound~\eqref{eq.m} for $|\Is|$.~\epf
%The convex optimization stage of
%Since Algorithm~\ref{alg:overlap_reduce} only shortens the Hoogeveen path. Hence $l(\mbox{\rm Alg.~\ref{alg:overlap_reduce}}) \leq l_H \leq \frac{5}{3} %l_{TSP}$ which gives the bound $l(\mbox{\rm Alg.~\ref{alg:overlap_reduce}}) \leq \tfrac{5}{3} ( 9 \cdot  l_{opt} + 8 \pi r  )$.

%\vspace{.02in}

%%------------------------------------------------- %%
\section{ Multi-Target Sensory Coverage Among Obstacles with Path Length Bound} \label{sec:obstacles}

\vspace{-.02in}
\noindent This section considers multi-target sensory coverage in the presence of obstacles. We first develop an {\small MINLP} problem formulation, then describe an algorithm that approximately solves this problem, followed by sensing node reduction along the sensory coverage path. The section ends with analysis of the algorithm's path length approximation bound.

%------------------------------------------------%
\begin{figure}
  \centering{\includegraphics[width=.5\textwidth]{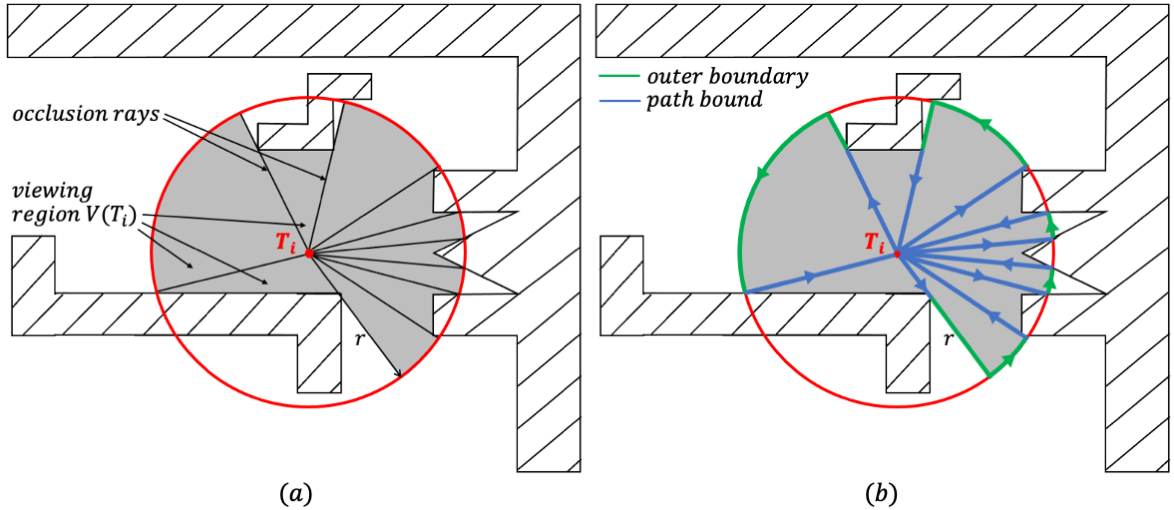}}
    \vspace{-.28in}
\caption{(a) The viewing region $V(T_i)$ of direct~lines~of~sight~that~start at $T_i$ and reach the sensing circle boundary or hit an~obstacle before reaching maximal detection range. Occlusion edges
are tangent to obstacles or reach sensing circle and obstacle intersection points. 
(b) The boundary that separates the viewing region $V(T_i)$ from other obstacle free regions, in green, consists of occlusion edge segments and sensing circle arcs.}
%are used to place the initial sensing nodes (see %text).}
    %The blue path connects these areas with a bounded %perimeter length.
    %(see text).}
\label{fig_view_cell}
\end{figure}
%------------------------------------------------%

The targets $T_1,\ldots,T_n$ lie in an~environment populated by polygonal obstacles that may obstruct sensor views of the targets.
The robot can detect targets only along direct lines of sight within detection range~$r$. Each target $T_i$ can thus be detected from a~{\em viewing region,} $V(T_i)$, defined as the set of rays that start at $T_i$ and either reach the sensor range limit or hit an~obstacle
%before reaching maximal detection range 
(Fig.~\ref{fig_view_cell}(a)). Each viewing region consists of triangular radial sectors
%Fig.~\ref{fig_view_cell}(a)). 
%Each radial sector is 
bounded by {\em occlusion edges.} Each occlusion edge is either tangent to 
an~obstacle vertex or edge, or reaches an~intersection point of the sensing circle boundary with an obstacle edge (Fig.~\ref{fig_view_cell}(a)). 

%Note each viewing region boundary can be fully %represented by a set of angles, $\theta_i$ where the %viewing region extends to a circular arc, and line %segments, enumerated by their boundary points in a set %$ls_i$. 

%The distal boundary consists of obstacle edges as well as circular arcs of radius %$r$ centered at $T_i$. % \textcolor{red}{(Fig.~\ref{fig_view_cell}(b))}.
In mobile robot sensory coverage, the presence of obstacles implies that  targets which can be observed by the robot may not be physically accessible by the robot. This issue is {\em not} directly included in our algorithm. Rather, the problem is viewed as optimizing sensing node placements and planning a safe collision-free path between them. We assume that the entire free space can be continually accessed by the robot sensor, and viewing regions are constructed accordingly. If desired, obstacles may be replaced by their configuration space representations.

%\textcolor{red}{We also assume that each viewing %region~$V(T_i)$
%%target~$T_i$ 
%contains an~obstacle free disc of radius~$\rho_i$.}

%(finite size mobile robots are discussed at the end of this section). 
%The {\small MINLP} formulation of multi-target sensory coverage among %obstacles is as follows.

{\bf {\small MINLP} formulation:} The robot path cost is measured as the sum  of the shortest collision free path lengths connecting successive sensing nodes along the robot path
\vspace{-.02in}
%   \begin{equation}\label{eq:min_Scost}
\[
       Cost(\{S_1,\ldots,S_n\}, \{\xi_{ij}\} )  =  \sum_{i=1}^n \sum_{j=1,j > i}^n \!   \xi_{ij} \! \cdot \! l_{min}( S_j, S_{j+1})  
\vspace{-.02in}
\]
where $l_{min}(S_j, S_{j+1}) \!\geq\! 0$ is the length of the shortest collision free path connecting $S_j$ and $S_{j+1}$. The incidence variables, $\xi_{ij}$ for $i,j \!\in\! \{ 1,\ldots,n \}$,~are~defined as before, $\xi_{ij}=1$ if $T_j$ is observed right after $T_i$, and 0 otherwise.

%\[
%\xi_{ij} \!=\! \xi_{ji} \!=\! \begin{cases} 1 \! &
%                  \mbox{\rm $\!\! T_j$~is~observed~immediately~after~or~before~$\! T_i$ } \\
%                              0 \! & $\!\!$ {\rm otherwise} \end{cases}
%\vspace{-.06in}
%\]
%We assume symmetric travel cost, so that $j \!>\! i$ in %the travel cost expression. 
%In order to present a~compact 
The {\small MINLP} formulation assumes
one sensing node per target, without Big-M constants that allow
multi-target views.
%in the {\small MINLP} formulation. 
However, sensing node reduction will be performed along the optimized sensory coverage path among obstacles. Each sensing node, $S_i \!\in \real^2$, must lie inside the viewing region $V(T_i)$.  This requirement is captured by two constraints 
  \begin{equation} \label{eq.vregion}
     ||S_i - T_i|| \leq r \hspace{.6 em}  \mathrm{and} \hspace{.6 em} l_{min}(S_i, T_i ) = ||S_i - T_i|| 
  \end{equation}
where $l_{min}(S_i, T_i )$ is the shortest collision free path~bet\-ween $S_i$ and~$T_i$. The inequality $\mbox{\small  $||S_i \!-\! T_i||$}  \!\leq\! r$ ensures that $S_i$ lies inside 
%the sensing circle 
$R(T_i)$, while the equality $l_{min}(S_i, T_i ) \!=\! \mbox{\small  $||S_i - T_i||$}$
ensures that  $S_i$ has a~direct line of sight to target $T_i$.
%for $i \!=1\! \ldots n$.

{\bf Problem \boldmath{$\# 3$}:} Given targets $T_1,\ldots,T_n$ in an~environment populated by polygonal obstacles, compute the shortest $S$-to-$T$ path, $\Ps$, that visits sensing nodes $S_1,\ldots, S_n$ such that all targets can be observed from the sensing nodes along direct lines of sight.  This problem is formulated as the {\small MINLP} optimization problem
% \sum_{i=1}^n \xi_{ij}\cdot d(S_1,S_i)
\vspace{-.08in}
%\begin{equation}\label{eq:OptProblemFull}
\[
    (\mathcal{S}^* , \{\xi_{ij}^* \})  \!=\!  \argmin_{ \mathcal{S}, \{\xi_{ij}\}, }\
      \!\!   \bigg(\!    \sum_{j=i,j > i}^{n} \! \!\xi_{ij} \cdot l_{min}(S_i,S_j) \!\bigg)
\]
where the sensing nodes, $\mathcal{S} \!=\! \{S_1,\ldots,S_n \}$,  and incidence variables, $\xi_{ij}$, are subject to the constraints of Eqs \eqref{eq:edge}-\eqref{eq:binary} while each sensing node $S_i$  must satisfy the viewing constraints of Eq.~\eqref{eq.vregion} for $i=1\ldots n$.
%in the polygonal environment.

%for each target $T_i$  
%the visibility constraints
%\[
%||S_i - T_i|| \leq r \hspace{.5 em}  \mathrm{and} \hspace{.5 em} %l_{min}(S_i, T_i ) \leq ||S_i - T_i|| \hspace{.8em} \mbox{$i=1\ldots %n$} \]
%where $l_{min}(\mbox{\small $S$}_i,\mbox{\small $S$}_j)$ 
%and $l_{min}(S_i, T_i )$ are shortest collision free path lengths 

{\bf  Approximation algorithm for multi-target views among obstacles:} 
The algorithm that approximately solves Problem~$\# 3$ selects initial sensing nodes, finds a sensor node visitation order, optimizes the ordered sensing node locations, then reduces the number of sensing nodes using multi-target views along the optimized path.

The algorithm first selects a~{\em base set} of disjoint viewing regions
from which all targets can be observed along direct lines of sight. The technique is similar to the one used in the previous section. Consider the set of viewing regions, $\Ws \!=\! \{ V(T_i) : i=1\ldots n \}$. One randomly selects a viewing region from $\Ws$, then removes this region and its overlaps with other viewing regions from $\Ws$. This iteration repeats until all viewing regions are removed from $\Ws$.

%The indices of the base set viewing regions %are stored in the set $\Is$.

Computation of the $n$-target viewing regions requires \mbox{$O(n \!\cdot\! m^2_r \log m_r)$} time, where $m_r \leq m$ is the maximum number of obstacle vertices and edges within detection range $r$ of each target, after an \mbox{$O(n \!\cdot\! m)$} pre-processing step to identify the set of intersecting vertices and edges of cardinality $m_{r}$.
%Computation of the overlap between a~base set viewing %region and all remaining viewing regions repeats %$O(n^2)$ times for $n$ targets. 
%To quantify the overlap computation time for two 
%viewing regions,  
Each viewing region boundary consists of at most $m_r$ occlusion edges and circular arcs (due to sensor range limits). The existence of an overlap between two viewing regions can be found in $O(m_r  \log m_r)$ time using the sweepline algorithm~\cite{4567905}.  The entire base set can thus be computed in  $O( n^2 \!\cdot\! m_r  \log m_r )$ time. The indices of the base set viewing regions are stored in the set $\Is$.
 
%The base set  are computed in $O(n)$
%$O(n^2\log n)$ time \\

%A base set of non-overlapping viewing regions is selected randomly. A viewing region, $V_j$ is selected to join the base set. All $n-1$ or fewer remain viewing regions are checked for overlaps, which per potential neighbor takes, at maximum, $O(m \cdot log(m))$. This can be validated as the overlapping algorithm requires checking two circles for an overlap, and determining if the potential overlap is in the range of $\theta_i$ and $\theta_i$. Next, all occlusion rays, of which there are maximum $m$ from the neighbor must be checked against the base circle for intersections in range of $\theta_j$. This is repeated with the base occlusion rays and neighboring circle. Finally, all occlusion rays must be checked for overlaps. Checking a set of $2m$ lines for overlaps can be accomplished used a sweepline algorithm in $O(m \cdot log(m))$. Therefore  making the overall base selection and node placement process $O(m \cdot log(m) \cdot n^2)$.  The final condition is a neighboring target within the potential bse viewing region. If $||T_j - T_i|| \leq r$, the line segment $[T_j  T_i]$ is checked for intersections with line segments in $p_i$. S If there are no intersections, the target is within the viewing region. If the base region has any occlusion rays, the sensing node is placed at the base target. Else, the sensing node is placed where the extended ray from $T_i$ to $T_j$ intersects the boundary.

%----------------------------------------------------------%
\begin{figure}
%\vspace{-.05in}
%\centering \includegraphics[width=.5\textwidth]{fig11v3.png}
\centering \includegraphics[width=.5\textwidth]{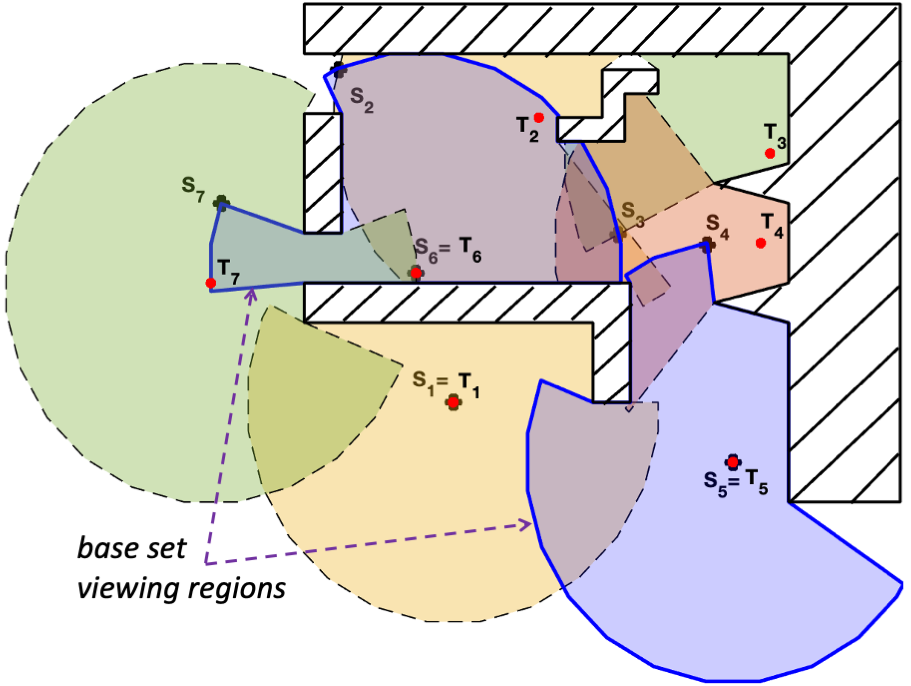}
\vspace{-.285in}
\caption{A seven target example showing a~base set of viewing regions $V(T_5),V(T_6)$. Initial sensing node locations $S_1,\ldots,S_7$ are selected on the boundaries and centers of the base set viewing regions (see text).}
\label{fig:base_obs}
\end{figure}
%-------------------------------------------------------%

\indent {\bf Example:} Fig.~\ref{fig:base_obs} illustrates the selection of base set viewing regions on a~six target example. Initially all viewing regions in $\Ws$ have overlaps. Viewing region $V(T_5)$ is randomly selected and removed from $\Ws$ (including its overlap with region $V(T_4)$). Viewing region $V(T_6)$ is selected next, and removed with its overlapping regions $V(T_2),V(T_3),V(T_7)$ from $\Ws$. Hence, the base set is $\Is \!=\! \{ 5,6 \}$.~\eex

%---------------------------------------------------%
%\begin{figure}
%%\vspace{-.05in}
%\centering \includegraphics[width=.4\textwidth]%{maxgraphv2.png}
%\vspace{-.125in}
%\caption{The shortest collision free paths between all %initial sensing nodes for the six targets example of %Fig.~\ref{fig:base_obs}. The shortest paths are %embedded in the visibility graph which is not shown for %clarity.}
%\label{fig:visb}
%\end{figure}
%---------------------------------------------------%

\indent {\bf Initial sensing node selection:} Each base set viewing region, $V(T_i)$ for $i \!\in\! \Is$, is used to select initial ~sensing~nodes as follows. If $V(T_i)$ is an~isolated viewing region or intersects an obstacle vertex, the target $T_i$ at its center is selected as a~sensing node. If $V(T_i)$ overlaps a~neighboring viewing region, $V(T_j)$, a sensing node $S_j$ for the neighboring target $T_j$ is selected on the boundary of $V(T_i)$. These are the sensing nodes $S_2,S_3,S_4,S_7$ in Fig.~\ref{fig:base_obs}. Base regions $V(T_i)$ with neighbors but without obstacle vertex intersections have $S_i$ placed on the boundary of $V(T_i)$. Sensing nodes selection can be integrated into the overlap computation of the base set viewing regions, thus requiring only $O(n)$ additional time.
%, with one sensing node per target.

%The base set viewing regions, $V_{b}(T_i)$ %such~that~\mbox{$i \!\in\! \Is$,}  are used to select %initial sensing nodes as overlaps are detected. 
%If $V_i$ is an isolated viewing core, the target %T_i$ at its center is selected as a~sensing node. 

%As previously described, the process of detecting %overlaps arbitrarily looks for base-outer-arc to %neighbor-outer-arc overlap, base-outer-arc to neighbor-%occlusion-ray overlap, neighbor-outer-arc to base-%occlusion-ray overlap, and finally base-occlusion-ray %to neighbor-occlusion-ray overlaps. The sets describing %the viewing regions, $\theta_i$ and $p_i$, are ordered %clockwise from $-\pi$, thus setting detection order %within each category. The node for the neighboring %viewing region is placed at the first detected overlap, %subject to this ordering. This could of course be re-%ordered or randomized without affecting the bound.

%inally, if there are occlusion rays a~sensing node for %$V_i$ itself is selected at $T_i$, otherwise the node %is placed on its perimeter, arbitrarily. 

%for targets beside the base set is part of the %overlap detection process, so the only new step is %base target node placement, in $O(n)$ time. 

%that is furthest from $T_i$  

%\textcolor{red}{\indent {\bf Example:} Fig.~11(b) illustrates initial %sensing nodes selection using the base set of viewing cores from the %previous example.~\eex}

\indent {\bf Hoogeveen  \boldmath{$S$}-to-\boldmath{$T$} shortest path approximation:} The algorithm that solves Problem~$\# 3$ proceeds much like the previous problems' solution.  Algorithm~\ref{alg:hoogeveen} ~is ~used ~to compute a ~Hoogeveen path that starts at $S_1 \!=\! T_1$,~visits~the 
sensing nodes $S_2,\ldots,S_{n-1}$ in some order, and ends at~sensing~node~$S_n$ that observes target $T_n$. Hoogeveen's algorithm  needs as input a~complete graph of shortest collision free paths connecting all sensing nodes in the polygonal environment. The {\em visibility graph}~\cite{WELZL1985167} %connects all sensing nodes and convex obstacle vertices with collision free lines.  It 
contains the shortest collision free paths between all sensing nodes and can be computed in $O((m_{cvx} \!+\! n)^2 \log(m_{cvx} \!+\! n))$ time~\cite{WELZL1985167}, where $m_{cvx} \leq m$ is the total number of convex obstacle vertices in the environment, which simplifies the visibility graph. From the constructed visibility graph, the shortest collision free paths between all sensing node pairs can be computed in $O((m_{cvx} \!+\! n)^3)$ additional time using Johnson's algorithm \cite{10.1145/321992.321993}.
%additional time. 
The Hoogeveen path itself then takes $O(n^3)$ time to compute.

%---------------------------------------------------%
\begin{figure}
%\centering{\includegraphics[width=.5\textwidth]{fig13v2.png}}
%\centering{\includegraphics[width=.5\textwidth]{fig12v1nhoognoptnsens.png}}
\centering{\includegraphics[width=.5\textwidth]{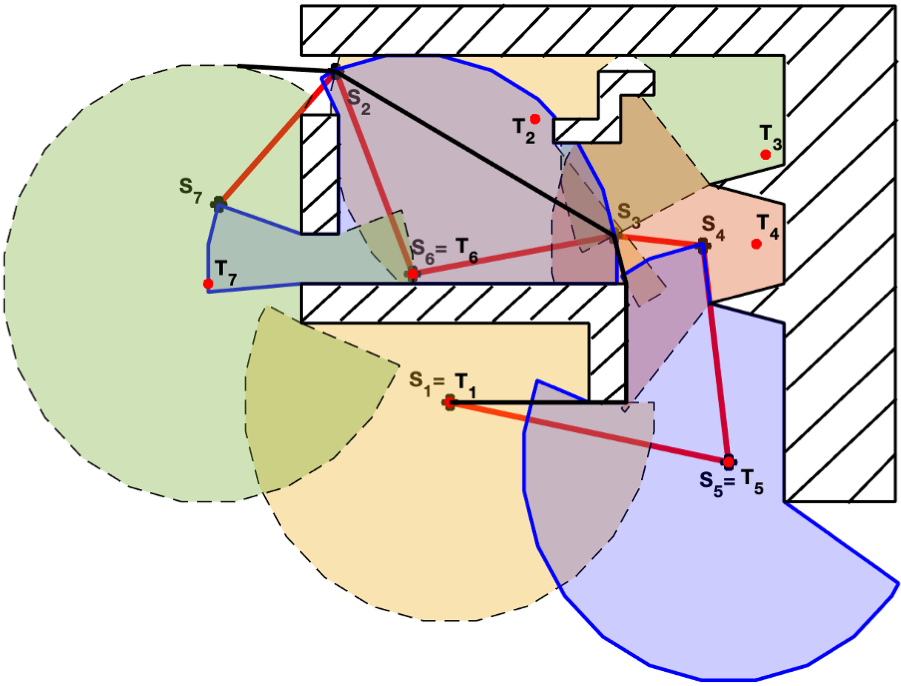}}
    \vspace{-.28in}
\caption{ The Hoogeveen $S$-to-$T$ path, shown in red, passes through the sensing nodes $S_1,\ldots,S_7$ shown in Fig.~\ref{fig:base_obs}. The optimized path is shown in black.}
\label{fig:hoog_obs}
\end{figure}
%---------------------------------------------------%

\indent {\bf Example:} Fig.~\ref{fig:hoog_obs} shows the Hoogeveen path on the seven target example from Fig.~\ref{fig:base_obs}, computed in .0195 seconds, connecting the initial sensing nodes $S_1,\ldots,S_6$, shown in red.~\eex

\begin{figure}
%\centering{\includegraphics[width=.5\textwidth]{fig14v2.png}}
\centering{\includegraphics[width=.5\textwidth]{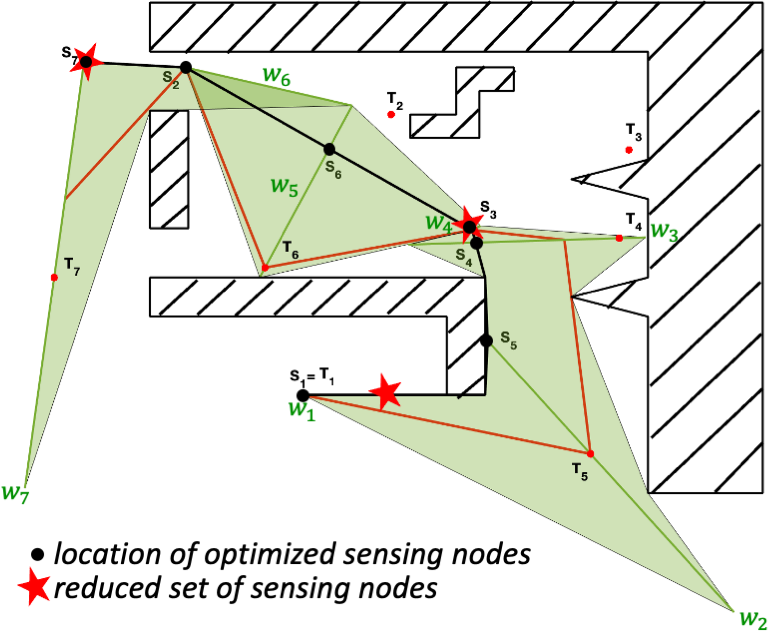}}
    \vspace{-.28in}
\caption{The seven target example of Fig.~\ref{fig:base_obs} showing the connected six corridors bounded by the sensing node $S_1 \!=\! T_1$ (a single-point segment), and then by linear segment of size $\leq 2r$ that lie in the viewing regions $V(T_{H}(i))$ and contain the sensing nodes~$S_{H(i)}$ for $i \!=\! 2\ldots 7$ (bold green segments marked with $w_i$). The optimized coverage path shown in black, with full set of sensing nodes marked. After sensing node reduction, the first reduced sensing node, the red star, observes $T_1$ and $T_5$, the second star at $S_3$ observes $T_2,T_3,T_4,T_6$, and the final reduced sensing node, coincident with $S_7$ observes $T_7$}
\label{fig:ltc}
\end{figure}

\indent {\bf Sensing node location optimization:} The sensing nodes are indexed according to the Hoogeveen path visitation order as $S_1,\ldots,S_n$. The algorithm~optimizes~the sensing node locations inside {\em obstacle free corridors}, defined as follows.  The $i^{th}$ corridor is bounded at both ends~by~linear~seg\-ments, $w_i$ and $w_{i+1}$, that pass through sensing nodes $S_i$ and $S_{i+1}$ (bold green segments in Fig. ~\ref{fig:ltc}).  The $i^{th}$ corridor's {\em left wall} is the shortest collision free path connecting the segments' left endpoints, its {\em right wall} is the shortest collision free path connecting the segments' right endpoints (Fig.~\ref{fig:ltc}).  Both walls are computed in the path homotopy class of the  Hoogeveen path connecting $S_i$ and $S_{i+1}$ in the polygonal environment.\footnote{Two paths that connect endpoints of $w_i$ and $w_{i+1}$  belong to the same {\em path homotopy class} when the loop formed by the two paths with $w_i$ and $w_{i+1}$  does {\em not} surround any internal obstacle.}  The $i^{th}$ corridor, called an {\em hourglass}~\cite{Kapoor1997-bq}, contains all shortest collision free paths that connect points of $w_i$ with those of $w_{i+1}$ within the path homotopy class, provided that $w_i$ and $w_{i+1}$ do not intersect each other (nor their obstacle-free linear extensions).

%Computation of the i'th corridor takes $O(m)$ time, %where $m$ is the total number of obstacle %vertices~\cite{HERSHBERGER199463}. 
%\note{Figure \ref{fig:ltc} caption seems odd, and has some typos %%%$V(T_{H(}i)$.  There are not six corridors highlighted in this figure.}
%-------------------------------------------------%
%---------------------------------------------------%

The corridors 
%that surround the Hoogeveen path 
are constructed as follows. The first corridor is spanned by the sensing node $S_1 \!=\! T_1$ (a~single-point segment) and the linear segment of size $\leq 2r$ that lies in the viewing region~$V(T_2)$ and contains sensing node~$S_2$ ($w_2$ in Fig.~\ref{fig:ltc}). Each subsequent corridor is bounded by linear segments $w_i$ and $w_{i+1}$ of size $\leq 2r$ that lie in their viewing regions $V(T_i)$ and $V(T_{i+1})$ and contain the sensing nodes $S_i$ and~$S_{i+1}$ (Fig.~\ref{fig:ltc}). The segments $w_i$ and $w_{i+1}$ are clipped into collision free sub-segments when they overlap nearby obstacles. Computation of the $n \!-\! 1$ corridors takes $O(nm^2 \log m \!+\! m \!\cdot\! n )$ time.

%These line segments will produce valid results as long as they intersect the Hoogeveen path at sensing nodes and are fully within the viewing region for a sensed target, so maximizing length is preferred. Two methods are attempted: connect the sensing node to its target and extend this segment until it leaves the viewing region, and bisect the incoming and outgoing Hoogeveen path angles with a line segment that emanates in either direction from the sensing nodes until it intersects the viewing region boundary. The longer of the two, at worst the sensing node itself, is used for the window.

\indent The corridors are next partitioned into triangular cells whose vertices lie on the corridor walls (Fig.~\ref{fig:ltc}). The triangulation is {\em linear} in the sense that the Hoogeveen path successively crosses each of these triangles, thus imposing linear order on the triangles of each corridor.  Using {\em boundary constrained triangulation} ~\cite{preparata2012computational}, triangulation of the $n \!-\! 1$ corridors  takes $O(n\!\cdot\! m \log( m))$ time.

\indent {\bf Example:} Fig.~\ref{fig:ltc} shows the corridors that surround the Hoogeveen path for the seven target example of Fig. ~\ref{fig:base_obs}, computed and triangulated in 1.77 seconds. 
~\eex

The sensing nodes locations are next optimized within the %linearly ordered 
corridors as follows.
%The  Hoogeveen path serves as a~seed path while the optimization %variables are 
%The {\em path via points} 
%that vary on edges of the corridor triangles.  
Let the $i^{th}$ corridor consist of $k_i$ triangles.  Let $e_{i,0}, \ldots, e_{i, k_i}$ denote the triangle edges crossed by the Hoogeveen path, indexed according to the linear order imposed by the Hoogeveen path crossing the edges. 
%In particular,
%$e_{i, k_i} \!=\! e_{1, k_{i+1}}$ for $i \!=\! 1 \ldots n-1$.
The triangle edges $e_{i,0} \!=\! w_i$ and $e_{i, k_i} \!=\! w_{i+1}$ bound the $i^{th}$ corridor and contain the sensing nodes $S_i$ and~$S_{i+1}$.  To ensure that each sensing node observes its target,
each $S_i$ varies in a~segment formed by the intersection of $e_{i,0}$ with the viewing region $V(T_i)$, such that the segment contains the initial sensing node~$S_i$.

%located on the Hoogeveen path. Windows are selection %within the viewing region guarantees this condition.

The path cost is now measured as a~function of the {\em path via points} that vary on the triangle edges crossed by the Hoogeveen path. These via points are denoted $\{ p_{i,0}, \ldots, p_{i,k_i}\}$ for $i \!=\! 1 \ldots n \!-\! 1$, and the path cost is %the sum of norms
\vspace{-.03in}
%\begin{eqnarray} \label{eq:min_obs_cost}
\[
       Cost( \{p_{1,j}\}_{i=1..n-1,j=0..k_i}  ) \\
       \! =  \! \sum_{i=1}^{n-1} \sum_{j=0}^{k_i - 1}  \norm{p_{i,j+1} \! -  \! p_{i,j} }  
\vspace{-.01in}
\]
\noindent 
such that  $p_{1,0} \!=\! \mbox{\small $T$}_1$ and \mbox{$p_{i+1,0}  \!=\! p_{i,k_i}$}~for~\mbox{$i \!=\! 1 \ldots n \!-\! 2$.} Each corridor entry point, $p_{i,0}$,
%except $p_{1,0}$  
is subject to the viewing constraint 
%points $p_1, \ldots, p_k$ vary in  
\[
p_{i,0} \in e_{i,0} \cap V(T_i)  \rtxt{\hspace{.5em} $i= 1 \ldots n \!-\! 1$}
\]
such that the $i^{th}$ viewing sub-segment contains the initial sensing node $S_i$ while the remaining via points are subject to the triangle edge constraints
\vspace{-.08in}
\[
p_{i,j} \in e_{i,j} \rtxt{\hspace{1.5em} $i= 1 \ldots n-1, \, j= 1 \ldots k_i$.}
\vspace{-.06in}
\]
\noindent 
The path cost forms a~convex function of the via points when they vary on linear segments. Moreover, the straight line segments between neighboring via points lie in obstacle free triangular cells. Convex optimization can thus efficiently optimize the sensing node locations within the corridors that surround the Hoogeveen path, as illustrated in the following example.

\indent {\bf Example:} Consider again the seven target example. Fig. ~\ref{fig:ltc} shows the sensory coverage path (in black) obtained by optimizing the path via point locations, starting from the initial sensing nodes located on the Hoogeveen path (shown in red). The optimized path was computed in .589 seconds, with sensing nodes located at $S_1 \!=\! T_1$, then at $S_i \!=\! p_{i,0}$ for $i \!=\! 2 \ldots n-1$ and finally at $S_n \!=\! p_{n-1,k_{n-1}}$ where $n=7$.~\eex

{\bf Sensing node reduction:}  Sensing node reduction %along the optimized path 
is performed by Algorithm~\ref{alg:node_reduction}. 
%from Section~\ref{sec:fewer}. 
%The optimized sensory coverage path is piecewise linear %with two types of sensing nodes (Fig.~13). Sensing %nodes at path vertices are essential and cannot be %reduced.
%since they represent single-point visit
%in a~particular viewing region along the optimized %path. 
%Sensing nodes in the interior of linear segments of the %optimized path 
%are {\em non-essential} and 
%can potentially be reduced.  
The algorithm now requires the overlaps of the optimized path with the $n$ viewing regions. These overlaps can be computed in $O(n^2 \!\cdot\! m_r)$ time,
%where $m_r$ is the maximum number of obstacle vertices within %detection range $r$ surrounding each target. The overlaps 
then fed into Algorithm~\ref{alg:node_reduction} which selects in  $O(n)$ additional time a~reduced set of sensing nodes. % along the optimized path.
These nodes observe all targets using multi-target line of sight views. Fig.~\ref{fig:ltc} shows the reduced sensing nodes %$S_1,S_2,\ldots$ 
along the optimized path, computed in .0279 seconds.

%{\bf Remark:} Further sensing nodes reduction and shortening of the sensory coverage path can be achieved
%by iteratively running Algorithms~\ref{alg:node_reduction}~and~\ref{alg:overlap_reduce}.
%%The optimized sensing node positions, $\Ss_{opt}$,
%%have changed as a~result of the convex optimization performed in line~\ref{line:convex} of Algorithm~\ref{alg:overlap_reduce}.
%In each iteration one runs Algorithm~\ref{alg:node_reduction} on the optimized nodes of $\Ss_{opt}$ for sensing nodes reduction. If the number of sensing nodes
%has been reduced,  one runs Algorithm~\ref{alg:overlap_reduce} to optimize the positions of the smaller number of sensing nodes, until no more sensing node reduction is achieved.~\eex
%\vspace{.02in}

%%% -------------------------------------------------------------- %%%
%\vskip -0.25 true in
\begin{algorithm}[ht]
%\textcolor{blue}{
\caption{Multi-Target Cover Among Obstacles}
\hspace{.2em} {\bf Input:} $\!$ Targets \mbox{\small $\Ts \!=\! 
\{T_1,\ldots ,T_n\}$,} polygonal obstacle ver- \\
\mbox{\hspace{.5in}} tices~$\Os$, $|\Os| = m$.
%~\mbox{Sensing}~\mbox{nodes}~\mbox{\small $\Ss \!=\! \{S_1\ldots S_m\}$}
\begin{small}
\begin{algorithmic}[1]
\Procedure{ReducedSensingPath}{$\Ts$,  $\Os$}
\State {\! $\{ V(T_1), \ldots, \! V(T_n) \} = ViewingRegions(\Ts, \Os)$ } \label{line:regions}
\State {\! $\Is = BaseSetCicrcles(\Ts, \mbox{\small  $\{ V(T_i) \}$})$ \label{line:sensing_nodes} \label{line:base_set} }
\State {\! $\{ S_1, \ldots, S_n \} = BaseSetSensingNodes(\Is, \mbox{\small  $\{ V(T_i) \}$})$ }
\State {$\Vs = VisibilityGraphShortestPaths(\{S_1, \ldots, S_n\},\Os)$}
%\State { Set $\Ss = \{ S_1, \ldots, S_n \}$ }
\State {\! $\mathcal{P}_{order}$\ =\ {\rm Hoogeveen}($S_1, \ldots, S_n,\Vs$) 
\label{line:h_path} }
\State {\!  $\{\xi_{ij}\}\ =\ {\rm Incidence}(P_{order})$ }
\State { ($\Ps, \Ss_{opt} ) = \argmin {\rm Cost}\big\{ \{\xi_{ij}\},\Ss,\mathcal{T} \big\}$ } \label{line:path_opt}
\State { $\quad  {\rm subject \  to}  \ S_1 \!=\! T_1 \ {\rm and} \ S_i \!\in\! e_{i,0} \cap \mbox{\small $V(T_i)$}$  $i =1 \ldots n$ }
\State { $\Ss_{min}$ = SensingNodeReduction($\Ts$, $\Ps$, $\Ss_{opt}$) 
\label{line:path_reduce} }
%\State{Set $n_s = |\Ss_{min}|$, \ $S_1 = T_1$}
\State { $Return\big( \mbox{\rm path} \ \Ps, \mbox{\rm reduced sensing nodes} \ \Ss_{min} \big)$ }
\EndProcedure
\end{algorithmic}
\end{small}
\label{alg:overlap_obstacles}
%}
\end{algorithm}
\vspace{-0.075 true in}
%%% -------------------------------------------------------------- %%%

The approximation algorithm that solves Problem~$\# 3$ is summarized as Algorithm~\ref{alg:overlap_obstacles}. It consists of seven stages. The $n$ viewing regions are computed in $O( n \!\cdot\! m^2_r \log m_r)$ time, where $m_r$ is the maximum number of obstacle vertices and edges within  detection range $r$ of each target. Next, a~base set of viewing regions and initial sensing nodes are selected in $O(n^2 m_r \log m_r )$ time. Shortest obstacle free paths between sensing nodes are computed in $O((n+m_{cvx})^3)$ time. The Hoogeveen path that passes through the sensing nodes is computed 
%in line~\ref{line:h_path} 
in $O(n^3)$ time.  Obstacle free corridors 
%that surround the Hoogeveenpath 
are next computed and triangulated in  $O(m^2_r \log m_r \!+\! m \!\cdot\! n )$  time. Convex optimization of the path via points %located on the triangle edges 
is performed 
%in line~\ref{line:path_opt} 
in $O(k n^2 \ln(1/\epsilon))$ time.  Finally, sensing node reduction 
%along the optimized path 
is performed 
%in line~\ref{line:path_reduce} 
in $O(n^2 \cdot m_r)$ time.

%\begin{figure}
%\centering{\includegraphics[width=.5\textwidth]%{sensredux.png}}
%    \vspace{-.28in}
%\caption{The six targets example of %Fig.~\ref{fig:base_obs} showing the optimized $S$-%to-$T$ path passing through the sensing nodes %$S_1,\ldots,S_3$ shown, in black. $S_1$ observes $T_1$, %$S_2$ observes $T_3,T_4$ and $S_3$ observes %$T_2,T_5,T_6$ }, 
%    %(see text).}
%\label{fig:sensredobs}
%\end{figure}

%----------------------------------------------------------------------------------------------------------%

{\bf  Path length approximation bound:} The  following proposition 
%describes the path length 
%approximation bound for %Algorithm~\ref{alg:overlap_obstacles}. 
assumes that all base set viewing regions have some area $ A_i > 0$ which is captured by a radius $\rho_i$ such that $A_i = \pi \rho_i^2 $ for $i \!=\! 1\ldots n$. 
%\vspace{.01in}

%Recall that each 
%viewing region, $V(T_i)$, contains $d_i$ occlusion %rays that emanate from the target $T_i$. 
%The following proposition 
%The parameters $\bar{d}_i \! \leq\! d_i$ ($i \!=\! %1\ldots n$) is the number of occlusion rays of $%V(T_i)$ that contain sensing nodes belonging to %neighboring viewing regions, termed the {\em engaged} occlusion rays of $V(T_i)$. 

\begin{prop} \label{prop:bound3}
Let targets $T_1,\ldots,T_n$ be located among polygonal obstacles. Let a~point robot move in the freespace using a line-of-sight coverage sensor of detection range~$r$.
%possibly overlapping viewing regions of maximal radius $r$.
%centered at the targets.
Algorithm~\ref{alg:overlap_obstacles} computes in $O((m \!+\! n)^3 \log (m \!+\! n) )$ time a~sensory coverage path whose length satisfies the approximation bound
\vspace{-.05in}
%\begin{equation} \label{eq:bnd_obs}
\[ 
l(\mbox{\rm Alg.~\ref{alg:overlap_obstacles}})  \leq 
\tfrac{5}{3} \big( ( 1 \!+\! 8 \frac{r^2}{ \bar{\rho}^2}) \cdot l_{opt} + 2r \!\cdot\! m \big)
 + c  
  \vspace{-.05in}
\]
\noindent where $l_{opt}$ is the length of the optimal $S$-to-$T$ coverage path in the polygonal environment, $m$ is the total number of obstacle vertices, $\bar{\rho} \!\leq\! r$ is the radius of a circle with area equal to the average base set viewing region areas and $c \!=\! (\tfrac{5}{3} \!\cdot\! \tfrac{8 r^2}{\bar{\rho}^2} \!\cdot\!  \pi + 2) r$ is a constant.
\end{prop}
\vspace{.02in}

%the parameter $\bar{d}$ is the average number of %engaged radial sectors of the viewing regions.

%{\bf Remark:} When the targets viewing regions  do %{\em not} overlap, the path length bound becomes %essentially the path length bound in the obstacle %free environment, multiplied by the ratio 
%$(r/\bar{\rho})^2$ that measures obstacle congestion %about the targets and additional path length  $2r m$
%incurred by occlusion edges in the environment.~\eex

%in the polygonal environment, 
%$\bar{d}_i \!=\! 0$ for $i \!=\! 1 \ldots n$ and then %$\bar{d}\!=\! 0$ in Eq.~\eqref{eq:bnd_obs}.  

{\bf Proof:}
Let $\Ag_{opt}$ be the shortest $S$-to-$T$~sensory coverage path, with length $l_{opt}$. The optimal path must visit every base set viewing region, $V(T_i)$ for $i \!\in\! \Is$, otherwise the target~$T_i$  at the center of $V(T_i)$ cannot be observed by the robot. Consider the area swept by a~disc of radius $2r$ while its center moves along $\Ag_{opt}$
\vspace{-.06in}
\begin{equation} \label{eq.area_obs}
\As = 4r \cdot l_{opt} + 4 \pi r^2
\vspace{-.06in}
\end{equation}
\noindent where $4 \pi r^2$ is the area of two half-discs at the endpoints of~$\Ag_{opt}$.  Since each $V(T_i)$ lies in a~sensing circle of radius~$r$ centred at~$T_i$, the area swept by the disc of radius $2r$ along $\Ag_{opt}$ covers all base set viewing regions. By construction, the base set viewing regions are {\em disjoint} and contain areas equal to circles with radius $\rho_i$ for $i \!\in\! \Is$. Hence
\vspace{-.08in}
\begin{equation}  \label{eq:right_av}
\As \geq \pi  \cdot \sum_{i=1}^{|\Is|} \rho^2_i 
%\geq 
%\pi \cdot \big( \frac{1}{|\Is|} \sum_{i=1}^{|\Is|} \rho_i \big)^2
\vspace{-.06in}
\end{equation}
\noindent where $|\Is|$ is the number of base set viewing regions. 
%Eq.~\eqref{eq:right_av} can be equivalently written in %terms of 
%\[
%\As \geq |\Is| \cdot  \pi \bar{\rho}^2 
%\vspace{-.06in}
%\]
Using the identity
\[
\sum_{i=1}^{|\Is|} \rho^2_i \geq \frac{1}{|\Is|} \cdot \big( \sum_{i=1}^{|\Is|} \rho_i \big)^2 
\]
we obtain the lower bound on $\As$
\[
\As \geq  |\Is| \cdot \pi \bar{\rho}^2
\]
where $\bar{\rho} \!=\! \sum_{i=1}^{|\Is|} \rho_i / |\Is|$ is the {\em average radius} of the base set viewing region areas.  Substituting for $\As$ according to Eq. ~\eqref{eq.area_obs} gives
%into the latter inequality gives
%\vspace{-.06in}
\begin{equation} \label{eq:m_obs}
 |\Is|  \leq \frac{4r}{\pi \bar{\rho}^2} ( l_{opt} + \pi r ).
%\vspace{-.02in}
\end{equation}
\noindent % Next consider 
The $S$-to-$T$ Hoogeveen path computed by Algorithm~5
%in the polygonal environment. 
%This path 
starts at~$S_1 \!=\! T_1$, visits $S_2,\ldots,S_{n-1}$ in any order, then ends at $S_n$ within detection range of~$T_n$. Using Lemma~\ref{lemma:hoogeveen}, the length of this path, $l_H$, satisfies the bound $l_H \leq \tfrac{5}{3} l_{TSP}$ where $l_{TSP}$ is the length of the shortest $S$-to-$T$ path. 
% in the polygonal environment  that starts at~$S_1$, visits $S_2,\ldots,S_{n-1}$ in any order and ends at~$S_n$.   
The convex optimization step of Alg.~5 only {\em shortens} the Hoogeveen path.  Hence $l(\mbox{\rm Alg.~5)} \leq l_H \leq \frac{5}{3} l_{TSP}$.

An alternative path, $\mathcal{P}$, takes advantage of the sensing node locations on the base set viewing regions' perimeters and centers. 
%Each base set viewing region either has sensing %nodes on its boundary, or a~single sensing node at %its center.
Path $\mathcal{P}$ starts at $S_1$ and moves along $\Ag_{opt}$. When $\Ag_{opt}$ reaches a~base set viewing region, $\mathcal{P}$ circumnavigates the viewing region's outer perimeter when the perimeter contains %perimeter
sensing nodes, or executes a straight in-and-out motion to its center when a~single sensing node lies at its center. The path then continues along $\Ag_{opt}$ to the next base set viewing region.  

The length of $\mathcal{P}$, $l_b$, satisfies the following upper bound. Because the base set viewing regions are disjoint,
%Using the fact that non-intersection of base viewing %regions necessitates a bound on $\sum_{i=1}^{|\Is|} %\bar{d}_i \leq 2 \cdot m $ 
each obstacle vertex lies in a~single base set viewing region and bounds at most one occlusion edge. A~full tour of a~base set viewing region perimeter traverses each occlusion edge and each sensing circle arc once.
%Also, if a vertex does {\em not} appear in any %viewing region, it cannot affect the length of any %viewing region perimeter. Thus, only convex vertices %in the %base set 
%viewing regions affect 
If $m_r$ denotes the total number of obstacle vertices within detection range of the targets, the length of $\mathcal{P}$ is bounded by
\[
l_b \leq l_{opt} + 2 r \cdot (\pi \!\cdot\! \mbox{\small  $|\Is|$} + m_r + 1)
\]
where the term $2 \pi r \mbox{\small  $|\Is|$}$ bounds the total length of the {\em circular arcs and obstacle edges} on the  viewing regions' perimeters, while the term $2  r m_r$ bounds the total length of the viewing regions occlusion edges. When a single sensing node lies at the center of a~base set viewing region, the detour of length $2 r $ to this node is upper bounded by $2 \pi r $.

%The term  $2 \bar{d}_i r$ bounds the length of the %path required to navigate the engaged occlusion rays %and return to the arc. 

%The perimeter of each viewing core $V(T_i)$ consists of circular arcs of %maximal radius~$r$ centered at~$T_i$, \textcolor{red}{circular arcs of %radius $\rho_i$ containing the target}, a~single obstacle edge in each %radial sector of $V(T_i)$ and $d_i$ {\em radial occlusion rays} whose %maximal length is also~$r$. Based on this insight, 
%\begin{equation} \label{eq:obs_lb}
%\[
%l_b \leq l_{opt} + \sum_{i=1}^{|\Is|} \big(2 \pi r  +  2 %\bar{d}_i   r \big)
%\]
%having the following terms.
%The term $2 \pi r $  bounds the length of the circular %arcs and obstacle edges on the i'th viewing region %perimeter. The term  $2 \bar{d}_i r$ bounds the length %of the path required to navigate the engaged occlusion %rays and return to the arc. Should there only be one %node at the center, this detour of $2 r $ is lower %bounded by the $2 \pi r $.

%\textcolor{green}{Define the {\em average number} of %engaged occlusion rays 
%%over the base set 
%as $\bar{d} \!=\! \sum_{i=1}^{|\Is|} \bar{d}_i / |\Is|$. % Using this parameter, the bound on $l_b$ can be written %as
%\[
%l_b \leq l_{opt} + 2 r \big(\pi  + \bar{d}  \big)  \cdot %|\Is| + r
%\]}

The alternative path $\mathcal{P}$ starts at~$S_1$, circumnavigates the base set viewing regions while visiting the sensing nodes $S_2,\ldots,S_{n-1}$, then ends at $S_n$. This path is {\em longer} than the optimally shortest  path that starts at~$S_1$, visits the same intermediate nodes, then ends at $S_n$. Thus $l_b \geq l_{TSP}$ and the path length bound on Algorithm~\ref{alg:overlap_obstacles} becomes

\begin{equation} \label{eq:obs_p_b}
l(\mbox{\rm Alg.~\ref{alg:overlap_obstacles}})
 \leq  \tfrac{5}{3}  l_{TSP} \! \leq \! \tfrac{5}{3} l_b  \leq \tfrac{5}{3} \big( l_{opt} \!+\!  2 r \cdot (\pi \!\cdot\! |\Is| + m + 1) \big) \, 
\end{equation}
% \tfrac{5}{3} \big( l_{opt} \!+\!  2 r (\pi \!+ \! %\bar{d} ) \cdot |\Is|  \big) \, .
Substituting the upper bound for $|\Is|$ specified in Eq.~\eqref{eq:m_obs} and $m_r \leq m$ gives the path length bound
\[
l(\mbox{\rm Alg.~\ref{alg:overlap_obstacles}}) \leq
\tfrac{5}{3} \big( ( 1 \!+\! 8 \frac{r^2}{ \bar{\rho}^2}) \cdot l_{opt} +  2r \!\cdot\! m \big) + c 
\]
%\tfrac{5}{3} \big( \frac{8 r^2}{\pi \bar{\rho}^2}  
%( \pi  \!+\!  \bar{d} ) \big) \cdot l_{opt}  + c  
where $c \!=\! (\tfrac{5}{3} \!\cdot\! \tfrac{8 r^2}{\bar{\rho}^2} \!\cdot\! \pi + 2) r$ is a constant.~\epf
\vspace{.03in}

{\bf Remark:} One can construct several variations on the path length bound of Algorithm~\ref{alg:overlap_obstacles}. One example follows. 
%In particular, the assumption that the base set viewing %regions contain obstacle free circular areas of radius %$\rho_i$ for $i \!\in\! \Is$ can be prohibitive. There %are many alternative approaches. 
Given an~environment and targets, $|\Is|$ is a~known quantity. Additionally, the viewing region areas can easily be computed with the knowledge $\As \geq  \sum_{i=1}^{|\Is|} area(V(T_i)) $. Hence, one can replace $\As$ with the average, $\As \geq |\Is| \cdot E[area(V_{i \in \Is})]$.  This can serve as the starting point for developing probabilistic path length bounds. By using sample data points or known distributions, the expected value for the base set area or number of obstacle vertices could be substituted into Eq. (\ref{eq:obs_p_b}) to generate a predicted bound

\[
l_b \leq l_{opt} + 2 \cdot  r \big(\pi \cdot \frac{4r l_{opt} + 4 \pi r^2}{E[area(V_{i \in \Is})]} + m  \big)
\]
\[
l_b \leq (1 + \frac{8 \pi \cdot r^2  }{E[area(V_{i \in \Is})]})\cdot l_{opt} + 2 \cdot r \cdot m + \tilde{c}
\]

where $\tilde{c}$ is a constant.
{\eex}

%%% ------------------------------------------------------------- %%%
\section{Conclusion}
\vspace{-.02in}

%Motivated by robotic search and inspection tasks,

\noindent This paper considered three successively more demanding mobile robot sensory coverage problems: multi-target coverage in obstacle free environments with non-overlapping sensing circles centered at the targets, multi-target coverage in obstacle free environments with possibly overlapping sensing circles that allow multi-target views, and multi-target coverage in environments populated by obstacles that may obstruct sensor views of the targets. All three problems were formulated as {\small NP-Hard MINLP} optimization problems. 
%that combine the selection of sensing nodes visitation order with optimization of their location for the shortest sensory coverage path, under limited detection range. Obstacle constraints appear in the third problem.

For each problem, we introduced novel approximation algorithms that can be computed in polynomial time. All three problems were solved in roughly two stages. Initial sensing node locations were selected and a~visitation order for these nodes was computed in polynomial time.  Then efficient convex optimization was used to optimize the sensing node locations, yielding the shortest sensory coverage path under the set visitation order. In the second and third problems, multi-target views were used to reduce the number of sensing nodes, when possible. The algorithm solving the third problem uses obstacle obstructed viewing regions and shortest collision free path between sensing nodes to enable highly competitive execution times.

The paper developed path length bounds for the three algorithms, expressed in terms of each problem's optimal path length. These path length bounds 
provide limits for the gap between the exact solutions for these {\small NP}-hard problems and our polynomial time approximate solutions. Finally, all algorithms have been fully implemented in MATLAB
%and illustrated with execution examples, 
with software available as part of the paper. 

Future research will focus on two areas. First, there is a~need to efficiently search for and inspect multiple targets whose position is only approximately known to the robot. The robot must efficiently inspect {\em viewing neighborhoods} surrounding these targets, with provable efficiency bounds. Next, while the general environment layout might be known to the robot, a priori unknown obstacles may be encountered during the robot's actions. Here, too, there is a~need forprovable efficiency bounds, most likely expressed in terms of learned probabilistic obstacle distributions  and percentile sensory coverage bounds that use these distributions. The techniques and algorithms described in this paper provide a~strong foundation for future work on these problems.

Our future research also aims to extend the {\small MINLP} framework to  {\small 3-D} robot sensory coverage~and~inspect\-ion problems. Robotic agents can efficiently inspect exteriors of {\small 3-D} structures that form natural extensions of the planar environments considered in this paper.  When a~structure's exterior~is~only~approximately known to the robot, sensory coverage and inspection becomes a~volumetric {\small 3-D} planning problem. Our {\small MINLP} optimization framework can model sensors with convex {\small 3-D} footprints. 
%For instance, it seems that a~small number of great-circle paths suffice to completely inspect a~ball of maximal detection range centered on each target. 
Hence, the techniques described in this paper have a good chance of providing highly efficient algorithms even for the most challenging {\small 3-D} sensory coverage and inspection planning problems. 

\begin{appendices}
\label{app:ratio_proof}

\section{Proof of Proposition \ref{prop:bound1}}. 
To analyze the approximation bound, we must find the worst case ratio $(l_{TSP}/l_{alg})$ over all possible arrangements of targets, where $l_{TS}$ is the TSP path ($\mathcal{P}_{TSP}$) through all targets, and $l_{alg}$ is the shortest possible sensing path ($\mathcal{P}_{CPP}$) that views all targets .  We  begin by assuming that the worst case arrangement of $n$ targets consists of sensing circles arranged in a closely packed linear fashion (see Figure \ref{fig:RatioDiagram}).  This assumption will be analyzed below. 

\begin{figure}
\includegraphics[width=0.48\textwidth]{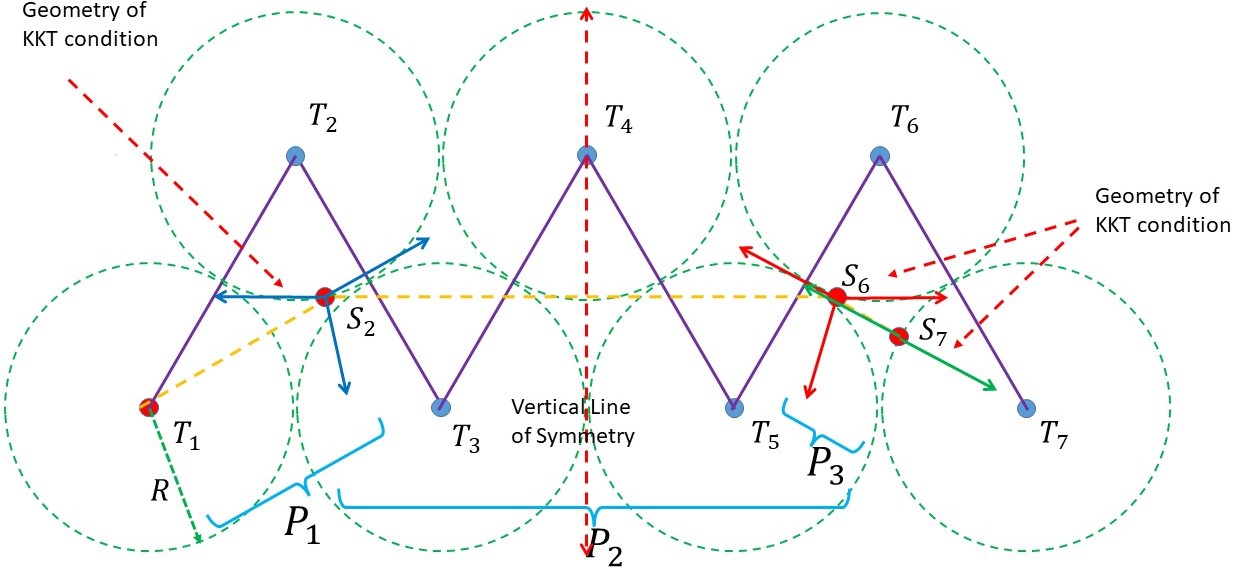}
\caption{Geometry of KKT conditions for odd number of targets.}
\label{fig:KKT_odd}
\end{figure}

The shortest TSP path between the targets in this arrangement is easily characterized.  To characterize the shortest sensing path and its associated sensing locations, we use the Karush-Kuhn-Tucker (KKT) optimality conditions. We first analyze the case of an odd number of targets. Let $T_1$ and $T_n$ be the start node and final target, respectively (see Fig. \ref{fig:KKT_odd} for a seven target example).  The optimal sensing path must contain a sensing node at $T_1$ and $n-1$ sensing nodes lying in the remaining target sensing regions. One can show that in a linear packing of sensor regions, the shortest path is defined by four sensing node locations: $S_1$ (located at $T_1$), $S_2$ in the sensing region around $T_2$, $S_{n-1}$ in the sensing region around $T_{n-1}$, and $S_n$ on the boundary of the sensing region surrounding $T_n$.  The length of such a path, $||P||$, takes the form (see Fig. \ref{fig:KKT_odd}):
   \begin{eqnarray*}  ||P|| &=& ||P_1|| + ||P_2|| + ||P_3|| \\ 
         &=&  ||S_2-T_1|| + ||S_{n-1} - S_2|| + ||S_n - S_{n-1}||.
   \end{eqnarray*} 
We seek to minimize $||P||$, subject to the constraint that the key sensing nodes lie in their associated sensing regions:
  \[ ||S_2 - T_2|| \le R; \ \ \ \ ||S_{n-1}-T_{n-1}||\le R; \ \ \ \ 
     ||S_n - T_n|| \le R. \] 
Define the {\em Path Lagrangian}, $\mathcal{L}_P$ as:
  \begin{eqnarray*} \mathcal{L}_P &=& ||P|| + \lambda_1 (||S_2 - T_2|| - R) \\
     &+&  \lambda_2 (||S_{n-1}-T_{n-1}|| 
     - R) + \lambda_3(||S_n - T_n|| - R). \end{eqnarray*}
The  minimal path length necessary KKT conditions are:
  \begin{eqnarray}
    \frac{\partial \mathcal{L}}{\partial S_2} &=& 
      \frac{S_1 - T_2}{||S_2 - T_2||} - \frac{S_{n-1}-S_2}{||S_{n-1}-S_2||} 
             + \lambda_1 \frac{S_2 - T_2}{||S_2 - T_2||} \nonumber \\
    \frac{\partial \mathcal{L}}{\partial S_{n-1}} &=& 
       \frac{S_{n-1} - S_2}{||S_{n-1} - S_2||} - \frac{S_n-S_{n-1} }{||S_n -S_{n-1}||} \label{eq:KKT} \\
       && \quad + \lambda_2 \frac{S_{n-1} - T_{n-1}}{||S_{n-1} - T_{n-1}||} \nonumber \\
    \frac{\partial \mathcal{L}}{\partial S_n} &=& 
       \frac{S_n - S_{n-1}}{||S_n - S_{n-1}||} 
       + \lambda_3 \frac{S_n - T_n}{||S_n - T_n||} \nonumber 
  \end{eqnarray}
Since the ratios in these expressions can be interpreted as unit vectors, the KKT conditions yield insight into the shortest sensing path geometry.  Figure \ref{fig:KKT_odd} shows the geometry of the KKT conditions for an odd number of targets in the worst case target geometry.

\begin{figure}
\centerline{ \includegraphics[width=0.3\textwidth]{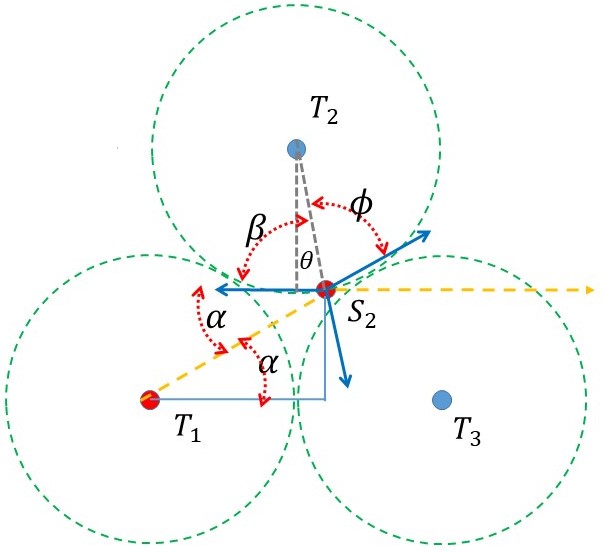} }
\vskip -0.07 true in
\caption{Detailed Geometry of KKT condition around sensing node $S_2$}
\label{fig:Node2_KKT}
\end{figure}

The first condition in Eq. (\ref{eq:KKT}) (see Fig. \ref{fig:Node2_KKT}) dictates that sensing node $S_2$ must be positioned so that the
line underlying $\vec{S_2 T_2}$ bisects the angle between the lines  underlying $\vec{T_1 S_2}$ and $\vec{S_2 S_{n-1}}$.  From Figure \ref{fig:Node2_KKT}, one can calculate that at a minimal path length, the angle between $\vec{T_2 S_2}$ and a vertical line, denoted $\theta^{*}_{o}$, is the solution to the following equation
  \begin{equation} \label{eq:thetastar_odd}
  \frac{(\sqrt{3}-\cos(\theta^{*}_{odd}))}{(1+\sin(\theta^{*}_{odd}))}=\tan(2\theta^*_{odd})
  \end{equation}
Using this result, the shortest path length for an odd number of sensing regions is
  \begin{equation} \label{eq:Podd}
      ||P_{odd}||\ =\ R[n-4+\kappa_{odd}(\theta^*_{odd})] \end{equation}
where 
   \begin{equation}\label{eq:kappa_odd} 
    \kappa_{o}(\theta^*_{o}) = 2\big[(1+\sin\theta^*_{o})^2 + (\sqrt{3}-\cos\theta^*_{o})^2\big]^{1/2} - 2\sin\theta^*_{o}
   \end{equation}
%%% ------------------------------------- %%%
%%% ------------------------------------- %%%
The shortest path length for an even number of targets can be analyzed in a similar way.  The shortest path length in this case is 
  \begin{align}
  \begin{split}
   R\big[(\frac{n-3}{2}-\sin\theta^*_{even})^2+(\cos\theta^*_{even} -
  \frac{\sqrt{3}}{2})^2\big]^{1/2} -R \nonumber
  \end{split}
  \end{align}
where $\theta^*_{even}$ is the solution to the following equation:
\begin{equation*}
\frac{\sqrt{3}-\cos\theta^*_{even}}{1+\sin\theta^*_{even}} =
\frac{\tan\theta^*_{even}-\eta}{1+\eta\tan(2\theta^*_{even})}
\end{equation*}
and $\eta=(2\cos\theta^*_{even}-\sqrt{3})/(n-3-2\sin\theta^*_{even})$.

One can show that as $n$ increases, $\theta^*_{even}\rightarrow \theta^*_{odd}$, and that $||P_{even}||$ is very well approximated by the expression of $||P_{odd}||$.  In fact, $||P_{odd}||$ in Eq. (\ref{eq:Podd}) is an under approximation of the path length, which ensures that  Eq. (\ref{eq:ratio}) is the worst case bound. Thus, $\kappa'_o = \kappa_o(\theta^*_o)$ yields the shortest optimal path, and the tightest bound in Eq. (\ref{eq:Podd}).

{\bf Worst Case Target Arrangement}.  We next analyze the claim that the packed linear arrangement of target sensing regions in Figure \ref{fig:KKT_odd} leads to the worst case ratio of the target TSP path length, $l_{TSP}$, to the shortest sensing path length, $l_{CPP}$.   Our goal is the find the target locations that maximize the ratio
    \[ \Delta \ \triangleq \ \frac{ l_{TSP}^{*}} { l^{*}_{CPP} }. \]
The general proof proceeds by induction.  Here we sketch the first few steps of the induction process. 

As above, we first develop conditions for extremal values of the ratio $\Delta$.  The goal is to {\em maximize} $\Delta$ with respect to the choice of target locations (which is the same as {\em minimizing} $-\Delta$), subject to the $(n-1)(n-2)/2$ {\em overlap constraints} that no two
sensing regions overlap, and that target node $T_1$ is fixed to a specific location:
\begin{eqnarray}
   \min_{T_2,\ldots,T_n} & - & \Delta(T_2, \ldots, T_n) \label{eq:delta_problem} \\
  {\rm subject\ to:} && \ ||T_j - T_i||^2 \ge 4r^2 \nonumber 
\end{eqnarray}
for $i=1,\ldots,(n-1)$ and $j=2,\ldots,n$.

If the {\em $\Delta$ Lagrangian} is defined by:
 \begin{equation*}
   \Ls_{\Delta}(\Ns,\vec{\lambda}) \ =\ - \Delta + \sum_{i=1}^{n-1}\ \sum_{j=i+1}^n\ \lambda_{i,j}
   \big(r^2- ||T_j - T_i||^2 \big)
 \end{equation*}
then the Karush-Kuhn-Tucker (KKT) conditions for a constrained minimum are:
  \begin{eqnarray}
  0 &\in & -\frac{\partial\Ls_{\Delta}(T_2,\ldots,T_n,\vec{\lambda})}{\partial T_k}  
       =   -\frac{\partial \Delta}{\partial T_k} \label{eq:deriv}\\
   && \ \ \ \ + 2 \sum_{j=k+1}^{n} \lambda_{j,k} (T_j-T_k)
     - 2 \sum_{i=k-1}^n \lambda_{k,j} (T_l-T_i)  \nonumber \\
  0 & \ge & 4r^2 - ||T_j-T_i||^2     \label{eq:ineq}\\
  0 & \le & \lambda_{i,j}           \label{eq:poslambda}\\
  0 &=& \lambda_{i,j} (4r^2 - ||T_j-T_i||^2)  \label{eq:complementarity}
  \end{eqnarray}
where $i=2,\ldots, (n-1)$ in (\ref{eq:deriv})-(\ref{eq:complementarity}) and $j=(i+1),\ldots, n$ (with
$i\ne j$) in Eq.s (\ref{eq:ineq})-(\ref{eq:complementarity}). To calculate $\partial\Delta/\partial
T_j$, for $j=2,3$,
\[ \frac{\partial \Delta}{\partial T_{j,i}}
    = \lim_{\vec{\mu_i}\rightarrow 0}
   \bigg[ \frac{ \Delta(T_{j,i} + \mu_i) - \Delta(T_{j,i}) } {\mu_i} \bigg] \]
where $i=x$ or $i=y$, and $\vec{\mu}=\begin{bmatrix} \mu_x & \mu_y\end{bmatrix}$ denotes
small perturbations of a target's Cartesian locations.

Since function $\Delta$ may not be convex, conditions (\ref{eq:deriv}) -
(\ref{eq:complementarity}) are only {\em necessary} conditions for a maximal value of $\Delta$.  Note that the geometry of $\Ps_{CPP}$ and/or $\Ps_{TSP}$ may change abruptly with small changes in target placements. In these cases, function $\Delta$ can be non-smooth, and $\partial \Delta/\partial T_i$ must be interpreted as a generalized (set-valued) gradient.

\subsection{Case of Two Targets}

Let two targets, $T_1$ and $T_2$, be separated by a distance $2r + \mu$, where $\mu \ge 0$.  One can see by inspection that $l_{TSP}^{*} = 2r+\mu$ and $l^{*}_{CPP} = r+\mu$. Thus,
     \[ \Delta = \frac{l_{TSP}^{*}}{l^{*}_{CPP}} = \frac{2r+\mu}{r+\mu}.\]
This ratio takes its largest value when $\mu\rightarrow 0$.  I.e., when the boundaries of the circular sensing regions $R(T_1)$ and $R(T_2)$ nearly touch.  Equivalently, the non-overlapping constraint is active at the maximum.

\subsection{Case of Three Targets}

For targets $T_1$, $T_2$, and $T_3$, let $T_1$ be fixed to a given location.  Since $\Delta$ may not be convex, its extrema may occur when no, some, or all constraints are active. We first show that $\Delta$ cannot be maximized under with no active constraints.  We then analyze the cases of one, two, and three active constraints, concluding that $\Delta$ is maximized when the three targets form the smallest possible equilateral triangle.

KKT condition (\ref{eq:complementarity}) implies that if each pair of targets is more than $2r$ apart, then $\lambda_{1,2} = \lambda_{1,3} = \lambda_{2,3} = 0$, and an extremum in $\Delta$ occurs when the following two conditions hold simultaneously:
\begin{equation}\label{eq:NoConstraints}
    \vec{0} \ \in \ -\frac{\partial \Delta}{\partial T_2}; \ \ \ \
    \vec{0} \ \in \ -\frac{\partial \Delta}{\partial T_3}\ .
\end{equation}

If any derivative in (\ref{eq:NoConstraints}) is non-zero for all unconstrained target locations, the unconstrained KKT conditions are not satisfied. Without loss of generality, initially assign a reference frame with origin at $T_1$ and whose $x$-axis is collinear with $\overrightarrow{T_1 T_2}$. In this reference frame, the problem is symmetrical with respect to the $x$-axis.  Hence, we only analyze the case where $T_{3,y}$, the $y$-coordinate of target $T_3$, is positive.  The symmetric case, where $T_{3,y}<0$, is identical.

%% ------------------------------------------------------ %%
\begin{figure}[h]
  \centerline{
    \includegraphics[width=3.125 true in]{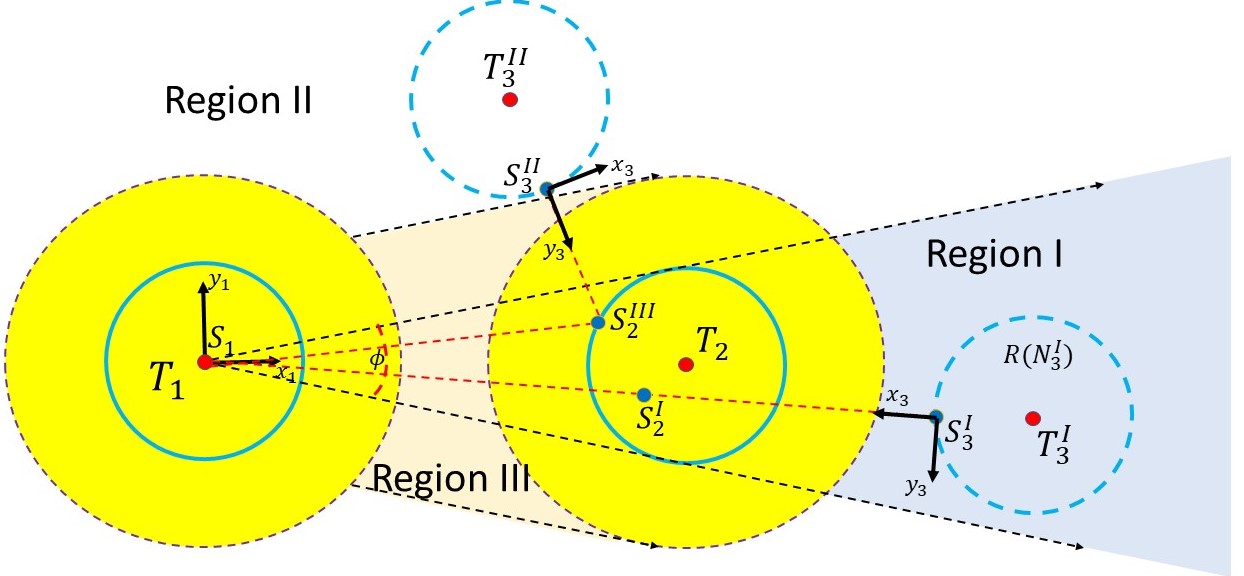}
    }
  \caption{Variations in the geometry of $\Ps_{CPP}$ with respect to target location when no constraints are active. Target $T_3$ cannot lie in circular yellow regions, where constraints must be active.  When $T_3$lies in Region I, $\Ps_{CPP}$ is a straight line from $T_1$ to the boundary of $R(T_3^I)$.  When $T_3$ lies in Region II, $\Ps_{CPP}$ is a two segment path intersecting the boundaries of $R(T_2)$ and $R(N^{II}_3)$.  When $T_3$ lies in Region III, the targets must be relabeled, with the relabeled problem falling into Region I or Region II. }
  \label{fig:CPP_Variation}
\end{figure}
%% ------------------------------------------------------ %%

Figure \ref{fig:CPP_Variation} shows how path $\Ps_{CPP}$ changes with respect to  placements of $T_3$. When $T_3$ lies in Region I, $\Ps_{CPP}$ is a straight line passing through $R(T_2)$, and it orthogonally intersects the boundary of $R(T_3)$. Sensing node $S_2$ can placed anywhere along the intersection of $\Ps_{CPP}$ and $R(T_2)$.

When $T_3$ lies in Region II, $\Ps_{CPP}$ is a two segment path.  The first segment intersects the boundary of $R(T_2)$ at sensing location, $S_2$, whose location is a solution to the following two equations, which are derived from shortest path conditions:
  \begin{eqnarray} \label{eq:cubic}
   \frac{(T_2-S_2)\cdot(T_3-S_2)}{ r ||T_3-S_2|| } &=& \frac{(T_1-S_2)\cdot(T_3-S_2)}{ r ||T_1-S_2|| }\\
     ||T_2 - S_2||^2 &=& r^2 \nonumber
  \end{eqnarray}
  
When $T_3$ lies in Region III, the path $\Ps_{TSP}$ changes characters: the path through targets $T_1$-$T_2$-$T_3$ is not the shortest TSP path through the targets--the visitation sequence must be reordered $T_1$-$T_3$-$T_2$.  This can be done by swapping the indices of $T_2$ and $T_3$.  The new relabeled problem lies in either Region I or Region II. Thus, Region III is irrelevant.

The shortest path geometry suggests a change of coordinates that simplifies the analysis of the KKT conditions.  In both Regions I and II of Figure \ref{fig:CPP_Variation}, let $\Fs_3$ denote a reference frame whose origin is coincident with $S_3$, and whose $x$-axis (denoted $x_3$) is collinear with $\overrightarrow{T_3 S_2}$.  See Figure \ref{fig:CPP_Variation}.  Let $\mu_3$ denote the distance along the positive $x_3$ axis, starting with $\mu_3=0$ at $S_3$.  Move target $T_3$ along the segment $\overrightarrow{T_3 S_2}$, while keeping $T_1$ and $T_2$ fixed.  Target $T_3$ can translate along $\overrightarrow{T_3 S_2}$ until the constraint $||T_3-T_2||\ge r$ becomes active.  Let $D$ denote the distance between its initial location and the location where the constraint becomes active.

The location of $T_3$ in either Region I or Region II can thus be uniquely parametrized by the location $S_2$ on the shortest path to $T_3$ and the distance $D$. Additionally, $\Fs_3$ is well defined throughout these regions.  Let $\eta$ denote the angle between $x_3$-axis and the line segment $\overrightarrow{T_3 T_2}$ as $T_3$ translates along axis $x_3$, this angle monotonically increases.

As $T_3$ translates along $x_3$, the value of $\Delta$ can be expressed as an {\em approximately linear fractional function} form:
  \begin{equation} \label{eq:RegionII_Delta}
    \Delta = \frac{l^o_{TSP} - \mu_3 \cos\eta(\delta_3)}{l^o_{CPP} - \mu_3}
  \end{equation}
where $l^o_{TSP}= ||T_2-T_1|| + ||T_3-T_2||$ and $l^o_{CPP} = ||S_3-T_1||$.  Note that the angle $\eta$ slowly increases as $T_3$ translates.  Thus, $\partial\eta/\partial \delta_3>0$ for all $\delta_3\ \in\ [0,T]$.

The following Lemma, whose proof is a simple calculation, yields insight into (\ref{eq:RegionII_Delta}).

\begin{lemma} \label{lemma:PositiveGradient}
Let $N$ and $M$ be positive constants, with $N>M$.  Let $0<\beta(\mu)<1$.  and let $\partial\beta(\mu)/\partial \delta >0$ for all $\mu \ \in\ [0,D]$, for some $\infty> T>0$.  Then, the gradient function (\ref{eq:proof_ratio}) with respect to $\mu$ is positive for all $\mu\ \in\ [0,D]$:
  \begin{equation}\label{eq:proof_ratio}
    \frac{N - \beta(\mu) \mu} {M-\mu}\ .
  \end{equation}
\end{lemma}

Note that the geometry of $\Ps_{CPP}$ is independent of small variations in $T_2$ when $T_3$ lies in Region I.  Hence, $\partial \Delta/\partial T_2 = \vec{0}$.  However, Lemma \ref{lemma:PositiveGradient} implies that $\partial\Delta/ \partial x_3 > \vec{0} $ for all $T_3$ located in Region I.  Thus, the KKT conditions cannot be satisfied in Region I. Moreover, the maximum value of $\Delta$ in Region I occurs when $R(T_3)$ nearly touches $R(T_2)$.

A similar condition holds in Region II of Figure \ref{fig:CPP_Variation}. In reference frame $\Fs_3$, the function $\Delta$ in Region II takes the same form as Equation (\ref{eq:proof_ratio}), and by Lemma \ref{lemma:PositiveGradient}, its derivative is always positive.  Hence, the KKT-conditions cannot be satisfied for any location of $T_3$ in Region II.  Thus, one or more constraints must be active at the maxima of $\Delta$.

%% ------------------------------------------------------ %%
\begin{figure}[h]
  \centerline{
    \includegraphics[height=1.8 true in]{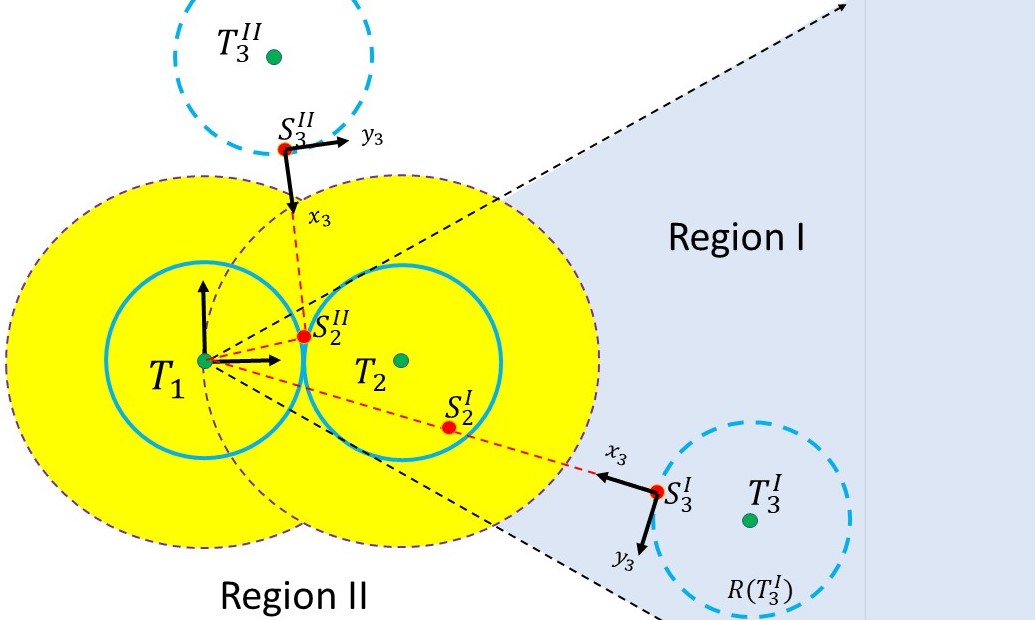}
    }
  \caption{Variations in the geometry of $\Ps_{CPP}$ with respect to target location for the
    constrained case of $||T_1-T_2||=r$.  When target $T_3$ is located in Region I, $\Ps_{CPP}$
    is a straight line from $T_1$ to the boundary of $R(T_3^I)$.  In Region II, $\Ps_{CPP}$ is a
    two-segment path that first intersects the boundary of $R(T_2)$,  and then intersects the
    boundary of $R(T_3^II)$.}
  \label{fig:CPP_Variation12}
\end{figure}
%% ------------------------------------------------------ %%

{\bf Case of One Active Constraint.} We will next consider the cases where a single constraint is active (one pair of target nodes are separated by distance $2r$).  There are three pairings to consider: $(R(T_1),R(T_2))$, $(R(T_1),R(T_3))$, and $(R(T_2),R(T_3))$.  Figure \ref{fig:CPP_Variation12} depicts the first constraint pairing, $(T_1,T_2)$.  As in Figure \ref{fig:CPP_Variation}, the problem is symmetric with respect to an axis passing through $T_1$ and $T_2$.  

The KKT conditions under this constraint take the form:
\begin{eqnarray}
  \vec{0} & \in & \frac{\partial \Ls}{\partial T_2} \ \ \ \Rightarrow \ \ \
         \vec{0}\ \in\ -\frac{\partial \Delta}{\partial T_2} - \lambda_{1,2} (T_2 - T_1) \label{eq:fub} \\
  \vec{0} & \in & \frac{\partial \Ls}{\partial T_3} \ \ \ \Rightarrow \ \ \
         \vec{0}\ \in\  -\frac{\partial\Delta}{\partial T_3} \label{eq:fub2}
\end{eqnarray}

Using the same analysis of the three unconstrained targets in Figure \ref{fig:CPP_Variation}, it is easy to see that in both Region I and Region II of Figure \ref{fig:CPP_Variation12} that $\partial \Delta/\partial x_3$ is always positive, and therefore the KKT conditions cannot be satisfied in the interiors of Regions I and II. Moreover, for both regions, $\Delta$ is maximized on the boundary, where $||T_3-T_2|| = 2r$.

When $T_1$ and $T_3$ are constrained to be a distance $2r$ apart, the path passing through the sequence $T_1$-$T_2$-$T_3$ is not the shortest possible path through the targets -- the visitation sequence must be reordered as $T_1$-$T_3$-$T_2$.  This can be done by swapping the indices of $T_2$ and
$T_3$. The newly relabeled problem thus has $T_1$ and $T_2$ constrained to be a distance $2r$ apart, and this case was analyzed above.

%% ------------------------------------------------------ %%
\begin{figure}[h]
  \centerline{
    \includegraphics[height=1.7 true in]{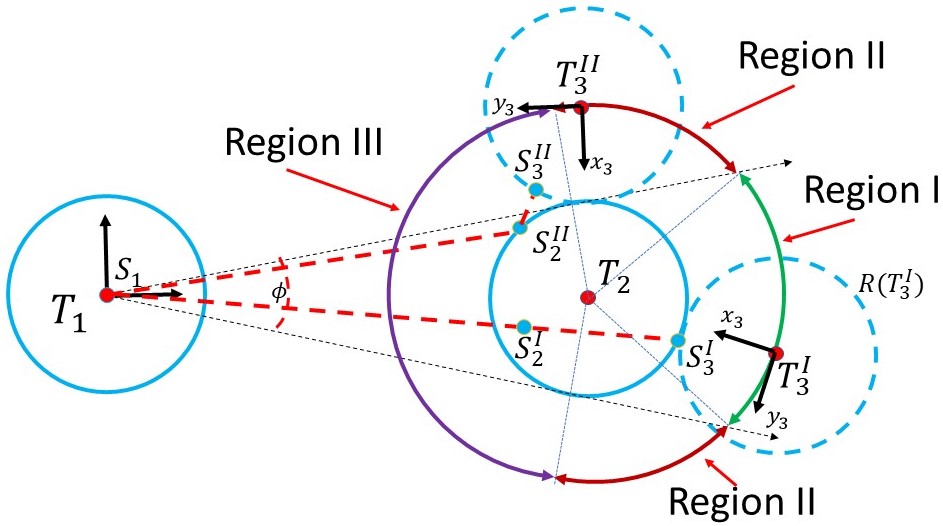}
    }
  \caption{Geometry of $\Ps_{CPP}$ when $||T_3 - T_2|| = 2r$. }
  \label{fig:CPP_Variation23}
\end{figure}
%% ------------------------------------------------------ %%

In the remaining  single constraint case to consider,  $T_2$ and $T_3$ lie a distance $2r$ apart (see Figure \ref{fig:CPP_Variation23}) so that target $T_3$ is constrained to lie on a circle of radius $2r$ centered on $T_2$.  The KKT-conditions in this case are:
  \begin{eqnarray}
    \vec{0} & \in & -\frac{\partial\Delta}{\partial T_2} + \lambda_{2,3}(T_3-T_2) \label{eq:C23a} \\
    \vec{0} & \in & -\frac{\partial\Delta}{\partial T_3} - \lambda_{2,3}(T_3-T_2).\label{eq:C23b}
  \end{eqnarray}

As seen in Figure \ref{fig:CPP_Variation23}, the geometry of $\Ps_{CPP}$ changes as $T_3$'s location varies over the circle.  In Region I of Fig. \ref{fig:CPP_Variation23}, $\Ps_{CPP}$ forms a straight line from $T_1$ that intersects the boundary of $R(T_3)$ orthogonally.  In Region II, $\Ps_{CPP}$ is a two segment path: the first segment connects $T_1$ to sensing location $S_2$, while the second segment departs $S_2$ and intersects the boundary of $R(T_3)$ orthogonally at $S_3$.  The location of $S_2$ is determined analogously to Equation (\ref{eq:cubic}).  The character of $\Ps_{TSP}$ changes in Region III: the path sequence $T_1$-$T_2$-$T_3$ is not the shortest path through the targets, starting from $T_1$.  As above, the indices of targets $T_2$ and $T_3$ are swapped to yield a correct path, resulting in a configuration that lies in either Region I or Region II.  Thus, Region III can be ignored. 

Let reference frame $\Fs_3$ be centered at $T_3$, with its $x$-axis colinear to $\overrightarrow{T_3T_2}$ (see Figure \ref{fig:CPP_Variation23}).  For all locations of $T_3$ in Region I, the $y_3$ component of $\lambda_{2,3}(T_3-T_2)$ in Equation (\ref{eq:C23a}) is zero. The magnitude of $l_{TSP}$ does not vary in Region I, and hence  $\partial l_{TSP}/\partial y_3 = 0$. Consequently, the gradient of $\Delta$ simplifies to
  \[ \frac{\partial \Delta}{\partial x_3} = -(\frac{\Delta}{l_{CPP}}) 
    \frac{\partial l_{CPP}}{\partial x_3}. \]  
This term take a nonzero value at all locations of $T_3$ except where $T_1$, $T_2$, and $T_3$ are colinear.  Thus, the $y$-component of (\ref{eq:C23b}) is
always nonzero, except at the colinear target condition. It can be shown that the colinear configuration is a saddle point.  Thus, $\Delta$ is not extremized in Region I.  

In Region II of Fig. \ref{fig:CPP_Variation23}, with $\Fs_3$ chosen as shown, the $y$-component of the term $\lambda_{23}(T_3-T_2)$ is zero.  But since $L_{CPP}$ varies as $T_3$ varies along Region II, Eq. (\ref{eq:C23b}) is not satisfied, and therefore there is no stationary point of $\Delta$ in Region II.

\begin{figure}[h]
  \centerline{
    \includegraphics[height=1.7 true in]{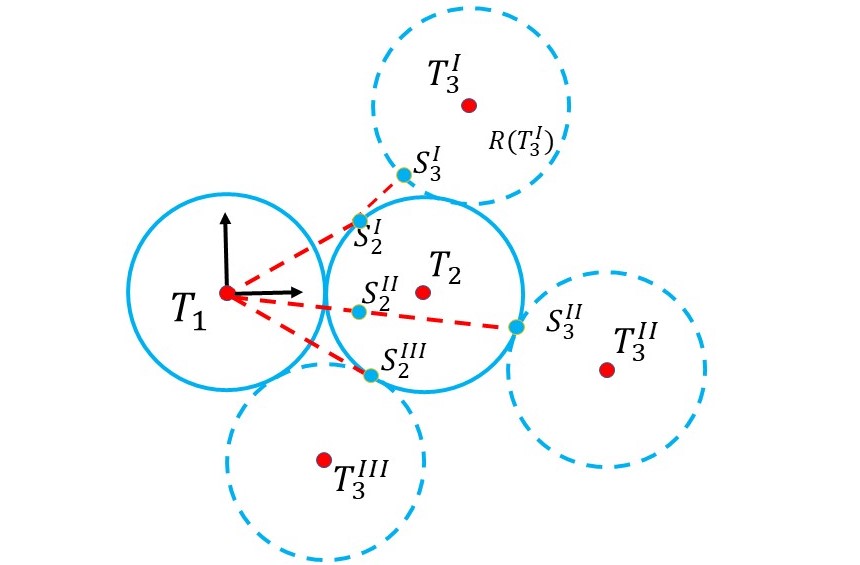}
    }
  \caption{Geometry of Sensing Regions and $\Ps_{CPP}$ when two constraints are active: $||T_2 - T_1|| = ||T_3 - T_2|| = 2r$. Three different configurations of $R(T_3)$ are shown. Path $\Ps_{TSP}$ has the same length for all configurations. Path $\Ps_{CPP}$ (red dashed line) is shortest when the sensing regions form an equilateral triangle.}
  \label{fig:2Constraints}
\end{figure}

{\bf Case of Two Active Constraints.}  The are possible three groupings of  paired constraints between sensing regions $R(T_1)$, $R(T_2)$, and $R(T_3)$: $((R(T_1),R(T_2)) \& (R(T_2),R(T_3))$; $((R(T_1),R(T_3))$ $\&$ $(R(T_2),R(T_3)))$ and $((R(T_1),R(T_2)) \& (R(T_1),R(T_3)))$. Figure \ref{fig:2Constraints} depicts the geometry when constraints are active between $R(T_1)$ and $R(T_2)$, as well as $R(T_2)$ and $R(T_3)$. Different candidate locations of $T_3$ are shown.
%% ------------------------------------------------------ %%
%% ------------------------------------------------------ %%

Note that $l_{TSP}$ is the same for all configurations of $T_3$ that satisfy the constraints.  Hence, the ratio $\Delta$ is maximized when $\Ps_{CPP}$ achieves its shortest length.  A direct calculation shows that $|\Ps_{CPP}|$ is minimized when the three sensing regions form a tightly nested equilateral triangle.  An analogous result holds for the other constraint combinations.

In summary, the worst case ratio of the ratio $\Delta$ for the situation of
3 targets occurs when all inter-target constraints are active, resulting in a equilateral triangle configuration.

\subsection{Case of 4 Targets}
%% ------------------------------------------------------ %%
\begin{figure}[h]
  \centerline{
    \includegraphics[height=1.7 true in]{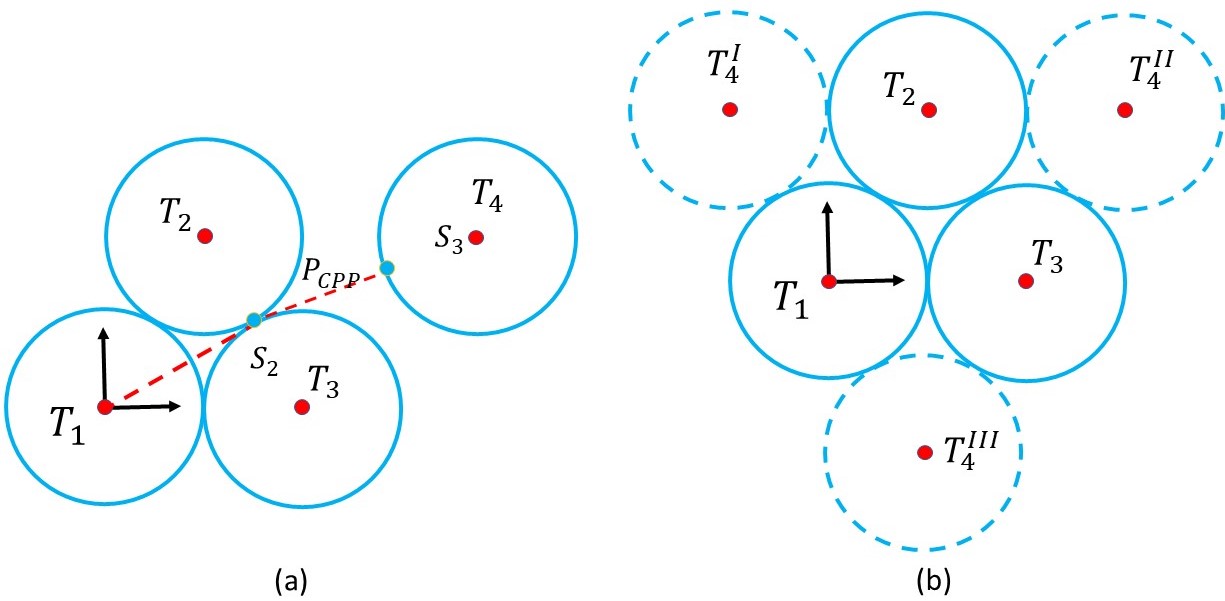}
    }
  \caption{Geometry of Sensing Regions and $\Ps_{CPP}$ for the case of 4 targets.  (a)  two constraints are active: $||T_2 - T_1|| = ||T_3 - T_2|| = 2r$, and $R(T_4)$ is unconstrained. (b) Three different extremizing configurations of $R(T_4)$ are shown.}
  \label{fig:4Targets}
\end{figure}
%% ------------------------------------------------------ %%

Figure \ref{fig:4Targets}(a) depicts a problem with 4 targets.  The first three targets are constrained to form an equilateral triangle, with the fourth target placed at an arbitrary, but unconstrained, location.  It can be shown, using the KKT-like analysis described above, that if targets $T_2$ or $T_3$ are displaced from their constrained configuration to a nearby unconstrained location, while the other targets remain fixed, then the value of $\Delta$ decreases.  That is, targets $T_2$ and $T_3$ must be constrained as shown to maximize $\Delta$, regardless of the location of $T_4$.  Again, using a KKT analysis, one can show that $\Delta$ is extremized when $R(T_4)$ is constrained with respect to the other sensing regions.  Figure \ref{fig:4Targets} shows the three extremizing locations of target $T_4$, indicated by $T_4^I$, $T_4^{II}$, and $T_4^{III}$. A direct calculation shows that target location $T_4^{II}$ maximizes $\Delta$.

\subsection{Case of 5 and more targets}

\begin{figure}[h]
  \centerline{
    \includegraphics[height=1.4 true in]{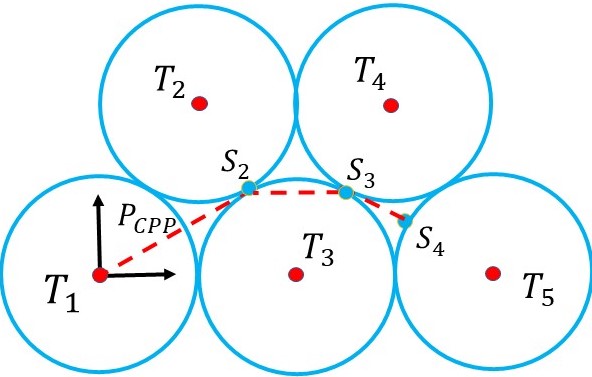}
    }
  \caption{Extremizing Geometry of 5 Targets. }
  \label{fig:5Targets}
\end{figure}

When a fifth target is added, the results above can be extrapolated to show that $\Delta$ is extremized when the sensing regions are tightly packed.  Direct calculations show that maximum magnitude of $\Delta$ is realized in the configuration shown in Figure \ref{fig:5Targets}.  For six or more targets, the same analytical approach leads to the linear packing arrangement of the sensing regions that is seen in Figure \ref{fig:RatioDiagram}.  As the number of targets increases, there are a discrete number of sensing circle packings that lead to the largest $\Delta$ ratio.  However, the linear packing arrangement is as short as any other packing arrangment.  Thus, the linear packing arrangement suffices for our analysis. 

%% ------------------------------------------------------ %%
%% ------------------------------------------------------ %%

%% ------------------------------------------------------ %%

\end{appendices}

%%% ------------------------------------------------------------- %%%
%\input{ProofAppendix.tex}
%%%%%%%%%%%%%%%%%%%%%%%%%%%%%%%%%%%%%%%%%%%%%%%%%%%%%%%%%%%%%%%%%%%%%%%%%%%%%%%%
%\bibliography{Coverage,CoveringTour}

\bibliography{new_minlp,Coverage,CoveringTour,CoveringTour2,TSP,SetCover,Algo,TSP_withNeighborhoods,elonbib,minlp}

\begin{thebibliography}{10}

\bibitem{acar&choset_06}
E.~Acar, H.~Choset, and J.~Y. Lee.
\newblock Sensor based coverage with extended range detectors.
\newblock {\em IEEE Transactions on Robotics}, 22(1):189--198, 2006.

\bibitem{demining_choset}
E.~Acar, H.~Choset, Y.~Zhang, and M.~Schervish.
\newblock A survey on inspecting structures using robotic systems.
\newblock {\em Int. J. Robotics Research}, 22:441--461, 2016.

\bibitem{inspection_survey}
Randa Almadhoun, Tarek Taha, Lakmal Seneviratne, Jorge Dias, and Guowei Cai.
\newblock A survey on inspecting structures using robotic systems.
\newblock {\em Int. J. Advanced Robotics Systems}, pages 1--18, 2016.

\bibitem{belotti_minlp}
Pietro Belotti, Christian Kirches, Sven Leyffer, Jeff Linderoth, James Luedtke, and Ashutosh Mahajan.
\newblock Mixed-integer nonlinear optimization.
\newblock {\em Acta Numerica}, 22, 05 2013.

\bibitem{bircher_3D}
Andreas Bircher, Mina Kamel1, Kostas Alexis1, Michael Burri, Philipp Oettershagen1, Sammy Omari, Thomas Mantel1, and Roland Siegwart.
\newblock Three-dimensional coverage path planning via viewpoint resampling and tour optimization for aerial robots.
\newblock {\em Autonomous Robots}, 40:1059--1078, 2016.

\bibitem{ellipsoid_survey}
R.~Bland, D.~Goldfarb, and M.~Todd.
\newblock The ellipsoid method: A survey.
\newblock {\em Operations Research}, 29, 12 1981.

\bibitem{BONMIN}
P.~Bonami, L.~T. Biegler, A.~R. Conn, G.~Cornuejols, I.~E. Grossmann, C.~D. Laird, A.~Lodi J.~Lee, F.~Margot, and A.~Waechter.
\newblock An algorithmic framework for convex mixed integer nonlinear programs.
\newblock {\em Discrete Optimization}, 5(2):186--204, 2008.

\bibitem{9561213}
Joel~W. Burdick, Amanda Bouman, and Elon Rimon.
\newblock From multi-target sensory coverage to complete sensory coverage: An optimization-based robotic sensory coverage approach.
\newblock In {\em 2021 IEEE International Conference on Robotics and Automation (ICRA)}, pages 10994--11000, 2021.

\bibitem{christofides_n_worst-case_1976}
{Christofides, N.}
\newblock Worst-{Case} analysis of a new heuristic for the travelling salesman problem.
\newblock Technical Report 388, Carnegie-Mellon University, Pittsburgh, PA, 1976.

\bibitem{coutinho_branch-and-bound_2016}
Walton~Pereira Coutinho, Roberto Quirino~do Nascimento, Artur~Alves Pessoa, and Anand Subramanian.
\newblock A {Branch}-and-{Bound} {Algorithm} for the {Close}-{Enough} {Traveling} {Salesman} {Problem}.
\newblock {\em INFORMS Journal on Computing}, 28(4):752--765, November 2016.

\bibitem{current_covering_1989-1}
John~R. Current and David~A. Schilling.
\newblock The {Covering} {Salesman} {Problem}.
\newblock {\em Transportation Science}, 23(3):208--213, August 1989.

\bibitem{danner_randomized_2000}
T.~Danner and L.E. Kavraki.
\newblock Randomized planning for short inspection paths.
\newblock In {\em {IEEE} {Int.} {Conf.} on {Robotics} and {Automation}}, pages 971--976 vol.2, San Francisco, CA, USA, 2000. IEEE.

\bibitem{dumitrescu_approximation_2003}
Adrian Dumitrescu and Joseph S~B Mitchell.
\newblock Approximation algorithms for {TSP} with neighborhoods in the plane.
\newblock {\em Journal of Algorithms}, 48(1):135--159, 2003.

\bibitem{christensen_planning_2017}
Brendan Englot and Franz Hover.
\newblock Planning {Complex} {Inspection} {Tasks} {Using} {Redundant} {Roadmaps}.
\newblock In Henrik~I. Christensen and Oussama Khatib, editors, {\em Robotics {Research}}, volume 100, pages 327--343. Springer International Publishing, Cham, 2017.

\bibitem{englot_three-dimensional_2013}
Brendan Englot and Franz~S. Hover.
\newblock Three-dimensional coverage planning for an underwater inspection robot.
\newblock {\em Int. J. Robotics Research}, 32(9-10):1048--1073, August 2013.

\bibitem{Galceran_underwater}
E.~Galceran and M.~Carrares.
\newblock Efficient seabed coverage path planning for asvs and auvs.
\newblock In {\em IEEE/RSJ Int. Conf. on Intelligent Robots and Systems}, pages 88--93, 2012.

\bibitem{galceran_survey}
E.~Galceran and M.~Carreras.
\newblock A survey on coverage path planning for robotics.
\newblock {\em Robotics and Autonomous Systems}, 61(4):146--155, 2013.

\bibitem{gentilini_travelling_2013}
Iacopo Gentilini, François Margot, and Kenji Shimada.
\newblock The travelling salesman problem with neighbourhoods: {MINLP} solution.
\newblock {\em Optimization Methods and Software}, 28(2):364--378, April 2013.

\bibitem{golden_generalized_2012}
Bruce Golden, Zahra Naji-Azimi, S.~Raghavan, Majid Salari, and Paolo Toth.
\newblock The {Generalized} {Covering} {Salesman} {Problem}.
\newblock {\em INFORMS Journal on Computing}, 24(4):534--553, November 2012.

\bibitem{gb08}
Michael Grant and Stephen Boyd.
\newblock Graph implementations for nonsmooth convex programs.
\newblock In V.~Blondel, S.~Boyd, and H.~Kimura, editors, {\em Recent Advances in Learning and Control}, Lecture Notes in Control and Information Sciences, pages 95--110. Springer-Verlag Limited, 2008.
\newblock \url{http://stanford.edu/~boyd/graph_dcp.html}.

\bibitem{cvx}
Michael Grant and Stephen Boyd.
\newblock {CVX}: Matlab software for disciplined convex programming, version 2.1.
\newblock \url{https://cvxr.com/cvx}, March 2014.

\bibitem{hoogeveen_analysis_1991}
J.A. Hoogeveen.
\newblock Analysis of {Christofides}'s heuristic: {Some} paths are more difficult than cycles.
\newblock {\em Operat. Res. Lett.}, 10:291--295, 1991.

\bibitem{OptiToolbox}
{Inverse Problem Limited}.
\newblock {{Opti Toolbox}} - {{A Free Matlab Tool for Optimization }}.
\newblock https://www.inverseproblem.co.nz/OPTI/, 2019.

\bibitem{10.1145/321992.321993}
Donald~B. Johnson.
\newblock Efficient algorithms for shortest paths in sparse networks.
\newblock {\em J. ACM}, 24(1):1–13, jan 1977.

\bibitem{Kapoor1997-bq}
S~Kapoor, S~N Maheshwari, and J~S~B Mitchell.
\newblock An efficient algorithm for euclidean shortest paths among polygonal obstacles in the plane.
\newblock {\em Discrete \& Computational Geometry}, 18(4):377--383, December 1997.

\bibitem{kolmogorov_blossom_2009}
Vladimir Kolmogorov.
\newblock Blossom {V}: a new implementation of a minimum cost perfect matching algorithm.
\newblock {\em Mathematical Programming Computation}, 1(1):43--67, July 2009.

\bibitem{minlp_solvers}
J.~Kronqvist, D.~Bernal, A.~Lundell, and I.~Grossmann.
\newblock A review and comparison of solvers for convex \mbox{MINLP}.
\newblock {\em Optimization and Engineering}, 20, 06 2019.

\bibitem{lauterbach_eins3d_2019}
Helge~A. Lauterbach, C.~Bertram Koch, Robin Hess, Daniel Eck, Klaus Schilling, and Andreas Nuchter.
\newblock The {Eins3D} project — {Instantaneous} {UAV}-{Based} {3D} {Mapping} for {Search} and {Rescue} {Applications}.
\newblock In {\em {IEEE} {Int} {Symp} {Safety}, {Security}, and {Rescue} {Robotics}}, pages 1--6, Würzburg, Germany, September 2019. IEEE.

\bibitem{loizou_cdc16}
S.~G. Loizou and C.~C. Constantinou.
\newblock Multi-robot coverage on dendritic topologies under communication constraints.
\newblock In {\em IEEE 55th Conference on Decision and Control}, pages 43--48, 2016.

\bibitem{tobia_obstacles}
T.~Marcussi, M.~Petersen, D.~Wrangel, and R.~Tedrake.
\newblock Motion planning around obstacles with convex optimization.
\newblock {\em arXiv:2205.04422}, May 2022.

\bibitem{mitchell-2017}
Joseph~S.B. Mitchell, Csaba~D. Tóth, Jacob~E. Goodman, and Joseph O'Rourke.
\newblock {\em {Shortest paths and networks}}.
\newblock Chapman \& Hall/CRC, 3rd edition, 2017.

\bibitem{plonski&isler}
Patrick~A Plonski and Volkan Isler.
\newblock Approximation algorithms for tours of height-varying view cones.
\newblock {\em The International Journal of Robotics Research}, 38(2-3):224--235, 2019.

\bibitem{preparata2012computational}
F.P. Preparata and M.I. Shamos.
\newblock {\em Computational Geometry: An Introduction}.
\newblock Monographs in Computer Science. Springer New York, 2012.

\bibitem{4567905}
Michael~Ian Shamos and Dan Hoey.
\newblock Geometric intersection problems.
\newblock In {\em 17th Annual Symposium on Foundations of Computer Science (sfcs 1976)}, pages 208--215, 1976.

\bibitem{tokekar_sensor_2016}
Pratap Tokekar, Joshua~Vander Hook, David Mulla, and Volkan Isler.
\newblock Sensor {Planning} for a {Symbiotic} {UAV} and {UGV} {System} for {Precision} {Agriculture}.
\newblock {\em IEEE Transactions on Robotics}, 32(6):1498--1511, December 2016.

\bibitem{vecchietti_modeling_2003}
A.~Vecchietti, S.~Lee, and I.E. Grossman.
\newblock Modeling of {Discrete}/{Continuous} {Optimization} {Problems}: {Characterization} and {Formulation} of {Disjunctions} and {Their} {Relaxations}.
\newblock {\em Computers and Chemical Engineering}, 27(3):433--448, 2003.

\bibitem{WELZL1985167}
Emo Welzl.
\newblock Constructing the visibility graph for n-line segments in o(n2) time.
\newblock {\em Information Processing Letters}, 20(4):167--171, 1985.

\bibitem{approx_alg}
D.~P. Williamson and D.~B. Shmoys.
\newblock {\em The Design of Approximation Algorithms}.
\newblock Cambridge University Press, N.Y., 2011.

\end{thebibliography}

%\bibliography{new_minlp}

%\bibliographystyle{unsrt}

%\printbibliography

%\bibliographystyle{ieeetr}
%%%%%%%%%%%%%%%%%%%%%%%%%%%%%%%%%%%%%%%%%%%%%%%%%%%%%%%%%%%%%%%%%%%%%%%%%%%%%%%%

\end{document}